\documentclass[twoside,twocolumn]{article}
\usepackage[left=2cm,right=2cm,top=2cm,bottom=2cm]{geometry}
\usepackage[belowskip=0pt,aboveskip=0pt, labelfont=bf]{caption}
\usepackage{lmodern}
\usepackage{float}
\usepackage{amssymb,amsmath}
\usepackage{mathtools}
\usepackage{ifxetex,ifluatex}
\usepackage[T1]{fontenc}
\usepackage[utf8]{inputenc}
\usepackage{textcomp}
\usepackage{siunitx}
\usepackage{textcomp}
\usepackage[toc,page]{appendix}
\usepackage{nicefrac}
\usepackage{bm}
\usepackage[utf8]{inputenc}
\usepackage[T1]{fontenc}
\usepackage[super,sort&compress,comma]{natbib}

\newcommand*{\citen}[1]{%
  \begingroup
    \romannumeral-`\x % remove space at the beginning of \setcitestyle
    \setcitestyle{square,numbers}%
    \cite{#1}%
  \endgroup   
}
\usepackage{balance}
\usepackage{times,mathptmx}
\usepackage{amsmath}
\makeatletter
\usepackage{physics}
\usepackage{graphicx}
\usepackage{threeparttable}
\usepackage{makecell}
\usepackage{array}
\usepackage{booktabs}
\usepackage[svgnames]{xcolor}
\usepackage{makecell}
\usepackage{array}
\usepackage{tabularx}
\usepackage{booktabs}
\usepackage{rotating} % To rotate a table 90 degree in a new page
\usepackage{pdflscape}
\def\fillandplacepagenumber{%
 \par\pagestyle{empty}%
 \vbox to 0pt{\vss}\vfill
 \vbox to 0pt{\baselineskip0pt
   \hbox to\linewidth{\hss}%
   \baselineskip\footskip
   \hbox to\linewidth{%
     \hfil\thepage\hfil}\vss}}

\usepackage{color, colortbl}
\definecolor{Gray}{gray}{0.75}

\definecolor{yellow}{rgb}{1,1,0.8039}
\definecolor{LightCyan}{rgb}{0.88,1,1}
\usepackage{arydshln}

\newcommand{\tabitem}{\makebox[1em][r]{\textbullet}}

\usepackage[noframe]{showframe}
\usepackage{framed}
\usepackage{chemformula}
 {\endMakeFramed}
\definecolor{shadecolor}{gray}{0.85}
\usepackage[super,sort&compress,comma]{natbib}
\usepackage{imakeidx}
\makeindex

\usepackage{nomencl}
\makenomenclature

\usepackage[toc,acronym]{glossaries}
\usepackage{glossary-mcols}
\usepackage{enumitem}
\makeglossaries

\usepackage{ifthen}
\usepackage{etoolbox}
\makeatletter
\preto{\@tabular}{\parskip=0pt}
\makeatother
\renewcommand\nomgroup[1]{%
  \ifthenelse{\equal{#1}{A}}{%
    \item[\textbf{Acronyms}]}{%                A - Acronyms
  \ifthenelse{\equal{#1}{L}}{%
    \item[\textbf{Latin Symbols}]}{%           R - Roman
  \ifthenelse{\equal{#1}{G}}{%
    \item[\textbf{Greek Symbols}]}{%           G - Greek
  \ifthenelse{\equal{#1}{S}}{%
    \item[\textbf{Superscripts}]}{%            S - Superscripts
  \ifthenelse{\equal{#1}{U}}{%
    \item[\textbf{Subscripts}]}{%              U - Subscripts
  \ifthenelse{\equal{#1}{X}}{%
    \item[\textbf{Other Symbols}]}{%           X - Other Symbols
  {}}}}}}}}

\newcommand{\nomunit}[1]{%
\renewcommand{\nomentryend}{\hspace*{\fill}#1}}
\usepackage[acronym, toc]{glossaries}
\newacronym{EV}{EV}{Electric Vehicles}
\newacronym{ROC}{ROC}{Receiver Operating Characteristics}
\newacronym{AUROC}{AUROC}{Area Under Receiver Operating Characteristics}
\newacronym{AUPRC}{AUPRC}{Area Under Precision-Recall Curve}
\newacronym{BEV}{BEV}{Battery Electric Vehicles}
\newacronym{HEV}{HEV}{Hybrid Electric Vehicles}
\newacronym{PHEV}{PHEV}{Plug-in Hybrid Electric Vehicles}
\newacronym{SOC}{SOC}{State-of-Charge}
\newacronym{OCV}{OCV}{Open-circuit Voltage}
\newacronym{SOH}{SOH}{State-of-Health}
\newacronym{Li-ion}{$\textrm{Li}^{+}$}{lithium-ions}
\newacronym{ECM}{ECM}{Equivalent Circuit Models}
\newacronym{BMS}{BMS}{Battery Management System}
\newacronym{SE}{SE}{Solid Electrolyte}
\newacronym{PE}{PE}{Positive Electrode}
\newacronym{NE}{NE}{Negative Electrode}
\newacronym{AM}{AM}{Active Materials}
\newacronym{ML}{ML}{Machine Learning}
\newacronym{AR}{AR}{Autoregressive}
\newacronym{SVM}{SVM}{Support Vector Machine}
\newacronym{RVM}{RVM}{Relevance Vector Machine}
\newacronym{GPR}{GPR}{Gaussian Process Regression}
\newacronym{ANN}{ANN}{Artificial Neutral Network}
\newacronym{RL}{RL}{Reinforcement Learning}
\newacronym{IQR}{IQR}{Interquartile Range}
\newacronym{SD}{SD}{Standard Deviation}
\newacronym{MAD}{MAD}{Median Absolute Deviation}
\newacronym{t-SNE}{t-SNE}{t-distributed Stochastic Neighbor Embedding}
\newacronym{LOF}{LOF}{Local Outlier Factor}
\newacronym{PCA}{PCA}{Principal Component Analysis}
\newacronym{TP}{TP}{True Positives}
\newacronym{TN}{TN}{True Negatives}
\newacronym{FP}{FP}{False Positives}
\newacronym{FN}{FN}{False Negatives}
\newacronym{IF}{iForest}{Isolation Forest}
\newacronym{KMC}{KMC}{K-Means Clustering}
\newacronym{DBSCAN}{DBSCAN}{Density-Based Spatial Clustering of Applications with Noise}
\newacronym{GMM}{GMM}{Gaussian Mixture Models}
\newacronym{KNN}{KNN}{K-Nearest Neighbors}
\newacronym{MCC}{MCC}{Matthew Correlation Coefficient}
\newacronym{RD}{RD}{Reachability Distance}
\newacronym{LRD}{LRD}{Local Reachability Distance}
\newacronym{PCs}{PCs}{Principal Components}
\newacronym{KPI}{KPI}{Key Performance Indicator}
\newacronym{OSBAD}{OSBAD}{Open-Source Benchmark of Anomaly Detection}
\newacronym{HP}{HP}{hyperparameter}

\newacronym{LFP}{LFP}{Lithium Iron Phosphate (\ch{LiFePO4})}
\newacronym{NMC}{NMC}{Nickel Manganese Cobalt oxides}
\newacronym{SI}{SI}{Supplementary Information}

\usepackage{expl3,xparse,xcoffins,titling,kantlipsum}
\ExplSyntaxOn
\coffin_new:N \l_my_abstract_coffin
\dim_zero_new:N \l_my_width_dim
\keys_define:nn { my / abstract }
  {
    width .code:n = {
      \dim_set:Nn \l_my_width_dim {#1\textwidth}
    }
  }
\NewDocumentCommand \myabstract { O {width=.8} m }{%
  \keys_set:nn { my / abstract } { #1 }
  \SetVerticalCoffin \l_my_abstract_coffin {\l_my_width_dim} {#2}
  \renewcommand\maketitlehookd{%
    \mbox{}\medskip\par
    \centering
    \TypesetCoffin \l_my_abstract_coffin
  }
}
\ExplSyntaxOff

\usepackage[hidelinks]{hyperref}
\usepackage{url}
\begin{document}

\author{Mei-Chin Pang,$^{a}$ Suraj Adhikari,$^{a}$ Takuma Kasahara,$^{b}$ Nagihiro Haba,$^{b}$ Saneyuki Ohno$^{b}$}
\title{An Open-Access Benchmark of Statistical and Machine-Learning Anomaly Detection Methods for Battery Applications}
\myabstract{Battery safety is critical in applications ranging from consumer electronics to electric vehicles and aircraft, where undetected anomalies could trigger safety hazards or costly downtime. In this study, we present OSBAD as an open-source benchmark for anomaly detection frameworks in battery applications. By benchmarking 15 diverse algorithms encompassing statistical, distance-based, and unsupervised machine-learning methods, OSBAD enables a systematic comparison of anomaly detection methods across heterogeneous datasets. In addition, we demonstrate how a physics- and statistics-informed feature transformation workflow enhances anomaly separability by decomposing collective anomalies into point anomalies. To address a major bottleneck in unsupervised anomaly detection due to incomplete labels, we propose a Bayesian optimization pipeline that facilitates automated hyperparameter tuning based on transfer-learning and regression proxies. Through validation on datasets covering both liquid and solid-state chemistries, we further demonstrate the cross-chemistry generalization capability of OSBAD to identify irregularities across different electrochemical systems. By making benchmarking database with open-source reproducible anomaly detection workflows available to the community, OSBAD establishes a unified foundation for developing safe, scalable, and transferable anomaly detection tools in battery analytics. This research underscores the significance of physics- and statistics-informed feature engineering as well as model selection with probabilistic hyperparameter tuning, in advancing trustworthy, data-driven diagnostics for safety-critical energy systems.}

\maketitle

\mbox{}

%%%FOOTNOTES%
{\let\thefootnote\relax\footnotetext{\textit{$^{a}$Data Analytics in Group Research, BASF SE, Carl-Bosch-Strasse 38, 67056 Ludwigshafen am Rhein, Germany. E-mail: mei-chin.pang@basf.com}}}
{\let\thefootnote\relax\footnotetext{\textit{$^{b}$Institute of Multidisciplinary Research for Advanced Materials,\newline Tohoku University. E-mail: saneyuki.ohno.c8@tohoku.ac.jp}}}
%%%END OF FOOTNOTES%
%%%MAIN TEXT%%%%

\section{Introduction}
\label{Sec: Introduction}

\begin{figure*}[ht!]
\centering
\includegraphics[width=\textwidth,trim=2 2 2 2,clip]{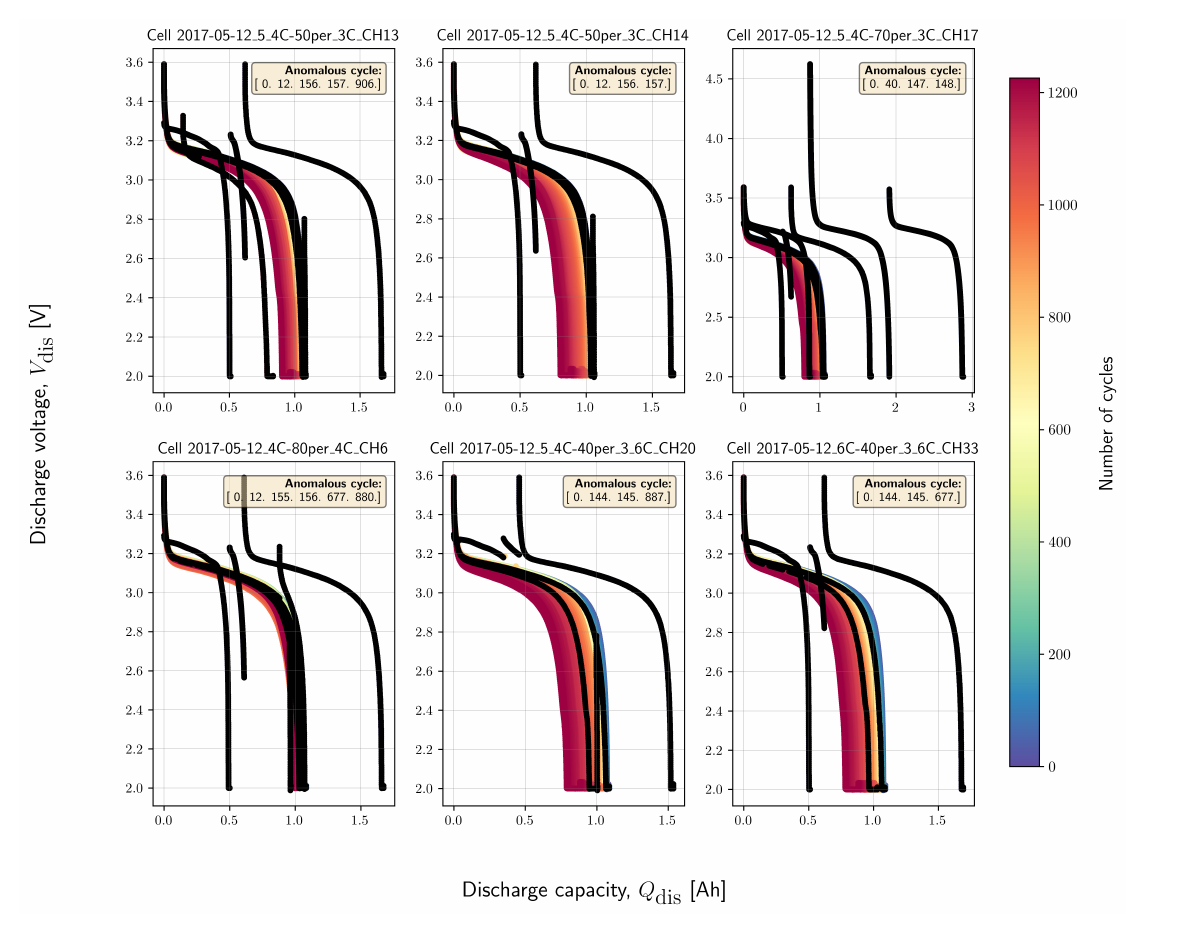}
\caption{\textbf{Identification of anomalous discharge cycles across multiple lithium-ion cells.} Discharge voltage-capacity profiles from the open-source datasets published by \citet{Severson.2019} are shown for six representative cells, with the unique cell index given as the subplot title and the color indicating the number of cycles. Black curves highlight discharge profiles corresponding to anomalous cycles identified in the benchmarking datasets, as listed in each panel's inset. The cells have a nominal capacity of \SI{1.1}{Ah} and a nominal voltage of \SI{3.3}{V}. Anomalous cycles were discovered not only in the first cycle, but also across multiple cycles beyond the first 100 cycles. }
\label{fig: outlier_examples}
\end{figure*}

With the increasing adoption of batteries across various applications, such as consumer electronics, electric vehicles, and short-range electric aircraft, battery performance monitoring is becoming increasingly important to ensure safe and reliable battery operation. Over hundreds to thousands of charge-discharge cycles, various anomalies can occur inevitably due to reasons such as equipment breakdown, manufacturing defects, sensor errors, or even cell failure.\cite{palacin2016batteries} For example, Figure \ref{fig: outlier_examples} illustrates the anomalies discovered in the battery cycling datasets published by Severson \textit{et al.}\cite{Severson.2019}, who had conducted the experiments using 124 commercial high-power lithium-ion A123 cells. The normal voltage operating window was specified to be between \SI{2}{V} and \SI{3.6}{V}, with a nominal capacity of \SI{1.1}{Ah} and a nominal voltage of \SI{3.3}{V}.\cite{Severson.2019} However, it is apparent that both the voltage and capacity measurements are contaminated with different anomalies exceeding the standard operating conditions. In the discharge cycle dataset from the cell-index \colorbox{Gainsboro}{2017-05-12\textunderscore 5\textunderscore 4C-70per\textunderscore 3C\textunderscore CH17}, the cell discharge voltage of the one cycle had reached even up to \SI{4.5}{V}. We found that these anomalies are not limited to specific cell chemistry or measurement protocols, as abnormal observations could also be observed in the \acrfull{OCV} temperature profile and dynamic voltage profile from different data sources and institutions (see Figure \ref{fig: other_anomalies} in the \acrfull{SI}).\cite{birkl2017diagnosis, Bole_dataset.2014, bole2014adaptation}

Divergent data points or anomalies not only affect data quality and model predictions, but the occurrence of anomalies also has profound impacts on battery safety and lifetime performance. One should note that batteries must be operated within a limited temperature and a voltage range to ensure a safe and optimal lifetime performance. The violation of these operating constraints due to overcharge, overdischarge, or extreme temperature conditions can result in internal short circuits, toxic chemical reactions, and self-ignition, which subsequently cause thermal runaway.\cite{sun2020review, wang2005lithium, kong2018li} \citet{sun2020review} have compiled and reviewed numerous real-life battery fire incidents, which highlight the importance of anomaly detection to ensure safety and reliability in battery operations. 

\begin{figure*}[ht!]
\centering
\includegraphics[width=\textwidth,trim=2 2 2 2,clip]{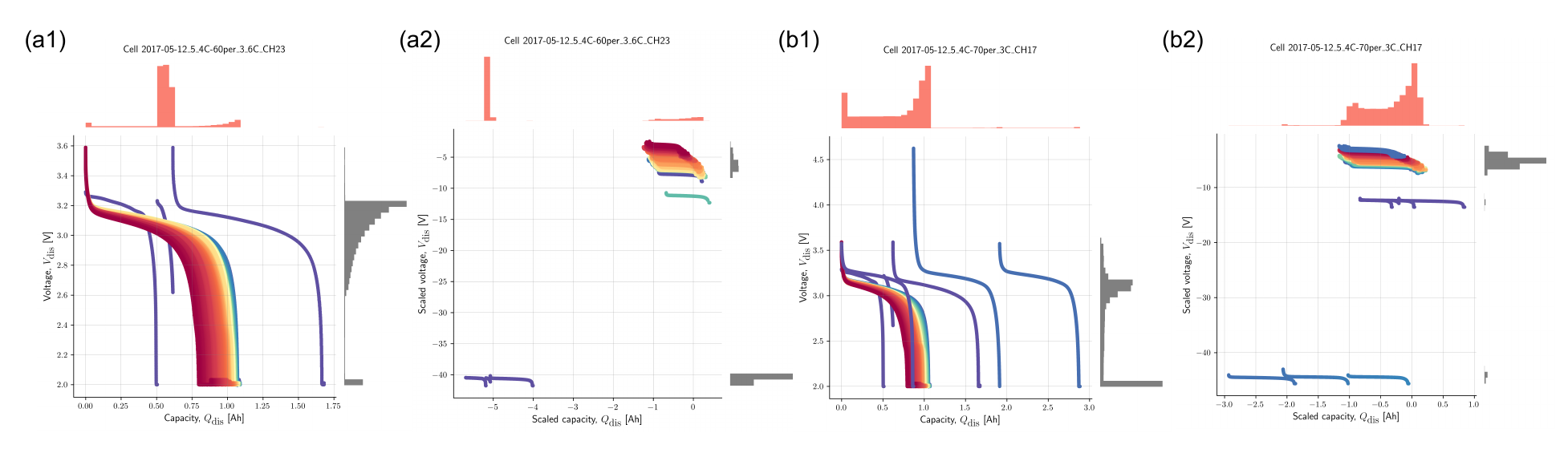}
\caption{\textbf{Voltage and discharge capacity features with marginal histograms before and after statistical transformation, where the transformation steps highlight abnormal cycles from normal ones.} Panels (a1)-(a2) illustrate results for cell \colorbox{Gainsboro}{2017-05-12\textunderscore 5\textunderscore 4C-60per\textunderscore 3\textunderscore 6C\textunderscore CH23} with a simpler anomalous pattern, while (b1)-(b2) show results for cell \colorbox{Gainsboro}{2017-05-12\textunderscore 5\textunderscore 4C-70per\textunderscore 3C\textunderscore CH17} with a more complex pattern.}
\label{fig: compare_feature_transformation_effects}
\end{figure*}

Although data-driven battery lifetime prediction has been extensively studied,\cite{Severson.2019, reniers2019review, Li.2019, Ng.2020} reports on data-driven anomaly detection remains limited. Haider et al.\cite{haider2020data} applied K-shape hierarchical clustering to monthly voltage profiles from a large number of batteries in the data center, flagging deviations from the baseline as anomalies. Xue et al.\cite{xue2021fault} combined K-means, Z-score, and 3$\sigma$ screening, along with operating-state classification, feature matrices, and t-SNE reduction, to identify faulty battery packs of electric scooters. While such clustering-based methods show promise, they rely heavily on hyperparameter tuning (\textit{e.g.}, optimal $k$), which is challenging in unsupervised settings without labeled data. Beyond clustering, Zhang et al. \cite{zhang2023realistic} introduced a dynamical autoencoder for EV charging data, but evaluated the anomaly detection performance using \acrfull{ROC} and \acrfull{AUROC}, which can misrepresent accuracy in highly imbalanced datasets where anomalies are rare. Metrics such as \acrfull{AUPRC} or \acrfull{MCC} offer more robust evaluation in such cases.\cite{baldi2000assessing,matthews1975comparison,chicco2020advantages} Limited understanding of battery anomalies, compounded by scattered reports, means that real-life battery monitoring often still relies on manual inspection to identify outliers. While spotting a few anomalies in laboratory tests is straightforward, scaling this process to hundreds of cells over many cycles is labor-intensive and inefficient. Moreover, no benchmarking study has compared simple statistical approaches with more advanced \acrfull{ML} anomaly detection methods.

To facilitate systematic comparison and evaluation of different anomaly detection methods, we introduce \acrfull{OSBAD} as a comprehensive and open-source framework designed for detecting irregularities in electrochemical applications. In this study, we have integrated 15 distinct algorithms spanning a broad spectrum of methodological paradigms into \acrshort{OSBAD}, which include statistical, distance-based, and \acrshort{ML} approaches. The benchmark encompasses classical statistical methods such as the \acrfull{SD}, \acrfull{MAD}, \acrfull{IQR}, Z-score, and modified Z-score, which identify anomalies based on the distributional properties of the data. Complementing these methods are distance-based metrics such as Euclidean, Manhattan, Minkowski, and Mahalanobis distances, where distances between the observations in the feature space are used to distinguish outliers. To capture more complex and nonlinear patterns of degradation and failure, \acrshort{OSBAD} also incorporates advanced \acrshort{ML} algorithms such as \acrfull{IF}, \acrfull{KNN}, \acrfull{GMM}, \acrfull{LOF}, \acrfull{PCA}, and Autoencoder. By integrating these different methods into OSBAD, we provide a comprehensive framework for anomaly detection in battery data analytics.

To ensure generalization across a wide range of battery systems, we evaluated \acrshort{OSBAD} in this work using two independent datasets that differ fundamentally in their electrode materials and electrolyte compositions. The first datasets from \citet{Severson.2019} features a liquid-electrolyte cell with \ch{LiFePO4} as the positive electrode and graphite as the negative electrode, whereas the second datasets from Tohoku University in this work uses solid-state cells with \ch{LiNi_{0.5}Co_{0.2}Mn_{0.3}O2} as the positive electrode and In/InLi as the negative electrode. By benchmarking anomaly detection on datasets with distinct electrochemical and material properties, we demonstrate the cross-chemistry generalization of \acrshort{OSBAD} and lay the foundation for robust anomaly detection beyond battery applications in future work. We created binary labels in the benchmarking datasets, where 0 denotes inliers and 1 denotes outliers. By providing benchmarking datasets with open-source, reproducible anomaly detection workflows, OSBAD establishes a unified framework for anomaly detection towards the development of trustworthy and data-driven diagnostics for safety-critical energy systems.

In this manuscript, we use the experimental cell with the index \colorbox{Gainsboro}{2017-05-12\textunderscore 5\textunderscore 4C-70per\textunderscore 3C\textunderscore CH17} from the Severson datasets\cite{Severson.2019} and \colorbox{Gainsboro}{Cell-1} from the Tohoku datasets as the exemplary case study to demonstrate the evaluation of various anomaly detection methods. In the Severson dataset,\cite{Severson.2019} four anomalies were identified through visual inspection, which include cycles 0, 40, 147, and 148. Cycles 0 and 40 were found to be major anomalies affecting performance and safety, whereas cycles 147 and 148 are minor anomalies due to measurement gaps (see Figure \ref{fig: visual_check_anomalies_major} and Figure \ref{fig: visual_check_anomalies_minor} in the \acrshort{SI}). In the Tohoku dataset, cycles 79, 429, and 476 constitute the capacity fade anomalies. While major anomalies can significantly impact battery performance and safety, minor anomalies may distort downstream \acrshort{ML} predictions, particularly in models sensitive to feature completeness, where discontinuities can lead to biased learned patterns or alter distance metrics.

\begin{figure*}[ht!]
\centering
\includegraphics[width=\textwidth]{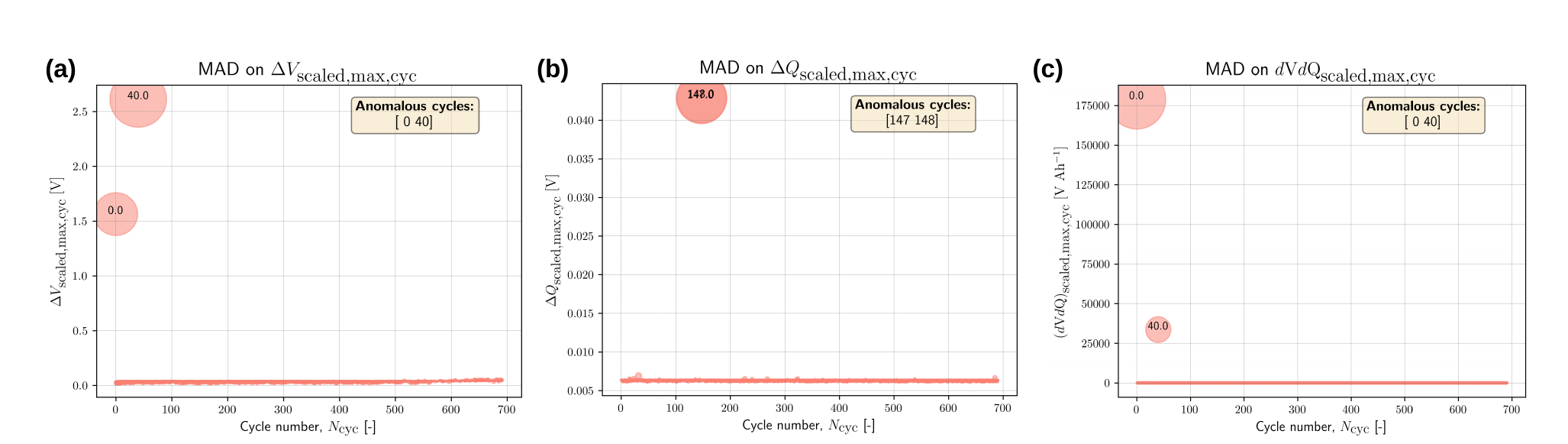}
\caption{\textbf{Univariate bubble plots of extracted features from the discharge voltage and capacity data in the Severson dataset.} The bubble chart of (a) maximum scaled voltage difference per cycle ($\Delta V_\textrm{scaled,max,cyc}$), (b) maximum scaled capacity difference per cycle ($\Delta Q_\textrm{scaled,max,cyc}$) and (c) maximum scaled ratio of voltage to capacity difference per cycle ($({dV}/{dQ})_\textrm{scaled,max,cyc}$) for \colorbox{Gainsboro}{Cell 2017-05-12\textunderscore 5\textunderscore 4C-70per\textunderscore 3C\textunderscore CH17}. Anomalous cycles can be identified as cycles with very high maximum voltage and capacity jump per cycle compared to normal cycles.}
\label{fig: bubble_chart}
\end{figure*}

\section{Feature Engineering for Anomalies Detection}
\label{sec: feature_engineer}

Anomalies can occur in various forms, including point anomalies, collective anomalies, local anomalies, and global anomalies (see Figure \ref{fig: anomaly definitions} in the \acrshort{SI}), where each anomalous pattern represents distinct deviations from normal cycling behavior. These anomalies may be univariate, where a single feature deviates from its typical range, or multivariate, where unusual combinations of features are anomalous. Because the datasets used in this benchmarking study contain both normal and abnormal measurements (see Figure \ref{fig: outlier_examples}), our study focuses on outlier detection. However, we use the terms “anomaly detection” and “outlier detection” interchangeably to refer to observations that deviate significantly from the main data distribution.

\subsection{Statistical Feature Transformation}
\label{sec: stats_feature_transformation}

As anomalies in the Severson datasets are convoluted within the normal cycling data,\cite{Severson.2019} statistical methods and \acrshort{ML} models cannot be effectively applied to the raw features. Figures \ref{fig: compare_feature_transformation_effects}(a1) and \ref{fig: compare_feature_transformation_effects}(b1) show examples of this feature entanglement, where the marginal histograms of abnormal voltage and capacity values overlap heavily with those from normal cycles. To help with the separation of abnormal cycles from normal cycles, we introduce a statistical feature transformation that uses the median and \acrshort{IQR} of the input features:

\begin{equation}
x_\textrm{scaled} = x_i - \left[\frac{\textrm{median}(X)^{2}}{\textrm{IQR}(X)}\right],
\label{eq: median_IQR_stats_feature_transformation}
\end{equation}

\noindent where the \acrshort{IQR} is computed as the difference between the third quartile ($75^\mathrm{th}$ percentile) and the first quartile ($25^\mathrm{th}$ percentile) of the input vector ($\textrm{IQR}(X) = Q_3(X) - Q_1(X)$). The term $\mathrm{median}(X)^2$ ensures that the transformed feature retains the original physical units.

Figure \ref{fig: explain_stats_feature_transformation} in the \acrshort{SI} illustrates how this transformation works: abnormal cycles often exhibit median and \acrshort{IQR} values that differ noticeably from those of normal cycles. By subtracting each data point from the ratio of the squared median to the \acrshort{IQR}, we amplify the difference between normal and anomalous cycles. Figures \ref{fig: compare_feature_transformation_effects}(a2) and \ref{fig: compare_feature_transformation_effects}(b2) show the marginal distributions of capacity and voltage after transformation, where the separation between normal and abnormal data clusters becomes much clearer. This disentanglement benefits downstream anomaly detection methods that rely on distinct clusters, densities, and distances in the feature space.

\begin{figure*}[ht!]
\centering
\includegraphics[width=\textwidth]{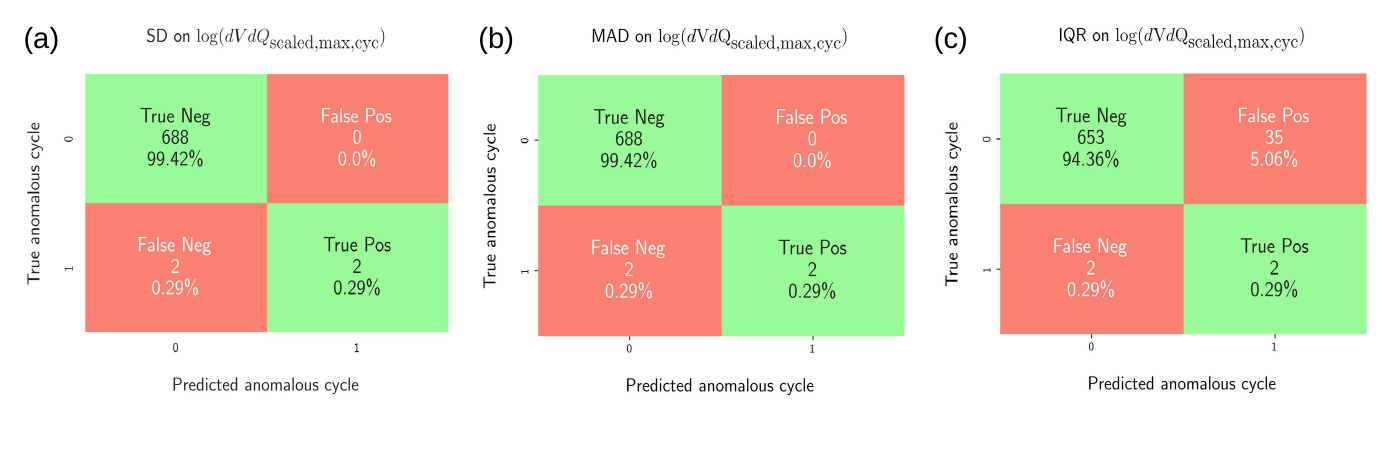}
\caption{\textbf{Confusion matrices for anomaly detection based on statistical methods applied to 
$\log({dV}/{dQ})_\textrm{scaled,max,cyc}$.} (a) \acrfull{SD}, (b) \acrfull{MAD} and (c) \acrfull{IQR}. The matrices compare predicted and true anomalous battery cycles from the benchmarking datasets, highlighting the classification performance of each statistical method. \acrfull{TP} means the specific statistical method correctly identifies an anomalous cycle, whereas \acrfull{TN} means a cycle is correctly predicted as normal. Both \acrshort{TP} and \acrshort{TN} are denoted by light green in the confusion matrix. The salmon color in the confusion matrix represents \acrfull{FP} and \acrfull{FN}, in which \acrfull{FP} indicates that the statistical method incorrectly identifies a normal cycle as anomalous (Type I error), and \acrfull{FN} means the method incorrectly predicts an anomalous cycle as normal (Type II error).}
\label{fig: stats_confusion_matrix_trim}
\end{figure*}

\subsection{Physics-informed Feature Extraction}
\label{sec: feature_extraction}
Because the anomalies in the Severson datasets arise from continuous sequences of abnormal voltage and capacity readings,\cite{Severson.2019} each affected cycle can be reduced from a collective anomaly to a cycle-level point anomaly. In other words, if a cycle has continuous abnormal voltage and capacity measurements, it will be labeled as an outlier.

Instead of using raw voltage and capacity data, we use scaled voltage and capacity from Section \ref{sec: stats_feature_transformation} for this step of feature extraction. Then, we calculate the scaled voltage difference $\Delta V_\textrm{scaled}$ and scaled capacity difference $\Delta Q_\textrm{scaled}$ for each measurement point in a cycle, enabling us to detect abrupt changes or discontinuities between consecutive readings. Finally, we extract point anomaly from collective anomalies by calculating the maximum of $\Delta V_\textrm{scaled}$ and $\Delta Q_\textrm{scaled}$ for each cycle, as well as their ratio as follows:

\begin{align}
\Delta V_\textrm{scaled,max,cyc} &= \underset{\textrm{cyc}}{\max}(V_{\textrm{scaled},{k+1}} - V_{\textrm{scaled},{k}}), \nonumber \\ 
\Delta Q_\textrm{scaled,max,cyc} &= \underset{\textrm{cyc}}{\max}(Q_{\textrm{scaled},{k+1}} - Q_{\textrm{scaled},{k}}),  \nonumber \\ 
\left[\frac{dV}{dQ}\right]_\textrm{scaled,max,cyc} &= \underset{\textrm{cyc}}{\max}\left[\frac{V_{\textrm{scaled},{k+1}} - V_{\textrm{scaled},{k}}}{Q_{\textrm{scaled},{k+1}} - Q_{\textrm{scaled},{k}}}\right], \\ \nonumber
\end{align} 

\nomenclature[L, 01]{$\Delta V_\textrm{scaled,max,cyc}$}{Maximum scaled voltage difference per cycle \nomunit{\si{V}}}
\nomenclature[L, 01]{$\Delta Q_\textrm{scaled,max,cyc}$}{Maximum scaled capacity difference per cycle \nomunit{\si{Ah}}}
\nomenclature[L, 02]{$({dV}/{dQ})_\textrm{scaled,max,cyc}$}{Maximum scaled ratio of voltage to capacity difference per cycle \nomunit{\si{V Ah^{-1}}}}

\noindent where $\Delta V_\textrm{scaled,max,cyc}$ is the maximum scaled voltage difference per cycle, $\Delta Q_\textrm{scaled,max,cyc}$ is the maximum scaled capacity difference per cycle and $({dV}/{dQ})_\textrm{scaled,max,cyc}$ is the maximum scaled voltage-to-capacity difference ratio per cycle, with $k$ indexing each measurement point.

Figure \ref{fig: bubble_chart} shows the bubble plot of the extracted cycle-wise point anomaly $\Delta V_\textrm{scaled,max,cyc}$, $\Delta Q_\textrm{scaled,max,cyc}$ and $({dV}/{dQ})_\textrm{scaled,max,cyc}$. Because anomalous cycles typically have a much higher maximum voltage or capacity difference compared to normal cycles, anomalies can be identified much faster after transformation. Although combining two features into a single metric may appear efficient, it often amplifies the anomaly scores  at the same time (see the y-axis range of $({dV}/{dQ})_\textrm{scaled,max,cyc}$ versus $\Delta V_\textrm{scaled,max,cyc}$ and $\Delta Q_\textrm{scaled,max,cyc}$). Extreme values from a few cycles can then overshadow smaller yet meaningful anomalies, making them invisible in the univariate bubble plot. For battery performance monitoring, missing a true anomaly (false negative) is much more detrimental than flagging a normal cycle (false positive), as a true anomaly could potentially pose safety hazards such as toxic chemical reactions or battery fire incidents.\cite{sun2020review} Therefore, to reduce the risk of missing a true anomaly, we retain all three features to safeguard against overlooking subtle but important anomalies, especially when benchmarking different univariate statistical anomaly detection methods.

\section{Statistical Anomaly Detection Methods}

Statistical methods can be used to detect extreme values of a univariate distribution, where extreme values denote the tail of the distribution. Due to the very low probability of the tail, the values in the distribution tail can be considered anomalies compared to the high-probability regions.

\begin{figure}[!ht]
\centering
\includegraphics[width=8.5cm]{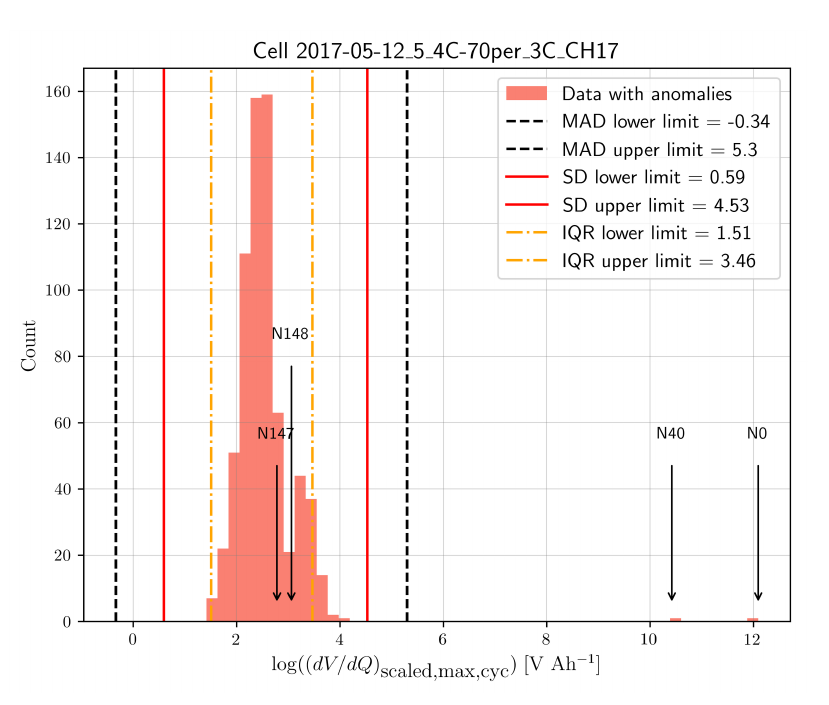}
\caption{\textbf{Comparing different statistical limits from \acrfull{MAD}, \acrfull{SD} and \acrfull{IQR} on the feature $\log({dV}/{dQ})_\textrm{scaled,max,cyc}$.} Due to the significant difference in the anomaly score between normal cycles and abnormal cycles for the feature $({dV}/{dQ})_\textrm{scaled,max,cyc}$ (see the bubble plot in Figure \ref{fig: bubble_chart}), natural logarithm is used in this comparison to minimize the variances. The \acrshort{MAD} detector has the widest limits, followed by \acrshort{SD} and \acrshort{IQR}. As the \acrshort{IQR} limits are most conservative, many normal cycles towards the distribution tail were flagged as outliers, leading to many \acrfull{FP} in the confusion matrix.}
\label{fig: compare_stats_limit}
\end{figure}

\begin{figure*}[ht!]
\centering
\includegraphics[width=\textwidth]{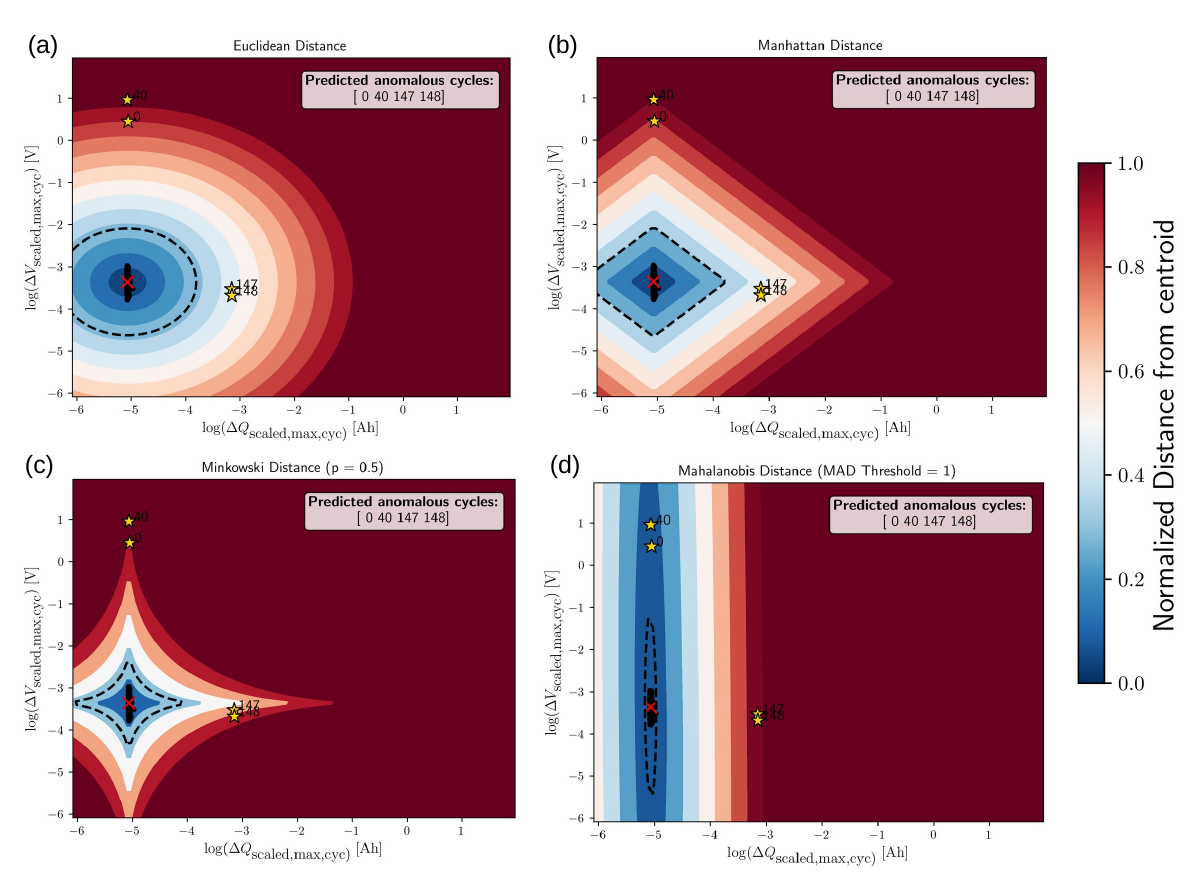}
\caption{\textbf{Visualization of distance-based anomaly detection methods applied to battery cycle data using two features: $\log(\Delta Q_\textrm{scaled,max,cyc})$ and $\log(\Delta V_\textrm{scaled,max,cyc})$.} Contour maps represent the normalized distance from the centroid in feature space for (a) Euclidean distance, (b) Manhattan distance, (c) Minkowski distance ($p=0.5$) and (d) Mahalanobis distance (with $\textrm{MAD threshold} = 1$). Predicted anomalous cycles are indicated by orange stars, while the centroid of normal cycles is marked by a red cross. The contour levels illustrate how each distance metric defines boundaries of normal and anomalous behavior in the battery cycle feature space.}
\label{fig: distance_contour_map_norm}
\end{figure*}

The confusion matrix in Figure \ref{fig: stats_confusion_matrix_trim} compares the performance of three statistical anomaly detection methods (\textit{i.e.} \acrshort{SD}, \acrshort{MAD} and \acrshort{IQR}) for detecting anomalous battery cycles using the feature $\log({dV}/{dQ})_\textrm{scaled,max,cyc}$. One should note that the statistical outlier detection methods rely on the statistical limits to distinguish normal from anomalous cycles. For example, the \acrshort{SD} method thresholds the anomaly score using the feature mean with three standard deviations ($\mu \pm 3\sigma$), whereas the \acrshort{MAD} method flags a cycle as anomalous if it exceeds the \acrshort{MAD} limits (see Equation \ref{eq: MAD} and \ref{eq: MAD-limit} in the \acrshort{SI}). The \acrshort{IQR} method detects anomalous cycles as points outside the lower and upper IQR limits, as shown by Equation \ref{eq: IQR} and \ref{eq: IQR_limits} in the \acrshort{SI}. Both \acrshort{SD} and \acrshort{MAD} show nearly identical, strong performance with very high true negative rates ($\approx 99.4\%$) and minimal false positives or false negatives, making them reliable for anomaly detection. In contrast, the \acrshort{IQR} detector produced the highest false positive rate (up to $\approx 5\%$) among the three methods. By comparing the statistical limits in Figure \ref{fig: compare_stats_limit}(a), two major anomalies (scores $12.09$ and $10.42$) were found to exceed all three statistical limits, producing true positives shown in the confusion matrix. In contrast, minor anomalies from cycle 147 ($2.78$) and cycle 148 ($3.06$), fell within the feature's distribution and limits, resulting in two false negatives. Figure \ref{fig: compare_stats_limit_IQR} in the \acrshort{SI} shows the statistical limits from the \acrshort{IQR} detector and illustrates how its conservative limits misclassified many normal cycles at the distribution tails as anomalies. The IQR range, commonly used in boxplots, flags data points outside the lower and upper whiskers ($IQR_\textrm{lower}$ and $IQR_\textrm{upper}$) as outliers. However, as shown in this benchmarking study, using a boxplot to identify anomalies may not always be reliable, as some normal observations towards the distribution tail may be wrongly classified as outliers. 

All statistical detectors missed some true anomalies (see Figure \ref{fig: stats_confusion_matrix_trim} and Figure \ref{fig: stats_confusion_matrix} in the \acrshort{SI}) because $\Delta V_\textrm{scaled,max,cyc}$ and $\Delta Q_\textrm{scaled,max,cyc}$ are univariate features measured along different dimensions, limiting cross-detection. Although the combined feature $({dV}/{dQ})_\textrm{scaled,max,cyc}$ merges the two dimensions and detects major anomalies, minor anomalies remain obscured within its distribution. As statistical approaches are usually implemented as univariate anomaly detection methods, they flag outliers in one variable at a time. Because they rely solely on the distribution of each feature, anomalies are often treated independently, without considering how different features may be related to one another. This comparison highlights the shortcomings of the univariate anomaly detection method compared to multivariate anomaly detection methods.

\begin{table*}[!h]
\caption{An overview of different distance metrics used for anomaly detection\cite{Jaroszewicz2014StatisticalMO, suárezdíaz2020tutorialdistancemetriclearning,
temple2023,
walterswilliams2010}}
\centering
% The number next to resizebox fit the table into page
% When setting to \textwidth, the table will be fitted into the A4 page
\resizebox{\textwidth}{!}{
\begin{tabularx}{23cm}{p{2.5cm}cccc} \\ \toprule

 & \makecell{Euclidean\\distance} & \makecell{Manhattan\\distance} & \makecell{Minkowski\\distance} & \makecell{Mahalanobis\\distance} \\ \midrule

% Euclidean distance header image
& \makecell{\raisebox{-0.5\totalheight}{\includegraphics[width=40mm ,height=38mm]{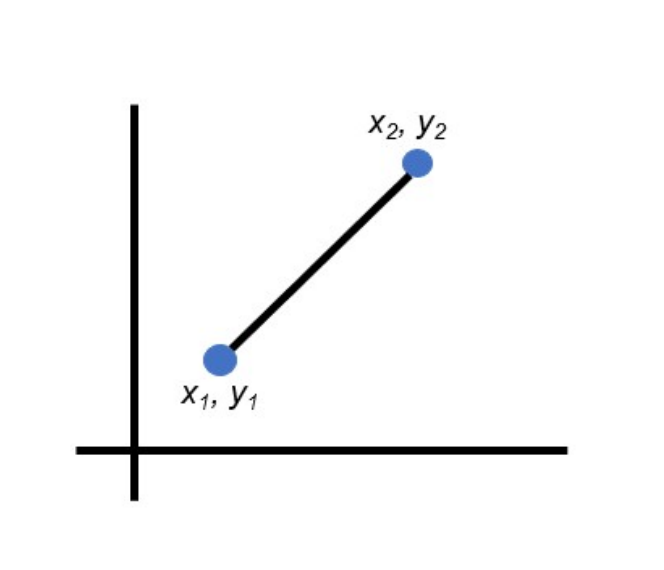}}}   

% Manhattan distance header image
& \makecell{\raisebox{-0.5\totalheight}{\includegraphics[width=40mm ,height=38mm]{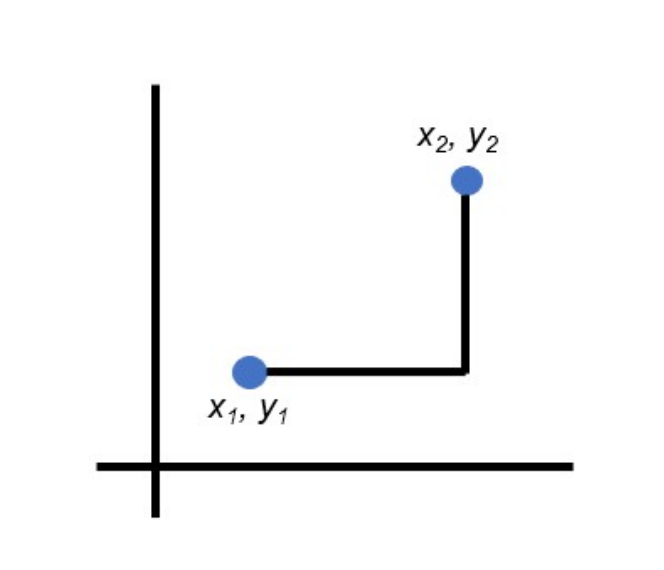}}} 

% Minkowski distance header image
& \makecell{\raisebox{-0.5\totalheight}{\includegraphics[width=40mm ,height=38mm]{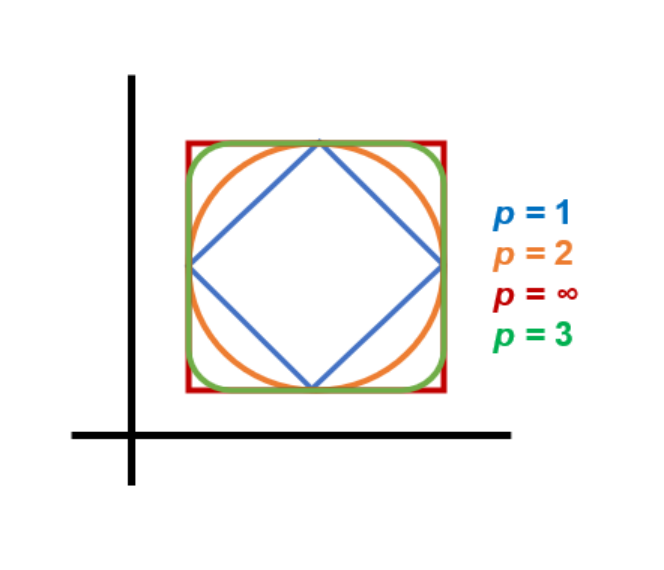}}} 

% Mahalanobis distance header image
& \makecell{\raisebox{-0.5\totalheight}{\includegraphics[width=40mm ,height=38mm]{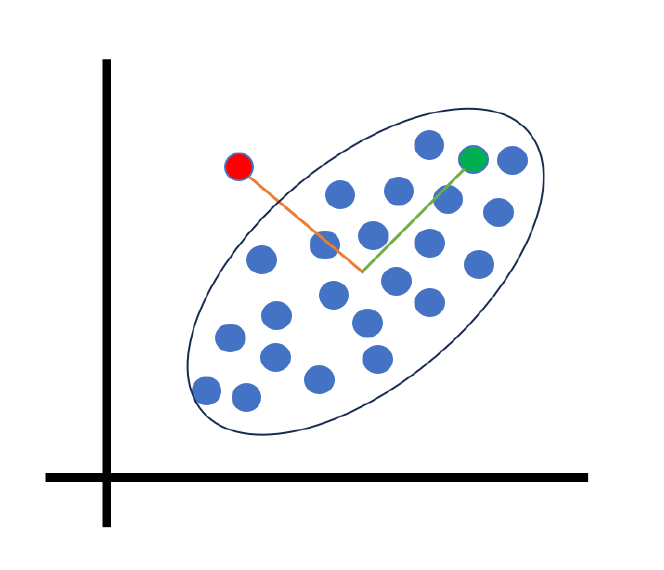}}} 

\\ \hdashline

\\ Equation

% Euclidean distance equation
& \makecell{$\sqrt{\sum_{i=1}^{n}(x_i-y_i)^2}$\\}

% Manhattan distance equation
& \makecell{$\sum_{i=1}^{n}|x_i-y_i|$\\}

% Minkowski distance equation
& \makecell{$\sqrt[p]{\sum_{i=1}^{n}|x_i-y_i|^p}$}

% Mahalanobis distance equation
& \makecell{$(x - \mu)^T S^{-1}(x - \mu)$}

\\ \\ \hdashline

Key characteristics 

% Euclidean distance descriptions
& \makecell*[{{p{4.5cm}}}]{\parbox{4.5cm}{\tabitem\;Straight-line distance between two points in the Euclidean space;\\
\tabitem\;Uses the L2 norm;\\
\tabitem\;Emphasizes larger deviations and assumes orthogonal feature space.}}

% Manhattan distance descriptions
& \makecell*[{{p{4.5cm}}}]{\parbox{4.5cm}{\tabitem\;Sum of absolute distances between two different data points;\\
\tabitem\;Also known as L1 norm, Taxicab distance, or city-block distance;\\
\tabitem\;Axis-aligned movement, mimicking movement along a grid;\\
\tabitem\;Each feature contributes linearly and independently to the total distance;\\
\tabitem\;Forms diamond-shaped contours.}}

% Minkowski distance descriptions
& \makecell*[{{p{4.5cm}}}]{\parbox{4.5cm}{\tabitem\;Generalized metric for measuring the distance between two points in a normed vector space;\\
\tabitem\;Amplifies larger differences in dimensions as $p$ increases;\\
\tabitem\;Includes Euclidean ($p=2$) and Manhattan ($p=1$) as special cases.\\}}

% Mahalanobis distance descriptions
& \makecell*[{{p{4.5cm}}}]{\parbox{4.5cm}{\tabitem\;Measures the distance between a point and a distribution.\\
\tabitem\;Distance that accounts for correlations and scale using the covariance matrix.\\
\tabitem\;Yields elliptical contours.}}

\\ \hdashline

Advantages 

% Euclidean distance advantages
& \makecell*[{{p{4.5cm}}}]{\parbox{4.5cm}{\tabitem\;Simple, computationally efficient, intuitive and widely used;\\
\tabitem\;Smooth and differentiable.}} 

% Manhattan distance advantages
& \makecell*[{{p{4.5cm}}}]{\parbox{4.5cm}{\tabitem\;Particularly effective in high-dimensional or sparse datasets;\\
\tabitem\;Robust to outliers;
\tabitem\;Computationally simpler and efficient (no squaring).}}

% Minkowski distance advantages
& \makecell*[{{p{4.5cm}}}]{\parbox{4.5cm}{\tabitem\;Unified frameworks with adaptability to different data geometries and different types of deviations via parameter $p$;\\
\tabitem\;Effective in high-dimensional spaces.}}

% Mahalanobis distance advantages
& \makecell*[{{p{4.5cm}}}]{\parbox{4.5cm}{\tabitem\;Accounts for feature correlation;\\
\tabitem\;Effective in multivariate outlier detection.}}

\\ \hdashline

Limitations 

% Euclidean distance limitations
& \makecell*[{{p{4.5cm}}}]{\parbox{4.5cm}{\tabitem\;Sensitive to scale and outliers ;\\
\tabitem\;Features must be normalized for meaningful results;\\
\tabitem\;Not ideal for high-dimensional data;\\
\tabitem\;Does not handle sparsity well.}} 

% Manhattan distance limitations
& \makecell*[{{p{4.5cm}}}]{\parbox{4.5cm}{\tabitem\;Like Euclidean distance, it still requires normalization or standardization;\\
\tabitem\;May not capture complex relationships.\\}}

% Minkowski distance limitations
& \makecell*[{{p{4.5cm}}}]{\parbox{4.5cm}{\tabitem\;Requires choosing $p$;\\
\tabitem\;The computational complexity increases with $p$.\\}}

% Mahalanobis distance limitations
& \makecell*[{{p{4.5cm}}}]{\parbox{4.5cm}{\tabitem\;Requires invertible covariance matrix;\\
\tabitem\;If matrix is singular or ill-conditioned, the distance cannot be computed reliably;\\
\tabitem\;Assumes data follows a multivariate normal distribution.}}

\\ \hdashline

Use-cases

% Euclidean distance use-cases
& \makecell*[{{p{4.5cm}}}]{\parbox{4.5cm}{\tabitem\;Distance-based outlier detection;\\
\tabitem\;Clustering-based anomaly detection algorithm;\\
\tabitem\;Autoencoder reconstruction error.}}

% Manhattan distance use-cases
& \makecell*[{{p{4.5cm}}}]{\parbox{4.5cm}{\tabitem\; \acrshort{KNN}, K-Medoids or DBSCAN clustering;\\
\tabitem\;Lasso Regression (uses L1 norm for regularization).}}

% Minkowski distance use-cases
& \makecell*[{{p{4.5cm}}}]{\parbox{4.5cm}{\tabitem\;Flexible \acrshort{KNN} outlier detection;\\
\tabitem\;Density-based algorithms (\textit{e.g.,} DBSCAN);\\ 
\tabitem\;Compute feature distances or correlations in high-dimensional spaces.}}

% Mahalanobis distance use-cases
& \makecell*[{{p{4.5cm}}}]{\parbox{4.5cm}{\tabitem\;Multivariate Gaussian-based outlier detection;\\
\tabitem\;Multivariate statistical analysis (measure similarity between multivariate observations);\\ 
\tabitem\;Elliptic envelope method.}} \\

\bottomrule
\end{tabularx}}
\label{table: distance_metrics_overview}
\end{table*}

\begin{figure*}[ht!]
\centering
\includegraphics[width=\textwidth]{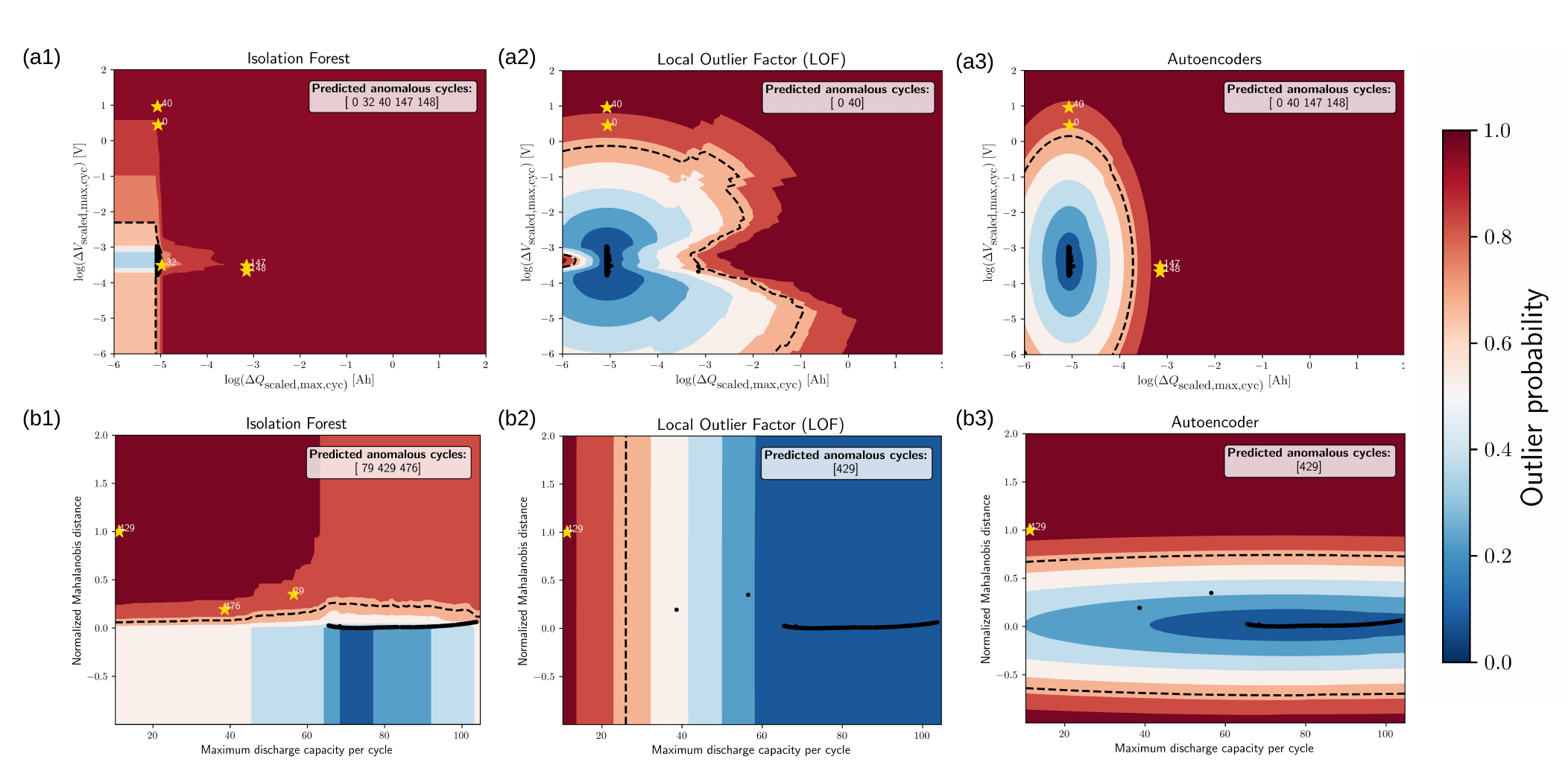}
\caption{\textbf{Probabilistic anomaly score maps to visualize predicted anomalies identified by different \acrshort{ML} algorithms.} Each subplot represents the anomaly detection decision boundaries for a given unsupervised ML model applied to the Severson datasets in Panel (a) and the Tohoku datasets in Panel (b). In the Severson datasets, the \acrshort{ML} algorithms were applied to the battery health features: $\log(\Delta Q_\textrm{scaled,max,cyc})$ and $\log(\Delta V_\textrm{scaled,max,cyc})$, whereas in the Tohoku datasets, the models were applied to normalized Mahalanobis distance and maximum discharge capacity per cycle. The color map encodes the outlier probability (dark blue = normal, dark red = highly anomalous), while the black dashed contour separates the predicted normal and anomalous regions. Yellow stars denote predicted anomalous cycles, with their cycle numbers labeled. Black scattered points indicate the predicted normal cycles. The true anomalous cycles labeled in the benchmarking dataset are cycles [0, 40, 147, and 148] for the Severson dataset and cycles [79, 429, 476] for the Tohoku dataset. For the models comparison in this figure, a consistent threshold of 0.7 is applied for all \acrshort{ML} methods. A cycle is predicted to be an outlier if its outlier probability score exceeds the threshold of 0.7.}
\label{fig: anomaly_score_map}
\end{figure*}

Among all the statistical anomaly detection methods (\textit{i.e.}, \acrshort{SD}, \acrshort{MAD}, \acrshort{IQR}, Z-score and modified Z-score), \acrshort{SD} is undoubtedly one of the easiest anomaly detection methods. Nevertheless, it is important to note that this method assumes that the data to follow a normal distribution. A probability plot\cite{wilk1968probability, chambers2018graphical} can be used to assess whether the extracted feature follows a normal distribution. If the points fall along a straight line, the feature distribution is approximately normal. Scattered tails indicate potential anomalies and curved patterns reveal skewness.\cite{wilk1968probability, chambers2018graphical} If the data is not normally distributed, data transformation such as Yeo-Johnson transformation can be used to reduce the skewness (see Figure \ref{fig: normality_check} in the \acrshort{SI}).\cite{yeo2000new} The same assumption applies to the Z-score method. However, unlike \acrshort{SD}, Z-score standardizes each feature to zero mean and unit standard deviation, producing thresholds between -3 and 3. This standardization enables easier comparison of the anomaly scores across different features, regardless of their original scales (see Figure \ref{fig: compare_zscore_sd} in the \acrshort{SI}). Compared to \acrshort{SD}, \acrshort{MAD} and Modified Z-score are more robust as these two detectors use median as the central point and measure how spread out the data points are depending on the median absolute deviation. However, for comparison with \acrshort{SD} under normal distributions, the \acrshort{MAD} score is typically multiplied by a constant factor $F_\textrm{MAD} = 1.4826$. \cite{rousseeuw1993alternatives,rosenmai2013using,leys2013detecting} We have also benchmarked the statistical prediction using different \acrshort{MAD}-factor and showed that the commonly used value ($F_\textrm{MAD} = 1.4826$) may lead to many false predictions if the underlying distribution is not Gaussian (see the comparison using different $F_\textrm{MAD}$ in Figure \ref{fig: stats_confusion_matrix_MAD} in the \acrshort{SI}). A more robust approach is to compute $F_\textrm{MAD}$ from the reciprocal of the 75th-percentile of the corresponding standard distribution (\textit{i.e.} a distribution with a mean of zero and a standard deviation of one).\cite{rosenmai2013using} Table \ref{table: stats_benchmark_outliers} in the \acrshort{SI} summarizes the characteristics as well as the advantages and limitations of each statistical anomaly detector.

\section{Distance-based Anomaly Detection}
\label{sec: distance_based_anomaly_detection}

To effectively quantify similarity between data samples, it is essential to define a suitable distance metric that computes pairwise dissimilarities within the feature space. This section benchmarks anomaly detection performance using four different distance metrics in the context of battery applications: Euclidean distance, Manhattan distance, Minkowski distance, and Mahalanobis distance. Table \ref{table: distance_metrics_overview} summarizes the key characteristics, advantages, and limitations of each distance metric. Although distance-based metrics are commonly integrated into various \acrshort{ML}-based anomaly detection frameworks, this section adopts a simpler centroid-based distance calculation. In scenarios where low latency, model interpretability, or limited computational resources are critical, such as in process-control hardware or embedded monitoring systems, a straightforward centroid-based method offers a more practical and efficient alternative to computationally intensive \acrshort{ML} algorithms.\cite{Samariya2023, Kloft:EECS-2010-22}

Figure \ref{fig: distance_contour_map_norm} shows the anomaly contour map of the four distance metrics. Each metric exhibits distinct classification behaviour due to its unique geometric interpretation of distance. For example, Euclidean distance in Figure \ref{fig: distance_contour_map_norm}(a) produces circular contours due to its isotropic variance assumption, whereas Manhattan distance in Figure \ref{fig: distance_contour_map_norm}(b) yields diamond-shaped contours, emphasizing axis-aligned deviations. Compared to Euclidean distance and Manhattan distance, Minkowski distance is a flexible distance metric depending on different $p$-values to measure the distance between two points in a normed vector space. Figure \ref{fig: distance_contour_map_minkowski} in the \acrshort{SI} illustrates how Minkowski distance contours change with varying $p$-values. As $p$ increases, the metric becomes more sensitive to large deviations, reducing the overall distance for smaller differences. While higher $p$-values ($p > 2$) increase the sensitivity towards extreme values, thus enhancing the separation in the red regions of the contour map, lower $p$-values ($p < 1$) reduce the influence of large deviations, resulting in more uniform distances and conservative anomaly detection. This means that aggressive anomaly detection may benefit from higher $p$-values, whereas lower $p$-values may be suitable for more conservative detection, focusing on subtle variations. Lastly, the Mahalanobis distance considers the covariance among variables to determine the multidimensional distance of a point to its distribution. This characteristic leads to the elliptical contours observed in Figure \ref{fig: distance_contour_map_norm}(d). Unlike Euclidean distance, which does not consider the correlation among features, the Mahalanobis distance metric excels in identifying anomalies that deviate from the joint distribution. However, it may underperform for anomalies aligned with the distribution (e.g., cycles 0 and 40), especially with a higher \acrshort{MAD} threshold of 3 (see Figure \ref{fig: distance_contour_map_mahalanobis} in the \acrshort{SI}).

Although distance metrics can be used to detect anomalies, this study demonstrates that they can also be employed in feature pre-processing steps. For example, in the Tohoku datasets, we computed normalized Mahalanobis distance from the cycle index and maximum discharge capacity per cycle and use this normalized distance metric as the input feature to \acrshort{ML}-based anomaly detection models (see Figure \ref{fig: anomaly_score_map}(b)).

\section{Data-driven Anomalies Detection}

Unlike univariate statistical anomaly detection method, where the anomalies of each feature are detected independently (see Figure \ref{fig: bubble_chart} and Figure \ref{fig: compare_stats_limit}), multivariate anomaly detection models can screen the anomalies of both features at the same time. For example, Figure \ref{fig: multivariate_anomalies} in the \acrshort{SI} shows a multivariate bubble plot where anomalies in both $\log(\Delta V_\textrm{scaled,max,cyc})$ and $\log(\Delta Q_\textrm{scaled,max,cyc})$ can be observed together. 

In this benchmarking study, we applied six unsupervised \acrshort{ML} models for multivariate anomaly detection: \acrfull{IF}, \acrfull{KNN}, \acrfull{GMM}, \acrfull{LOF}, \acrfull{PCA}, and Autoencoder. Table \ref{table: ml-models-comparison} summarizes these models, outlining their principles, important hyperparameters, main advantages, and limitations. These six approaches span diverse modeling paradigms, ranging from tree-based, distance-based, probabilistic, density-based, and linear projection to neural network representation, thereby providing a broad evaluation of unsupervised anomaly detection techniques.

\subsection{Probabilistic Anomaly Score Maps}
 
As different models produce different outlier scores, it is essential to note that these outlier scores are often not standardized, which makes them difficult to interpret directly.\cite{kriegel2011interpreting} As a result, benchmarking different anomaly detection methods in order to select the best model for battery applications is highly subjective. Even when working with the same datasets and the same anomalous cycles, each method often requires its own threshold.\cite{kriegel2011interpreting} Therefore, there is no single threshold that works across all the models in Table \ref{table: ml-models-comparison}. To overcome this inconsistency, we employ a probabilistic approach in this work to predict the probability that a cycle is an outlier, rather than making a binary prediction (\textit{i.e.}, whether a cycle is anomalous or not). The predicted outlier scores of each method are transformed into a range of [0, 1] using a min-max conversion.

Figure \ref{fig: anomaly_score_map} and Figure \ref{fig: anomaly_score_map_part2} in the \acrshort{SI} show the probabilistic anomaly contour plots of different methods using the Severson and Tohoku experimental cell datasets. \acrshort{IF} isolates anomalies by recursively splitting the feature space, which means that cycles that require fewer splits to be isolated are flagged as anomalous cycles. Due to the non-linear partitions typical of tree-based models, this process of recursive splitting produces an irregular decision boundary, as shown by Figure \ref{fig: anomaly_score_map}(a1) and Figure \ref{fig: anomaly_score_map}(b1) in both the Severson and Tohoku datasets. On the other hand, \acrshort{LOF} is a density-based model that compares the local density of a point with that of its neighbors. Therefore, its anomaly score map is also more intricate, producing irregular and density-based contours instead of smooth and symmetric shapes. Nevertheless, for the Tohoku datasets, only a partial section of the anomaly score map from the \acrshort{LOF} model is shown here to ensure a comparable axis-range across Figure \ref{fig: anomaly_score_map}(b1-b3). Figure \ref{fig: knn_lof_grid_contour_map}(a-b) in the \acrshort{SI} shows the complete anomaly score map, where a similar irregular and density-based contour as in Figure \ref{fig: anomaly_score_map}(a2) can be observed. Autoencoder, which detects anomalies through reconstruction errors from a neural network, produces a boundary that looks very similar to that of PCA (see Figure \ref{fig: anomaly_score_map_part2} in the \acrshort{SI}). The difference is that Autoencoder can also capture nonlinear patterns in the data, making them more flexible for complex cases.\cite{baldi2012autoencoders}

\subsection{Hyperparameter Tuning}

Hyperparameter tuning is the process of selecting the best combination of parameters for a \acrshort{ML} model to optimize the model performance. While hyperparameter tuning for supervised \acrshort{ML} is relatively straight-forward, this process becomes more challenging for unsupervised \acrshort{ML} due to the lack of labeled anomalies, class imbalance and sensitivity to different hyperparameters.\cite{zhao2022hyperparameter,fan2020hyperparameter,thomas2016learning} Proper hyperparameter tuning is important to ensure that the model accurately identifies anomalous cycles without overfitting to noise or underfitting to miss meaningful pattern. In this study, we implemented hyperparameter tuning for unsupervised anomaly detection models using two different methods:

\begin{enumerate}

\item Leveraging labeled outliers to train a model with hyperparameter tuning (training datasets) and subsequenty transfer the trained models with best average hyperparameters to predict anomalies on unlabeled test datasets\cite{zhao2022hyperparameter}

\item Using the output from the unsupervised model (\textit{e.g.}, predicted inliers from all six anomaly detection models) as features for a supervised task, and adjust the unsupervised model's hyperparameters to maximize the performance of the downstream supervised model.

\end{enumerate}

Instead of blindly trying random hyperparameters, as in random search, or computing expensive hyperparameter searches, as in grid search,\cite{bergstra2012random} we implemented Bayesian optimization to build a probabilistic model of the objective function and to efficiently sample the most promising hyperparameters for anomaly detection models.\cite{snoek2012practical,akiba2019optuna} Figure \ref{fig: hp_tuning_schema} in the \acrshort{SI} presents a concise overview of the hyperparameter tuning process.

\clearpage
% To position the table correctly within the paper margin in the landscape mode
% left or right adjustment
{\addtolength\textwidth{8cm}\addtolength\oddsidemargin{-2.5cm}
% ------------------------------------------------------
\begin{landscape}
\thispagestyle{empty}
\begin{table*}[ht!]
% \vspace{-12cm} moves the table upside down
\vspace{-8cm}
\centering
\caption{An overview of different \acrfull{ML} models used for anomaly detection}
\resizebox{21cm}{!}{\begin{threeparttable} % to use with adding table notes to the end of a table
\renewcommand{\arraystretch}{1.5} % Use this command to determine the spacing between each row; use before the begin{tabular} command
\begin{tabular*}{30cm}{lcccp{1.7cm}} \\ \toprule
ML Model & Key Concepts & Core Hyperparameters & Advantages and Limitations & Ref.  \\ \hline
% ---------------------------------------------------------
\makecell{\textbf{Tree-based ensemble method:} \\ \textbf{\acrfull{IF}} \\
\raisebox{-0.5\totalheight}{\includegraphics[width=40mm ,height=40mm]{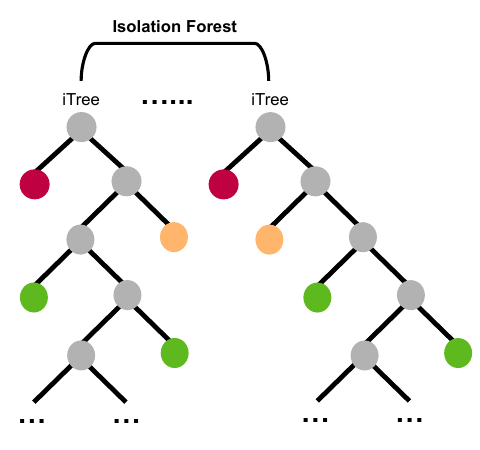}}}  

& \makecell*[{{p{6.75cm}}}]{\parbox{6.75cm}{\tabitem\;Random splitting of the feature space using multiple decision trees to isolate data points.\\
}
\parbox{6.75cm}{\tabitem\;\acrshort{IF} measures the number of splits (path length) required to isolate data points. Data points with short average path length from the root have high anomaly scores, whereas data points with more splits have low anomaly score.\\}
\parbox{6.75cm}{\tabitem\;The final anomaly score is calculated as the average normalized path length across all trees in the forest.}
} 

& \makecell*[{{p{6.75cm}}}]{
\parbox{6.75cm}{\textbf{n\textunderscore estimators}: The number of isolation trees constructed in the ensemble (default: 100). \\
}
\parbox{6.75cm}{\textbf{max\textunderscore samples}: The fraction of samples to draw from the datasets to train \acrshort{IF} algorithms (Range: $[0, 1]$).\\
}
\parbox{6.75cm}{\textbf{contamination}: The expected fraction of anomalies in the datasets to define the threshold for anomaly scores. Domain knowledge or exploratory analysis is required to define this hyperparameter (Range: $[0, 0.5]$).\\}
\parbox{6.75cm}{\textbf{max\textunderscore features}: The fraction of features to consider when training the model (Range: $[0, 1]$). \\
}
}

& \makecell*[{{p{6.75cm}}}]{
\textbf{Advantages} \\
\parbox{6.75cm}{\tabitem\;\acrshort{IF} does not assume specific statistical distribution about the feature, which makes it versatile across different datasets.\\
}
\parbox{6.75cm}{\tabitem\;\acrshort{IF} does not need distance or density metrics to calculate the anomaly scores.\\
\tabitem\;\acrshort{IF} has a linear time complexity with low memory requirements, making it scalable to large datasets and high-dimensional data.\\}
\textbf{Limitations} \\
\parbox{6.75cm}{\tabitem\;\acrshort{IF} only performs best when the anomalous data points are sparse and distinct from normal data clusters.\\
\tabitem\;All features are treated independently and only one feature per isolation operation is considered - may not work well on data with non-linear separable features space.}
}

& \citen{liu2008isolation, xu2023deep, zhao2019pyod, regaya2021point}

\\ \hdashline
% ---------------------------------------------------------
\makecell{\textbf{Distance-based method:} \\ \textbf{\acrfull{KNN}} \\
\raisebox{-0.5\totalheight}{\includegraphics[width=40mm ,height=40mm]{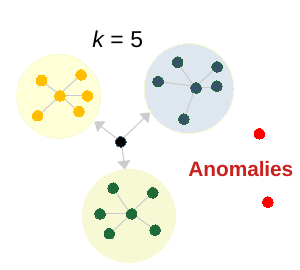}}}  

& \makecell*[{{p{6.75cm}}}]{\parbox{6.75cm}{\tabitem\;\acrshort{KNN} identifies a new data point to its \textit{k}-th closest data points by calculating the distance between the data point and its \textit{k}-th nearest neighbors.\\
}
\parbox{6.75cm}{\tabitem\;The anomaly score of a data point is determined by the distance to its \textit{k}-th nearest neighbors. A longer distance from the main data cluster implies that the data point is more likely to be an outlier.\\}
} 

& \makecell*[{{p{6.75cm}}}]{
\parbox{6.75cm}{\textbf{n\textunderscore neighbors}: Determines the number of nearest data points considered when making a prediction. \\
}
\parbox{6.75cm}{\textbf{method}: Determines whether the distance to the \textit{k}-th neighbors should be the largest, mean or median distance.\\
}
\parbox{6.75cm}{\textbf{metric}: Evaluates how the distance between data points is calculated. Commonly used distance metrics include Euclidean distance, Manhattan distance and Minkowski distance.\\}
}

& \makecell*[{{p{6.75cm}}}]{
\textbf{Advantages} \\
\parbox{6.75cm}{\tabitem\;It is easy to implement \acrshort{KNN} compared to other \acrshort{ML} models.\\
}
\textbf{Limitations} \\
\parbox{6.75cm}{\tabitem\;Because \acrshort{KNN} calculates the distance between the query point and all data points, the calculation can be computationally expensive, especially with large and high-dimensional datasets.\\}
\parbox{6.75cm}{\tabitem\;Finding an optimal \textit{k}-value is non-trivial: a small \textit{k}-value could cause \acrshort{KNN} to overfit (\textit{i.e.} capturing noises instead of finding the clustering patterns), whereas a large \textit{k}-value would overly smooth the decision boundary and causes the model to underfit.}
}

& \citen{steinbach2009knn,guo2003knn, zhang2017learning,zhang2021challenges, zhao2019pyod}

\\ \hdashline
% ---------------------------------------------------------
\makecell{\textbf{Probabilistic method:} \\ \textbf{\acrfull{GMM}} \\
\raisebox{-0.5\totalheight}{\includegraphics[width=40mm ,height=30mm]{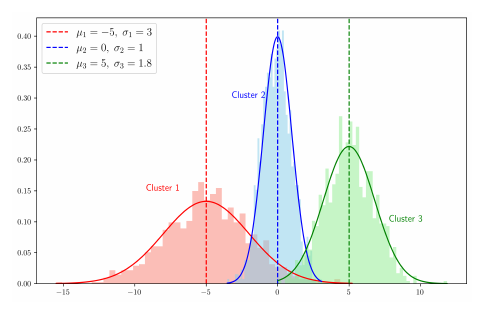}}}  

& \makecell*[{{p{6.75cm}}}]{\parbox{6.75cm}{\tabitem\;\acrshort{GMM} considers the input feature space to be generated from a mixture of different Gaussian distributions, where each distribution represents a different cluster.\\
}
\parbox{6.75cm}{\tabitem\;\acrshort{GMM} is a soft-clustering \acrshort{ML} model that assigns probabilities to each data point for belonging to different clusters. Overlapping data point assignments are possible depending on their probabilities.\\}
\parbox{6.75cm}{\tabitem\;\acrshort{GMM} uses the Expectation-Maximization algorithm to iteratively estimate the parameters that best fit the data.\\}
\parbox{6.75cm}{\tabitem\;Data points that have low probabilities of belonging to any of the Gaussian distributions are considered outliers.}
} 

& \makecell*[{{p{6.75cm}}}]{
\parbox{6.75cm}{\textbf{n\textunderscore components}: The number of Gaussian distributions (clusters) in the mixture (default: 1). \\
}
\parbox{6.75cm}{\textbf{covariance\textunderscore type}: Determines the shape of each Gaussian distribution. \textit{full}: each component has its own general covariance matrix; \textit{tied}: All components share the same general covariance matrix. \textit{diag}: Each component has its own diagonal covariance matrix. \textit{spherical}: Each component has its own single variance.\\
}
\parbox{6.75cm}{\textbf{contamination}: The expected fraction of anomalies in the datasets. Domain knowledge or exploratory analysis is required to define this hyperparameter (Range: $[0, 0.5]$).\\}
\parbox{6.75cm}{\textbf{init\textunderscore params}: Defines how initial parameters are defined, using either K-means clustering based initialization method or random initialization.\\
}
}

& \makecell*[{{p{6.75cm}}}]{
\textbf{Advantages} \\
\parbox{6.75cm}{\tabitem\;\acrshort{GMM} can represent a wide range of feature distributions by combining multiple Gaussian components, making them suitable for modeling complex multimodal distributions.\\
}
\parbox{6.75cm}{\tabitem\;\acrshort{GMM} implements soft clustering and offers probabilistic cluster assignments.\\}
\textbf{Limitations} \\
\parbox{6.75cm}{\tabitem\;Incorrect estimation of the component number may lead to overfitting or underfitting of the model.\\}
\parbox{6.75cm}{\tabitem\;\acrshort{GMM} assumes each cluster to follow a Gaussian distribution, which may not hold true for all datasets.}
}

& \citen{xuan2001algorithms, reynolds2009gaussian, mclachlan2014number, zhao2019pyod}

\\ 
\bottomrule
\end{tabular*}
%\begin{tablenotes}
%\item[$\dag$ The definition of these metrics, acronyms and symbols can be found in the Appendix.]
%\item[* Measurement at room temperature]
%\item[** The CCD of fluoroethylene carbonate (FEC)-based electrolyte solution is $\SI{2}{mA\;cm^{-2}}$ (1100 cycles).\cite{Markevich.2017}]
%\item[*** The shear modulus of Li metal is $\SI{4.25}{GPa}$.\cite{Seungho.2016}]
%\item[**** Calculated in this work]
%\end{tablenotes}
\end{threeparttable}} % to use with adding table notes to the end of a table
\label{table: ml-models-comparison}
\end{table*}
%\fillandplacepagenumber
\end{landscape}}

\clearpage
% To position the table correctly within the paper margin in the landscape mode
% left or right adjustment
{\addtolength\textwidth{8cm}\addtolength\oddsidemargin{-2.5cm}
% ------------------------------------------------------
\begin{landscape}
\thispagestyle{empty}
\begin{table*}[ht!]
% \vspace{-12cm} moves the table upside down
\vspace{-4cm}
\centering
\resizebox{21cm}{!}{\begin{threeparttable} % to use with adding table notes to the end of a table
\renewcommand{\arraystretch}{1.5} % Use this command to determine the spacing between each row; use before the begin{tabular} command
\begin{tabular*}{30cm}{lcccp{1.7cm}} \\ \toprule
ML Model & Key Concepts & Hyperparameters & Advantages and Limitations & Ref.  \\ \hline
% ---------------------------------------------------------
\makecell{\textbf{Density-based method:} \\ \textbf{\acrfull{LOF}} \\
\raisebox{-0.5\totalheight}{\includegraphics[width=40mm ,height=35mm]{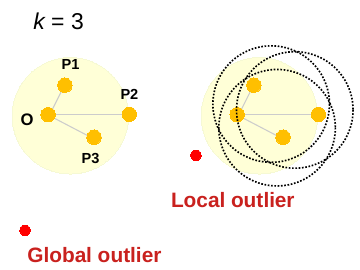}}}  

& \makecell*[{{p{6.75cm}}}]{\parbox{6.75cm}{\tabitem\;\acrshort{LOF} identifies anomalies by comparing the density of a data point relative to the densities of its neighbors.\\
}
\parbox{6.75cm}{\tabitem\;
\acrfull{RD}, which is the distance from each data point to its maximum distance of k-value data points is calculated to determine the perimeter area for each data point. \acrfull{LRD} is then calculated to determine the distance ratio for every nearest neighbor inside the cluster.\\}
\parbox{6.75cm}{\tabitem\;\acrshort{LOF} is the average \acrshort{LRD} of the all neighboring data points divided by the \acrshort{LRD} of point O. The \acrshort{LOF} of data points further away from their neighbors are usually greater than 1 and are therefore considered as outliers. Data points within densely packed areas are considered as normal and have $LOF(O)\leq1$.}
} 

& \makecell*[{{p{6.75cm}}}]{
\parbox{6.75cm}{\textbf{n\textunderscore neighbors}: Determines the number of neighboring data points to compute the local density. \\
}
\parbox{6.75cm}{\textbf{metric}: The distance metric (\textit{e.g.} Euclidean distance, Manhattan distance and Minkowski distance) used to compute distances between points and measure local densities.\\}
}

& \makecell*[{{p{6.75cm}}}]{
\textbf{Advantages} \\
\parbox{6.75cm}{\tabitem\;\acrshort{LOF} can be used to detect global outliers and local outliers if they are far away from the main clusters of data points.\\
}
\textbf{Limitations} \\
\parbox{6.75cm}{\tabitem\;\acrshort{LOF} is not suitable to identify collective anomalies that have high local densities, which will be considered as normal using the \acrshort{LOF} algorithms.}
} 

& \citen{zhao2019pyod,alghushairy2020review, cheng2019outlier, xu2022outlier}

\\ \hdashline
% ---------------------------------------------------------
\makecell{\textbf{Linear projection method:} \\ \textbf{\acrfull{PCA}} \\
\raisebox{-0.5\totalheight}{\includegraphics[width=40mm ,height=40mm]{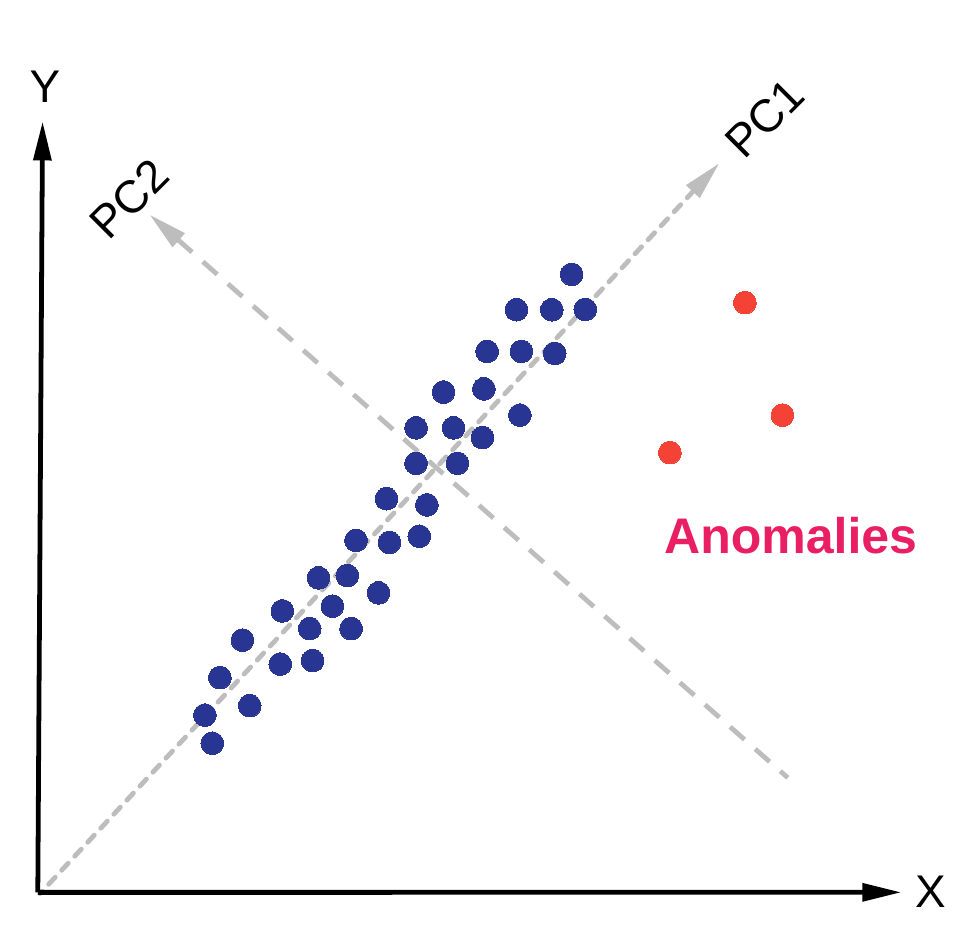}}}  

& \makecell*[{{p{6.75cm}}}]{\parbox{6.75cm}{\tabitem\;By calculating the linear combination of the input variables, \acrshort{PCA} is a method used to reduce the dimensionality of the datasets to \acrfull{PCs} while preserving its maximum variance.
}
\parbox{6.75cm}{\tabitem\;The 1$^{\textrm{st}}$ component (PC1) is found by finding the best line through the datasets that best describes the input data. Square distances of all points to this line are minimized. \acrshort{PCA} algorithms then continue to find the next line perpendicular to PC1 (\textit{i.e.,} PC2) to capture the remaining variance.}
\parbox{6.75cm}{\tabitem\;PC1 has the highest variance, followed by PC2 and so on. Anomalies are data points that deviate from the identified components.}
} 

& \makecell*[{{p{6.75cm}}}]{
\parbox{6.75cm}{\textbf{n\textunderscore components}: Number of components to keep.\\
}}

& \makecell*[{{p{6.75cm}}}]{
\textbf{Advantages} \\
\parbox{6.75cm}{\tabitem\;\acrshort{PCA} can be used to extract representative features from large datasets based on the linear correlation of the input variables with the \acrshort{PCs};\\
\tabitem\;When the input data is transformed onto the lower-dimensional hyperplane of the principal components, anomalies can be identified easily when their eigenvectors have smaller eigenvalues compared to the eigenvalues of principal components.\\
\tabitem\;\acrshort{PCA} is computationally efficient and easy to implement, which make it useful for large datasets with many features.\\
}
\textbf{Limitations} \\
\parbox{6.75cm}{\tabitem\;Features normalization is required as the variables with large standard deviations will dominate the determination of the principal components;\\
\tabitem\;\acrshort{PCA} may not work well if the input features have highly non-linear relationships.}}

&

\citen{zhao2019pyod,abdi2010principal,
daffertshofer2004pca,
karamizadeh2013overview,ringberg2007sensitivity} 

\\ \hdashline
% ---------------------------------------------------------
\makecell{\textbf{Neural-network based method:} \\ \textbf{Autoencoder} \\
\raisebox{-0.5\totalheight}{\includegraphics[width=40mm ,height=40mm]{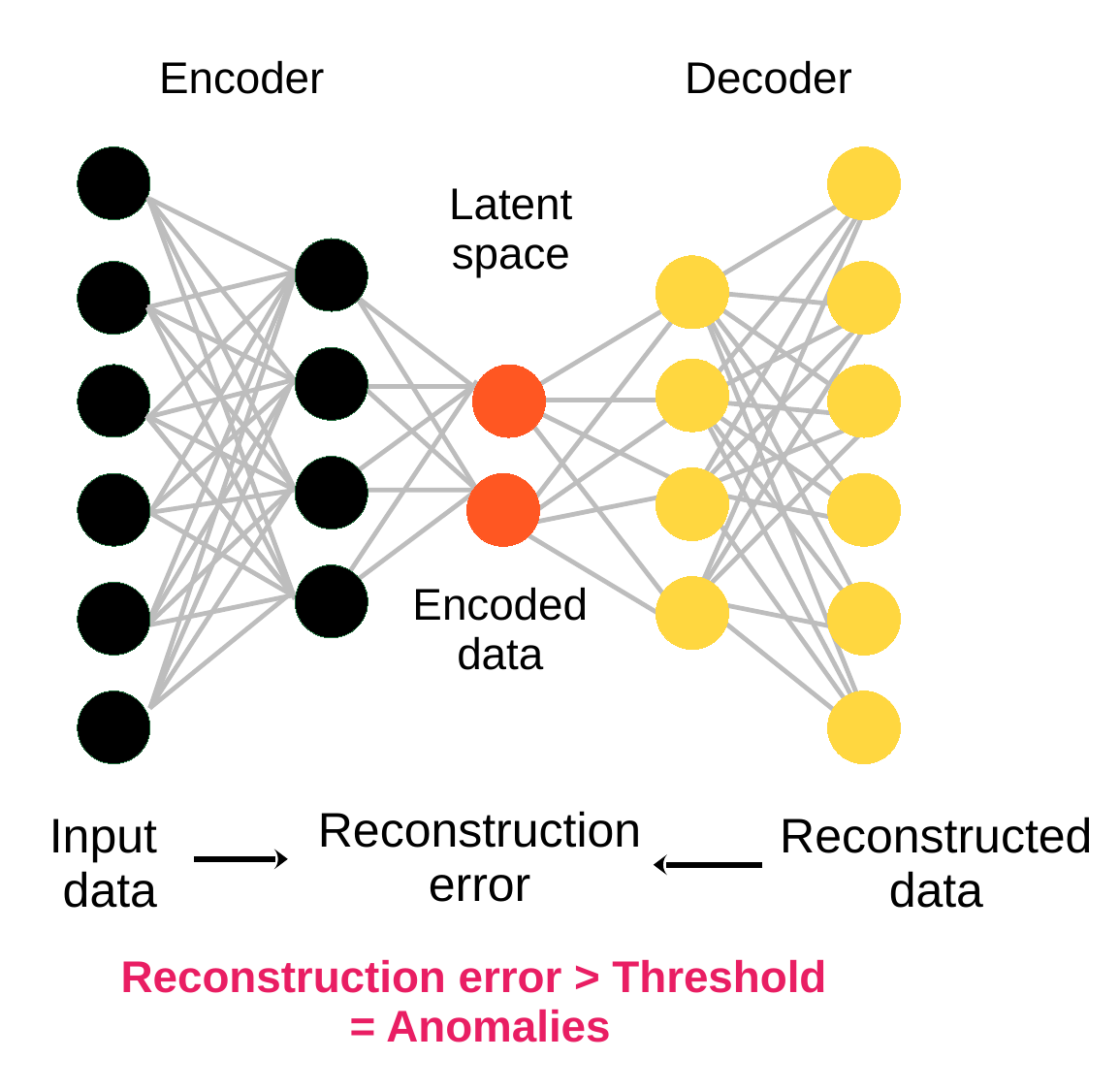}}}  

& \makecell*[{{p{6.75cm}}}]{\parbox{6.75cm}{\tabitem\;Autoencoder is a neural network trained to reconstruct the input data through a compressed latent representation;\\
\tabitem\;Autoencoder consists of an encoder, which compresses the input data into a latent space, and a decoder that reconstructs the output from the latent space;\\
\tabitem\;In anomaly detection, autoencoder learns to reconstruct the outputs from the dominant structure of the normal observations. Reconstruction errors (\textit{i.e.,} mean-squared errors) will be calculated after training;\\
\tabitem\;Due to the divergence from the norm, anomalous observations will have high reconstruction errors and can thus be flagged by autoencoder.\\
}} 

& \makecell*[{{p{6.75cm}}}]{
\parbox{6.75cm}{\textbf{epoch\textunderscore num}: The number of epoch used for training;\\
}
\parbox{6.75cm}{\textbf{batch\textunderscore size}: The batch size for training;\\}
\parbox{6.75cm}{\textbf{dropout\textunderscore rate}: The dropout to be used across all layers for regularizations;\\}
\parbox{6.75cm}{\textbf{hidden\textunderscore neuron\textunderscore list}: The number of neurons per hidden layers;\\}
\parbox{6.75cm}{\textbf{hidden\textunderscore activation\textunderscore name}: The activation function used in hidden layers;\\}
\parbox{6.75cm}{\textbf{optimizer\textunderscore name}: The name of the optimizer used to train the models.
}
}

& \makecell*[{{p{6.75cm}}}]{
\textbf{Advantages} \\
\parbox{6.75cm}{\tabitem\;Autoencoder is more flexible than \acrshort{PCA} if the input features have non-linear relationships.\\
}
\textbf{Limitations} \\
\parbox{6.75cm}{
\tabitem\;Clean data without anomalies is needed for training an autoencoder. For practical applications where the input data is often contaminated with anomalies, autoencoder will only be effective if the anomalies are rare and do not dominate the patterns of the normal observations;\\
\tabitem\;Training autoencoders requires longer computational time.}}

& \citen{zhao2019pyod,Bank2023, baldi2012autoencoders,berahmand2024autoencoders, michelucci2022introduction}

\\ 
\bottomrule
\end{tabular*}
%\begin{tablenotes}
%\item[$\dag$ The definition of these metrics, acronyms and symbols can be found in the Appendix.]
%\item[* Measurement at room temperature]
%\item[** The CCD of fluoroethylene carbonate (FEC)-based electrolyte solution is $\SI{2}{mA\;cm^{-2}}$ (1100 cycles).\cite{Markevich.2017}]
%\item[*** The shear modulus of Li metal is $\SI{4.25}{GPa}$.\cite{Seungho.2016}]
%\item[**** Calculated in this work]
%\end{tablenotes}
\end{threeparttable}} % to use with adding table notes to the end of a table
\end{table*}
%\fillandplacepagenumber
\end{landscape}}
\clearpage

\begin{figure*}[ht!]
\centering
\includegraphics[width=\textwidth]{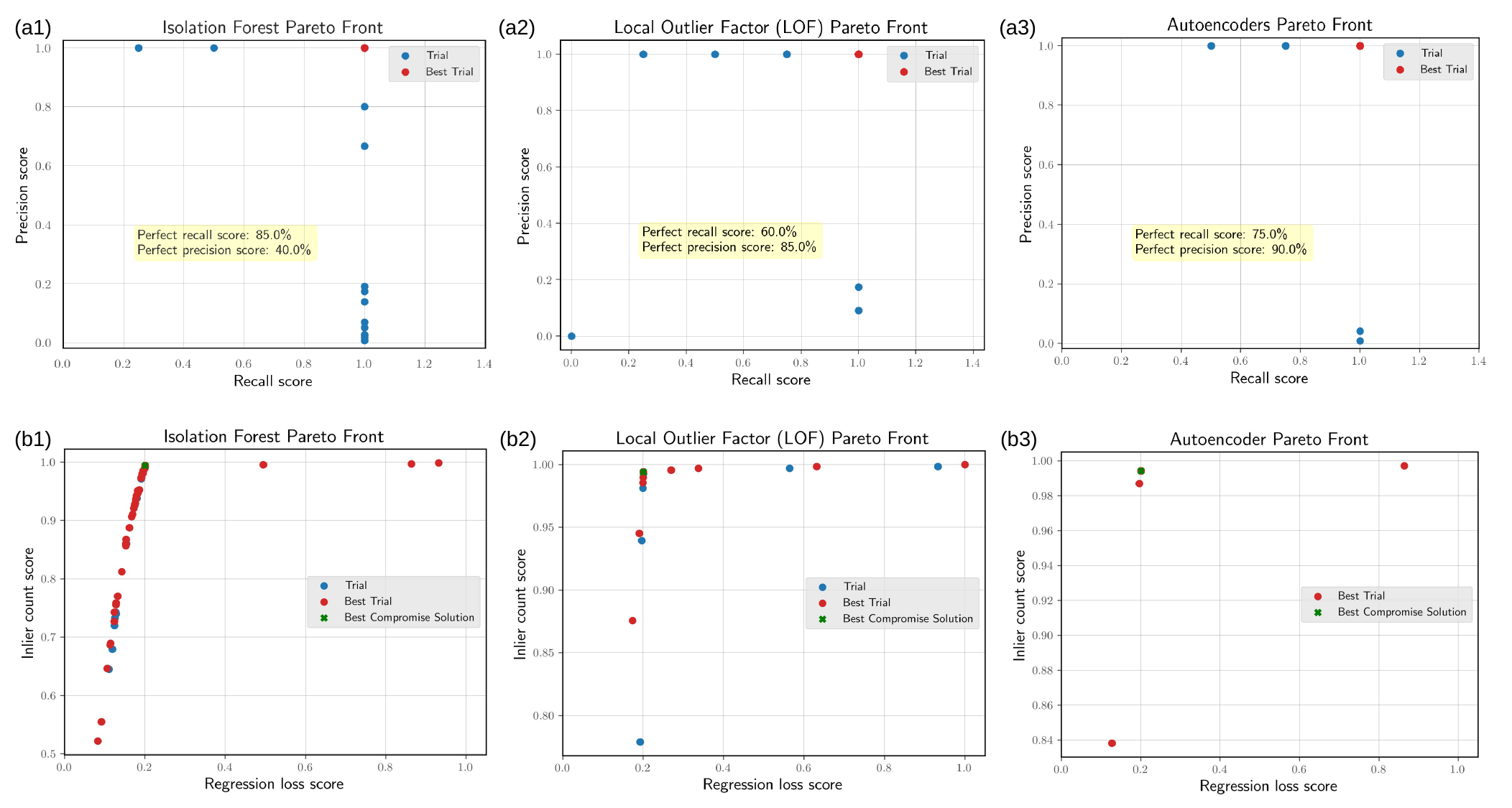}
\caption{\textbf{Bayesian hyperparameter optimization for classification and regression objectives across multiple models.} The top panel (a1-a3) illustrates the Pareto fronts obtained from Bayesian optimization performed to maximize the precision and recall scores for Isolation Forest, Local Outlier Factor (LOF), and Autoencoder models, respectively. The X-axis is the recall score (fraction of actual anomalies detected), whereas the Y-axis is the precision score (fraction of predicted anomalies that are correct). While the blue scattered points represent all the trials evaluated during the optimization process, the red dots denote the best trials, which achieved the maximum recall and precision scores. The percentage of perfect recall or precision scores in the yellow-annotated box denotes the percentage of trials with the highest value in that metric among the Pareto front. The bottom panel (b1-b3) displays the Bayesian optimization results, which minimize the regression loss and maximize the inlier count score for the same models. The best and compromise solutions are highlighted to demonstrate trade-offs between competing objectives during hyperparameter tuning.}
\label{fig: bayesopt_pareto_front}
\end{figure*}

\subsubsection{Hyperparameters Tuning with Transfer Learning}

\label{sec: hp_tuning_with_transfer_learning}

For hyperparameter tuning based on transfer learning, the hyperparameters are optimized using labeled anomalies during training for all 23 training cell datasets. For each model-cell combination, the best hyperparameters are recorded, then averaged across all 23 training cells. These averaged hyperparameters are subsequently fixed after tuning and transferred to evaluate the model performance using the unlabeled test datasets, therefore ensuring a consistent and unbiased performance evaluation. The details on the train-test split in this work can be found in the \acrshort{SI} (Section \ref{app: dataset}). The Tree-structured Parzen Estimator Algorithm is used to tune hyperparameters by jointly optimizing two internal performance metrics: recall score and precision score.\cite{akiba2019optuna} A Pareto front is constructed to visualize the trade-offs between optimizing the recall score and precision score (see Figure \ref{fig: bayesopt_pareto_front}). 20 trials were defined to optimize the hyperparameter tuning of all six models. By using Bayesian optimization, the hyperparameter search space for each model can be explored and exploited more efficiently, reducing trial-and-error, and providing interpretable trade-off solutions via Pareto fronts.

Figure \ref{fig: bayesopt_pareto_front}(a1)-(a3) shows the resulting Pareto fronts, where recall and precision were optimized simultaneously. \acrshort{IF} achieved the highest perfect recall rate (recall = 1 in \SI{85}{\%} of 20 trials). However, \acrshort{IF} reached perfect precision in only \SI{40}{\%} of the trials, reflecting a higher false-positive rate. Autoencoder achieved balanced recall and precision at Pareto-optimal points, making it preferable when both false positives and false negatives are costly. By contrast, \acrshort{LOF} prioritized precision over recall, reliably detecting true outliers but missing some anomalies.

\subsubsection{Hyperparameter Tuning with Regression Proxies}

In the absence of a labeled training datasets for the anomaly detection task, hyperparameter optimization techniques based on transfer learning are not directly applicable. As a result, redefining the objective function is needed as the conventional outlier detection performance metrics based on the precision and recall score become non-pertinent during the tuning process. Thus, a proxy multivariate regression model is introduced to facilitate the estimation of model performance using substitute indicators, such as regression loss, which can serve as viable alternatives in the absence of ground-truth labels. The core idea is that if the anomaly detection model successfully identifies the outliers, the remaining inlier data should, in principle, exhibit a more consistent structure, leading to enhanced regression loss. Conversely, if the model fails to detect anomalies due to suboptimal hyperparameters, the regression loss will deteriorate. In other words, this means that the selected model configuration cannot effectively separate outlier cycles from inliers. The regression loss thus serves as a proxy metric for assessing the efficacy of the selected hyperparameter in the absence of ground truth labels. However, if a model is very sensitive with a perfect recall score and low precision (\textit{i.e.}, no false negatives and high false positives), the regression loss may still decrease. In fact, this reduction in loss can become more pronounced as the number of false positives increases. To address this challenge, we introduce a second evaluation metric known as the inlier count score. As the name implies, this metric quantifies the number of cycles classified as inliers by the model configuration. It ensures that the reduction in regression loss does not come at the expense of excessive false positives. A high inlier count score is essential for preserving the integrity of the data distribution from normal cycles, especially in highly imbalanced datasets. Thus, the two contradicting metrics transform the hyperparameter tuning task into a multi-objective optimization or Pareto Optimization problem, where the goal becomes finding the best trade-off between minimizing the regression loss while maximizing the inlier count.

Figure \ref{fig: bayesopt_pareto_front}(b1)-(b3) illustrates the Pareto fronts generated for three different anomaly detection models. A sharp decline in the loss score is observed as the anomalies are progressively removed, followed by an inflection point, indicating the successful exclusion of true positives. Beyond this point, the rate of loss reduction slows, indicating that the model begins removing inliers (false positives), which still lowers the loss score but less dramatically. This geometric behavior is especially evident when utilizing the proposed statistical feature transformation framework, combined with the mean squared error as the regression loss function. Unlike the transfer-learning method, this optimization process yields a set of multiple Pareto-optimal solutions. Each of these solutions is mathematically plausible and thus equally valid from a multi-objective perspective. However, for practical implementation, it is often necessary to identify a single compromise solution that best balances the competing objectives. More detailed explanation of this selection procedure can be found in the \acrshort{SI} (Section \ref{app: hyperparameters_tuning}). The optimal hyperparameters identified through this process were then used to evaluate each model against the true labels using standard anomaly detection metrics for a post hoc evaluation and comparison.

\begin{figure*}[ht!]
\centering
\includegraphics[width=\textwidth]{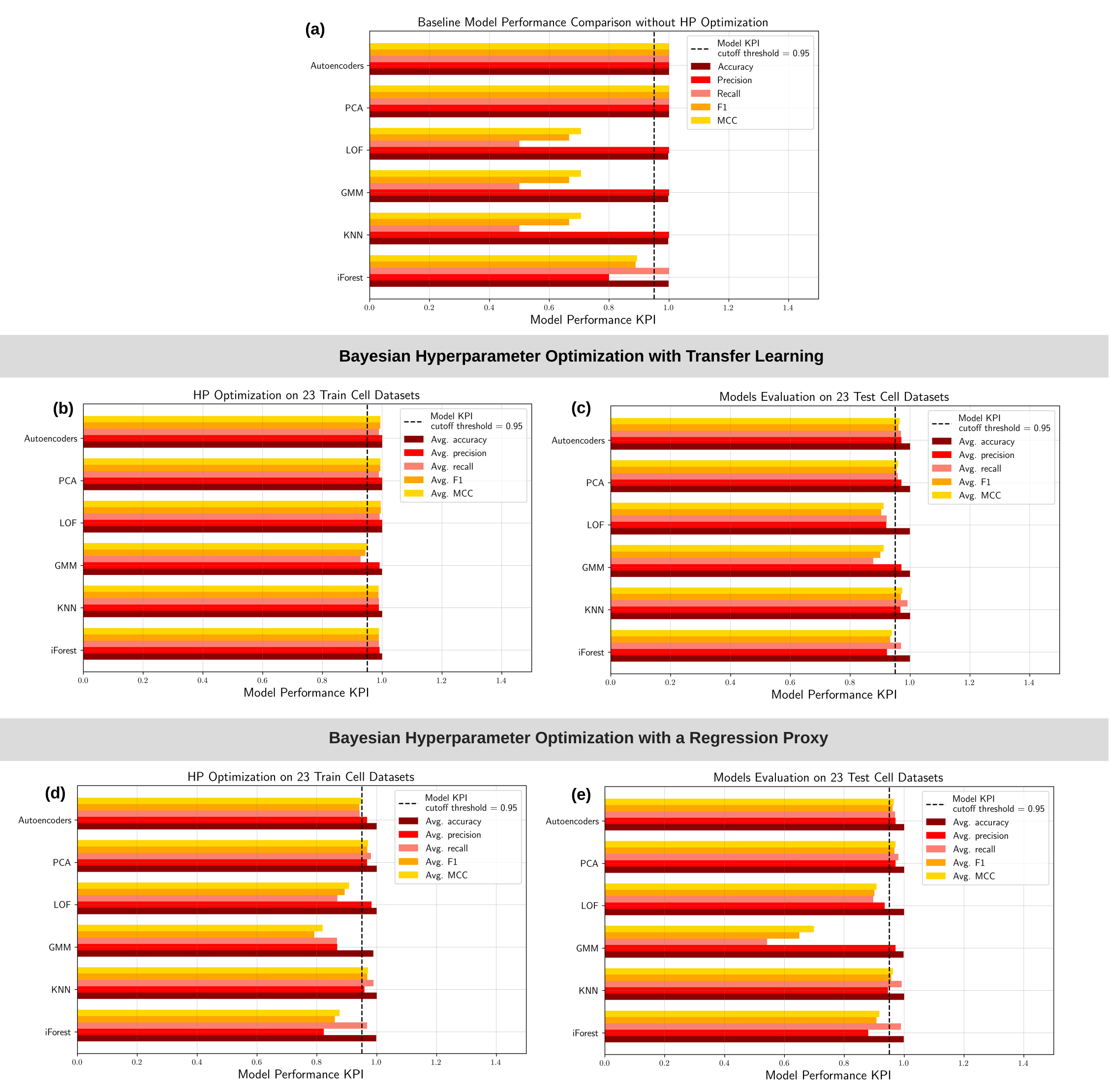}
\caption{\textbf{Comparison of baseline and Bayesian hyperparameter optimization strategies across multiple cell experimental datasets for both training and testing phases in the Severson datasets.} Panel (a) shows the baseline model performance without hyperparameter (HP) optimization across six \acrshort{ML} models: Autoencoder, \acrshort{PCA}, \acrshort{LOF}, \acrshort{GMM}, \acrshort{KNN}, and \acrshort{IF}. Panels (b)-(c) present results obtained from Bayesian hyperparameter optimization with transfer learning, where (b) shows optimization on 23 training cell datasets and (c) shows evaluation on 23 test cell datasets. Panels (d)-(e) depict results from Bayesian hyperparameter optimization using a regression proxy, with (d) representing optimization on the training datasets and (e) the corresponding evaluation on the test datasets. The color-coded bars indicate the performance metrics, which include accuracy, precision, recall, F1-score, and Matthew Correlation Coefficient (MCC). The KPI cutoff threshold of 0.95 serves as a benchmark for comparing model performance across different setups and optimization strategies.}
\label{fig: ml_models_eval}
\end{figure*}

\subsection{Model Evaluation Metrics}

The performance of anomaly detection methods are evaluated in this work based on five different metrics: (1) accuracy, (2) precision, (3) recall, (4) F1-score and (5) \acrfull{MCC}. The definitions and detailed descriptions of these metrics can be found in the \acrshort{SI} (Section \ref{app: model_evaluation_metrics}). While accuracy measures the proportion of correct predictions among all predictions, it is essential to note that this metric can be misleading in imbalanced datasets.\cite{chicco2020advantages} An imbalanced dataset occurs when one class significantly outnumbers the other class,\cite{chicco2020advantages} such as \SI{99.4}{\%} of the True Negative class (normal cycles) versus \SI{0.6}{\%} of the True Positive class (anomalous cycles). As a result, models trained on an imbalanced dataset tend to favor the majority class, reporting high accuracy, but exhibit poor performance on the minority class.\cite{chicco2020advantages} To properly assess the performance of each anomaly detection method, more informative metrics such as precision, recall, F1-score, and \acrshort{MCC} should be used.\cite{chicco2020advantages} While the metric precision penalizes false positive predictions, the metric recall penalizes false negative predictions. In other words, a model that produces no false positives has a precision of 1, whereas a model that predicts no false negatives has a recall of 1. The evaluation metrics of both precision and recall scores are often in tension, meaning that improving the precision score usually reduces the recall score, and vice versa. For anomaly detection in battery applications, predicting false negatives has a much higher cost than predicting false positives. This is because missing true anomalies in real-time battery operations could lead to severe consequences, such as toxic chemical reactions or even more severe battery fire incidents. Therefore, prioritizing the improvement of the recall score to reduce the probability of missing true anomalies is more critical for safe battery operations compared to addressing false alarms later.

Figure \ref{fig: ml_models_eval} shows the comparison of baseline and Bayesian hyperparameter optimization strategies across multiple cell experimental datasets for both training and testing phases in the Severson datasets. All models are first trained on the dataset from a single cell without hyperparameter tuning to evaluate the baseline performance of different anomaly detection models. This step is crucial for evaluating model performance without hyperparameter tuning, especially when true labels are scarce in real-world applications. As illustrated by Figure \ref{fig: ml_models_eval}(a), many models were shown to underperform, with several metrics well below the \acrfull{KPI} threshold of 0.95 (dashed vertical line). \acrshort{LOF}, \acrshort{GMM}, and \acrshort{KNN} show the largest gap from the targetted \acrshort{KPI}. \acrshort{IF} shows comparatively stronger performance (its recall score is above the model \acrshort{KPI} threshold) but is still not perfect. In the absence of hyperparameter tuning, Autoencoder and \acrshort{PCA} were found to have the best performance across all evaluation metrics. Figure \ref{fig: ml_models_eval}(b)-(c) presents results obtained from Bayesian hyperparameter optimization with transfer learning, where Figure \ref{fig: ml_models_eval}(b) shows the optimization on 23 training cell datasets and Figure \ref{fig: ml_models_eval}(c) shows the evaluation on the remaining 23 test cell datasets. To prevent overfitting, hyperparameter tuning was repeated independently for each of the 23 training cells. The best average parameters were then transferred to evaluate the model performance using the unlabeled test datasets. During training, \acrshort{IF}, \acrshort{KNN}, \acrshort{LOF}, PCA, and Autoencoder achieved near-perfect accuracy, precision, recall, F1, and \acrshort{MCC} scores. On the test set, \acrshort{KNN}, \acrshort{LOF}, PCA, and Autoencoder maintained high performance, while \acrshort{GMM} continued to underperform, suggesting inherent model limitations. Notably, \acrshort{IF} consistently sustained high recall across four experimental setups, despite a lower average precision score on the test set, highlighting its value in battery safety monitoring, where detecting all true anomalies outweighs the cost of false positives.

Figure \ref{fig: ml_models_eval}(d)-(e) presents the benchmarking results obtained using the proxy evaluation method. While the models demonstrate notable improvement in KPIs compared to the baseline, they still underperform compared to the transfer learning approach, as expected. Nevertheless, the effectiveness of the proxy evaluation approach becomes particularly evident in lower-performing models, such as \acrshort{LOF} and \acrshort{GMM}, where substantial performance gains are observed after hyperparameter tuning. In particular, the \acrshort{KNN} model exhibits the highest improvement following hyperparameter tuning. Furthermore, the results suggest that \acrshort{PCA} and Autoencoder emerge as the most robust and suitable candidates for deployment in truly unsupervised environments.

\section{Conclusions}

In real-world applications, the performance and reliability of \acrshort{ML} models are often hindered by poor data quality issues and undetected anomalies. In the context of battery systems, missing true anomalies is not only detrimental to downstream model accuracy but can also cause costly downtime and serious safety hazards such as thermal runaway and fire incidents. To address these challenges, we introduce \acrshort{OSBAD} as a systematic and open-source framework in this work for detecting anomalies across different battery applications. By integrating 15 diverse algorithms encompassing statistical, distance-based, and unsupervised machine-learning methods into OSBAD, we enable systematic anomaly detection benchmark across heterogeneous datasets.

While statistical anomaly detection methods based on the mean (\acrshort{SD} and Z-score) are simple to implement and useful when the data distributions are approximately Gaussian, their sensitivity towards extreme outliers could lead to missed anomalies, especially when subtle irregularities could be masked by extreme outliers in the datasets. In contrast, median-based statistical methods such as \acrshort{MAD}, \acrshort{IQR} and modified Z-score are more resilient towards skewed distributions. Nevertheless, these methods also have their limitations. \acrshort{MAD} and modified Z-score may yield inaccurate thresholds when applied with standard \acrshort{MAD} scaling factor, whereas \acrshort{IQR} may cause a high false positive prediction due to misclassifying normal observations at the distribution tails as anomalies. On the other hand, distance-based anomaly detection methods such as Euclidean, Manhattan and Minkowski distance compute distance between data points in a given feature space. Here, feature correlation in high-dimensional space is especially useful for multivariate anomaly detection. This approach also proves to be more pragmatic for low-latency, interpretable, and resource-constrained environments than computationally intensive \acrshort{ML} algorithms. Among all the unsupervised \acrshort{ML} methods examined in this work, \acrshort{IF} achieved a consistent high recall score for Severson datasets, which is especially important in safety-critical applications where missing anomalies can lead to catastrophic failures. \acrshort{PCA} and Autoencoder provided a more balanced trade-off between the recall and precision, helping to reduce false alarms. A key contribution of \acrshort{OSBAD} is its chemistry-agnostic design, which enables generalization across different battery systems. Through validation on two distinct datasets featuring both liquid and solid-state chemistries, we demonstrate the capability of \acrshort{OSBAD} to handle heterogeneous data and detect irregularities across different electrochemical systems with different input features. Its open-source implementation helps creating a valuable community resource towards advancing trustworthy data-driven diagnostics and promoting safer deployment of ML models in real-world energy applications.

In addition to its comprehensive algorithmic coverage, we also integrate a systematic hyperparameter tuning pipeline based on Bayesian optimization into \acrshort{OSBAD}, addressing a major bottleneck in unsupervised anomaly detection. Here, we implemented two complementary hyperparameter tuning methods: (1) transfer-learning-based tuning, where optimal hyperparameters derived from labeled data are transferred to unlabeled datasets, and (2) regression-proxy-based tuning, which adjusts hyperparameters to optimize the performance of downstream supervised tasks. Compared to baseline models without hyperparameter tuning, we show that models with hyperparameter tuning perform significantly better especially when some labels are available to guide the hyperparameter selection. By integrating a broad range of anomaly detection methods with advanced hyperparameter optimization strategies and validation across different chemistries, we provide a systematic benchmarking framework for robust and reproducible anomaly detection in battery data analytics. 

\section*{Data availability}
In this work, we utilize DuckDB, a state-of-the-art, fast, and open-source database system, to store all benchmarking datasets with labeled outliers. The database can be found here: \url{https://github.com/meichinpang/osbad/tree/master/database}

\section*{Code availability}
Code reproducing all the analysis and figures in this work is available in the following GitHub repository:

\noindent \url{https://github.com/meichinpang/osbad} \newline

\noindent The documentation with examples to install and run \acrshort{OSBAD} can be found here:
\noindent \url{https://osbad.readthedocs.io/en/latest/index.html}

\section*{Acknowledgements}
Part of this study is funded by the JSPS Summer Fellowship 2024 for Mei-Chin Pang during the academic research at Tohoku University in Japan. 

\bibliography{mybib}
\bibliographystyle{rsc}% This command sorts the bibs according to the appearance in the text.

\newpage
\onecolumn

\begin{appendices}
\renewcommand\theequation{\thesection\arabic{equation}}

\setcounter{equation}{0}

\renewcommand\thetable{\thesection\arabic{table}}
\setcounter{table}{0}
\renewcommand\thefigure{\thesection\arabic{figure}}
\setcounter{figure}{0}

\section{Benchmarking Dataset Descriptions}
\label{app: dataset}

Table \ref{table: benchmark_dataset} summarizes the details of two datasets (MIT/Stanford and Tohoku) used for benchmarking anomaly detection in battery systems. The MIT/Stanford data was published by \citet{Severson.2019}, whereas the Tohoku datasets is introduced in the present study.

The Severson datasets features \acrfull{LFP} as the positive electrode, graphite as the negative electrode with a liquid electrolyte. Here, we selected 46 cells out of the 124 cells to benchmark various statistical and \acrshort{ML} models. Each cell contains an average of 845 cycles, which is sufficient for evaluating different anomaly detection methods. The cells were cycled between \SI{2.0}{V} and \SI{3.6}{V}, where \SI{1}{C} is equivalent to \SI{1.1}{A}. While the cells were charged with different C-rates to 100\% \acrshort{SOC}, they were discharged at \SI{4}{C} to \SI{2.0}{V} with a current cut-off at C/50. Because the discharge C-rate is more uniform compared to the charge C-rate, we focus on benchmarking anomaly detection using the discharge section in this work. After manually labeling each cycle across all 46 cells as either an inlier (label 0) or an outlier (label 1), the datasets are split into two equal parts: (1) a training set with 23 cells, and (2) a test set with the remaining 23 cells. The labeled test set is kept completely separate until the final evaluation to prevent data leakage and inflated performance estimates. The training set serves two purposes: to assess baseline model performance without hyperparameter tuning and to optimize each model's hyperparameters through Bayesian optimization.

In contrast, the Tohoku datasets use \acrfull{NMC} as the positive electrode, indium/lithium–indium (In/InLi) as the negative electrode, and a solid electrolyte (\ch{Li6PS5Cl}), with 10 cells cycled between \SI{3.0}{V} and \SI{4.3}{V} vs. $\textrm{Li}^{+}/\textrm{Li}$ at \SI{0.1}{C} and \SI{25}{\celsius}. Out of these 10 cells, we selected datasets from four cells for training (Cells 1, 2, 5, and 6) and another four cells for testing (Cells 7, 8, 9, and 10) for anomaly detection algorithms. Both datasets span a 100\% state-of-charge ($\Delta \textrm{SOC}$) range and include anomalous discharge-capacity profiles.

\begin{table}[!ht]
\caption{Datasets selected for benchmarking anomaly detection}
\centering
% The number next to resizebox fit the table into page
\resizebox{8.5cm}{!}{\begin{threeparttable}
% Use this command to determine the spacing between each row;
% use before the begin{tabular} command
%\setlength\belowcaptionskip{-20pt}
\renewcommand{\arraystretch}{1.7}
% reduce the number next to \begin{tabular*} will increase the table size
% Increase the number next to \begin{tabular*} will reduce the table size
% This number should be slightly above the number in \resizebox{9cm} 
\begin{tabular*}{9.8cm}{p{3.4cm}cc} \\ \toprule

 & MIT/Stanford & Tohoku   \\ \midrule

Positive electrode & \acrshort{LFP} 	  & \makecell{\acrshort{NMC}523 \\ \ch{(LiNi_{0.5}Co_{0.2}Mn_{0.3}O2})}  \\
Electrolyte & \makecell{Liquid electrolyte} & \makecell{Solid electrolyte \\(\ch{Li6PS5Cl})} \\
Negative electrode & Graphite & In/InLi \\
Number of cells & 46 cells & 10 cells \\
Nominal capacity & \makecell{\SI{1.1}{Ah}} & \makecell{\SI{100}{mAh/g}} \\
Discharging C-rates & \makecell{\SI{4}{C}\\ ($\SI{1}{C} \approx \SI{1.1}{A}$)}  & \makecell{\SI{0.1}{C}\\ ($\SI{1}{C} \approx \SI{233}{\micro A}$)} \\
$\Delta$SOC in \% & 100\% & 100\% \\
Voltage limits &  \makecell{\SI{2.0}{V} - \SI{3.6}{V}}  & \makecell{\SI{3.0}{V} - \SI{4.3}{V} \\vs. Li$^{+}$/Li}\\
Operating temperature & \SI{30}{\celsius} & \SI{25}{\celsius} \\
Anomalous data type & \makecell{Discharge-\\capacity\\profile} & \makecell{Discharge-\\capacity\\profile} \\
Data source & \citet{Severson.2019} & This paper \\
\bottomrule
\end{tabular*}
%\begin{tablenotes}
%\item[${\dag}$ The list of acronyms in this proposal can be found in the Appendix.]
%\end{tablenotes}
\end{threeparttable}}
\label{table: benchmark_dataset}
\end{table}

\newpage

\section{Anomalies in other battery cycling dataset}
\label{app: other_cycling_anomalies}

\begin{figure}[!ht]
\centering
\includegraphics[width=10cm,trim=4 4 4 4,clip]{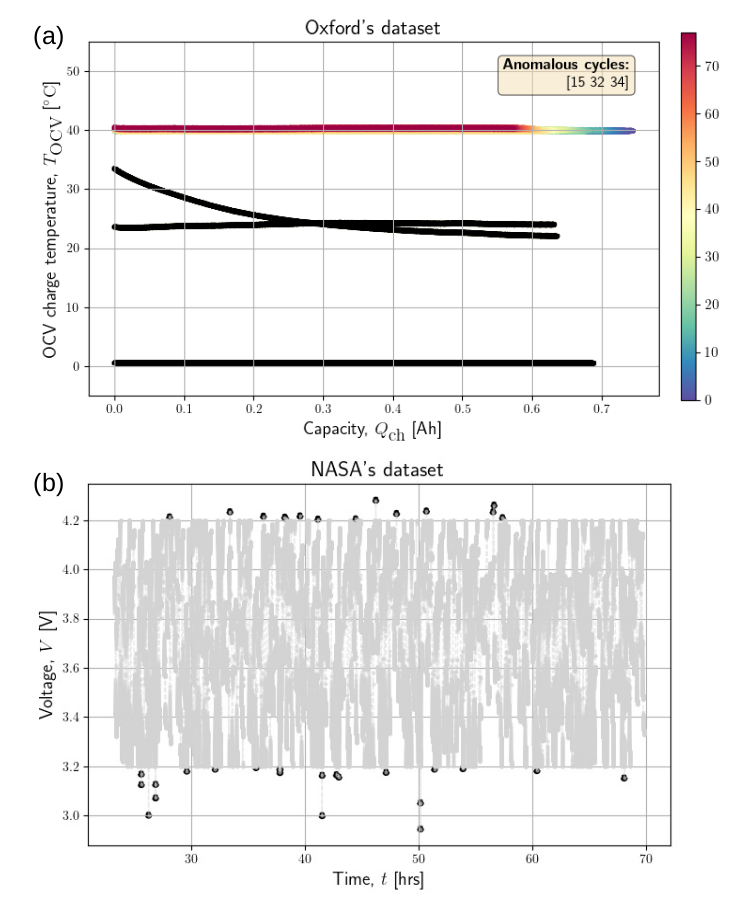}
\caption{\textbf{Anomalies observed in the battery cycling datasets across different cell chemistries.} (a) In the Oxford battery degradation datasets (NCA-graphite cell chemistry), the solid black curves denote anomalous cycles during the OCV charge temperature measurement. In a normal cycle, the OCV temperature in the thermal chamber should remain constant at \SI{40}{^\circ C}. However, some temperature measurements were found to deviate from the operating temperature in the thermal chamber at \SI{40}{^\circ C} for cycle 15, 32 and 34.\cite{birkl2017diagnosis} (b) In the dynamic cycling profile published by NASA (LCO-graphite cell chemistry), the black dots represent the voltage measurements that exceed the specified voltage operating range between \SI{3.2}{V} and \SI{4.2}{V}.\cite{Bole_dataset.2014,bole2014adaptation}}
\label{fig: other_anomalies}
\end{figure}

As shown in Figure \ref{fig: other_anomalies}, anomalies due to equipment breakdown, measurement errors or even cell failure are not limited to certain battery chemistries such as \acrshort{LFP}-graphite cells or \acrshort{NMC}622-\ch{Li_6PS_5CI}-indium solid-state battery cell chemistries,\cite{Severson.2019, puls2024benchmarking} but could also occur in other chemistries. For example, in the Oxford battery degradation datasets (NCA-graphite cell chemistry),\cite{birkl2017diagnosis} the cells were all tested inside the thermal chamber at \SI{40}{^\circ C}. Nevertheless, some outliers could be detected in the temperature profile during the \acrshort{OCV} period, where some measurements deviate from the operating temperature in the thermal chamber at \SI{40}{^\circ C} (see Figure \ref{fig: other_anomalies}(a)). In another dynamic cycling profile datasets published by NASA (LCO-graphite cell chemistry),\cite{Bole_dataset.2014,bole2014adaptation} some voltage measurements were found to exceed the specified voltage range between \SI{3.2}{V} and \SI{4.2}{V}. This problem highlights the need for a chemistry-agnostic anomaly detection tool that can be applied across various battery cell chemistries for diverse applications. 

\newpage

\section{Anomalies Definitions and Types}
\label{app: anomalies_definitions}

\begin{figure}[!ht]
\centering
\includegraphics[width=10cm,trim=4 4 4 4,clip]{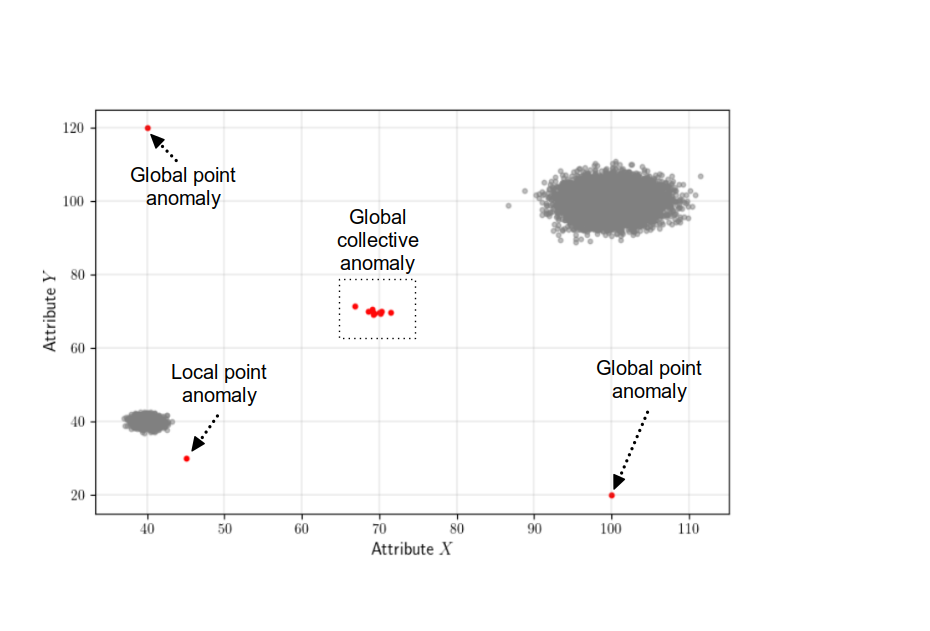}
\caption{\textbf{Illustration of different anomaly definitions.} (1) Point anomaly: a single rare observation; (2) Collective anomaly: a sequence of abnormal measurements; (3) Local anomaly: an unusual observation close to a normal cluster, and (4) Global anomaly: an isolated point far from the main data distribution.}
\label{fig: anomaly definitions}
\end{figure}

\citet{hawkins1980identification} defines an outlier as “an observation which deviates so much from the other observations as to arouse suspicion that it was generated by a different mechanism.” Because there are different scopes and applications of anomaly detection methods, it is important to distinguish different anomaly types (see Figure \ref{fig: anomaly definitions}):

\begin{itemize}[noitemsep]
\item Point anomaly: a single rare observation (for example, an abrupt change in the current measurement due to an internal short-circuit);
\item Collective anomaly: a sequence of abnormal measurements, such as continuous deviations in the recorded voltage and capacity data (see Figure \ref{fig: outlier_examples});
\item Local anomaly: an unusual observation close to a normal cluster;
\item Global anomaly: a distinctly isolated observation far from a normal data cluster.
\end{itemize}

In addition, outliers may also be univariate, where a single feature's value lies far outside its typical range or multivariate, where a combination of features is anomalous. In the \acrshort{ML} literature, three related terms are often distinguished:\cite{scikit-learn2024}

\begin{itemize}[noitemsep]
\item Outlier detection: Anomaly detection models are trained on data containing both normal and anomalous samples, where normal data is assumed to form dense clusters and anomalies occur in low-density regions.
\item Novelty detection: Anomaly detection models are trained on clean data. Anomalies are deviations from the learned normal pattern.
\item Anomaly detection: General term encompassing both approaches.
\end{itemize}

\section{Statistical Feature Transformation Framework}

\begin{figure*}[ht!]
\centering
\includegraphics[width=\textwidth,trim=2 2 2 2,clip]{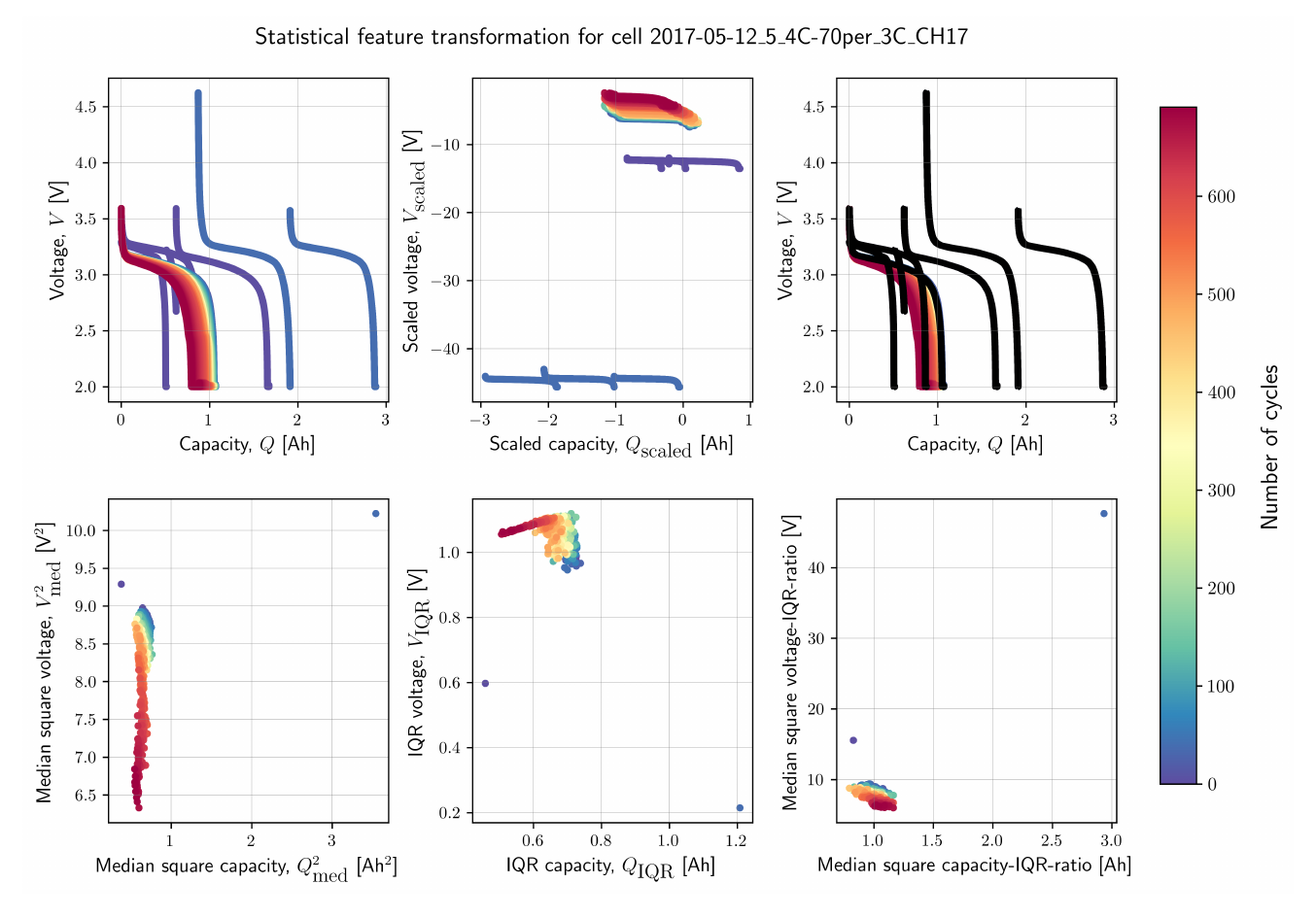}
\caption{\textbf{The inner working of statistical features transformation steps using median and \acrfull{IQR}.} (a) The original voltage and discharge capacity input features with abnormal cycles. (b) Voltage and capacity data after feature transformation, where the cluster of abnormal cycles can be separated from the normal cluster. (c) The original voltage and capacity features with the corresponding identified abnormal clusters (black curves). (d) The median capacity and voltage of each cycle, in which the median capacity and voltage of the abnormal cycle were found to have a distinct value compared to the normal cycles. (e) The \acrshort{IQR} capacity and voltage of each cycle: while the \acrshort{IQR} of normal cycles are densely clustered at the top left of the figure, the \acrshort{IQR} of the anomalous cycle can be found towards the bottom of the figure. (f) By taking the ratio of $\textrm{median}(X)^{2}$ and \acrshort{IQR}, the differences between the normal and abnormal clusters can be further amplified to help with the feature deconvolution.}
\label{fig: explain_stats_feature_transformation}
\end{figure*}

Because the anomalies in the Severson datasets are convoluted, meaning the marginal histograms of abnormal voltage and capacity measurements overlap heavily with those from normal cycles, statistical methods and \acrshort{ML} models cannot be applied directly to detect anomalies in this datasets. Therefore, statistical feature transformation is necessary to facilitate the separation of abnormal cycles from normal cycles. Figure \ref{fig: explain_stats_feature_transformation} illustrates the statistical feature transformation process for separating abnormal battery cycles using the median and \acrfull{IQR} method proposed in Equation \ref{eq: median_IQR_stats_feature_transformation}. Panel (a) shows the original voltage–capacity profiles containing abnormal cycles, while (b) presents the transformed features where abnormal and normal cycles become separable. Panel (c) overlays the original data with identified abnormal clusters (black curves). Panels (d) and (e) display the relationships between median and IQR-based capacity and voltage features, highlighting that abnormal cycles deviate from the tightly clustered normal cycles. Finally, Panel (f) shows that taking the ratio of the median square voltage to the IQR further amplifies the distinction between normal and abnormal clusters, aiding in anomaly identification.

\section{Statistical Anomaly Detection Methods}
\label{app: stats_methods_description}

In this section, we describe the theoretical foundation of the five statistical anomaly detection methods: (1) \acrfull{SD}, (2) \acrfull{MAD}, (3) \acrfull{IQR}, (4) Z-score, and (5) Modified Z-score.  

\subsection{\acrfull{SD}}

\acrshort{SD} is a measure of dispersion in a univariate distribution, which quantifies how spread out the values are around the mean of the datasets:

\begin{equation}
\sigma = \sqrt{\frac{1}{N-1}\sum_{i=1}^{N}(x_i - \overline{x})^2},
\label{eq: SD}
\end{equation}

\noindent Here, $x_i$ represents each data point in the feature, $\overline{x}$ is the sample mean and $N$ is the number of data points. The Bessel's correction, $N-1$, should be used in the denominator instead of $N$ when calculating the \acrshort{SD} using the sample mean and the population mean is unknown. A low standard deviation indicates that the data point is close to the sample mean, whereas a high standard deviation means they are spread out over a wider range of values.

$\mu \pm 3\sigma$ is commonly used to threshold anomalies based on \acrshort{SD}, where the lower bound corresponds to $\mu - 3\sigma$ and the upper bound corresponds to $\mu + 3\sigma$. While implementing this method is straightforward, we should be aware of the assumptions underlying the use of $\mu \pm 3\sigma$ to identify outliers. First and foremost, this statistical threshold method assumes that the data follows a normal distribution (\textit{i.e.} the distribution is approximately symmetric and bell-shaped). For a normal distribution, this assumption implies that 99.7\% of the data falls within $\mu \pm 3\sigma$, and any data points beyond $\mu \pm 3\sigma$ are considered extremely rare and potentially outliers. If the data is not normally distributed, additional data transformation steps are required, or alternative methods, such as median-based anomaly detection methods, might be more appropriate.

\begin{figure}[!ht]
\centering
\includegraphics[width=\textwidth,trim= 4 4 4 4,clip]{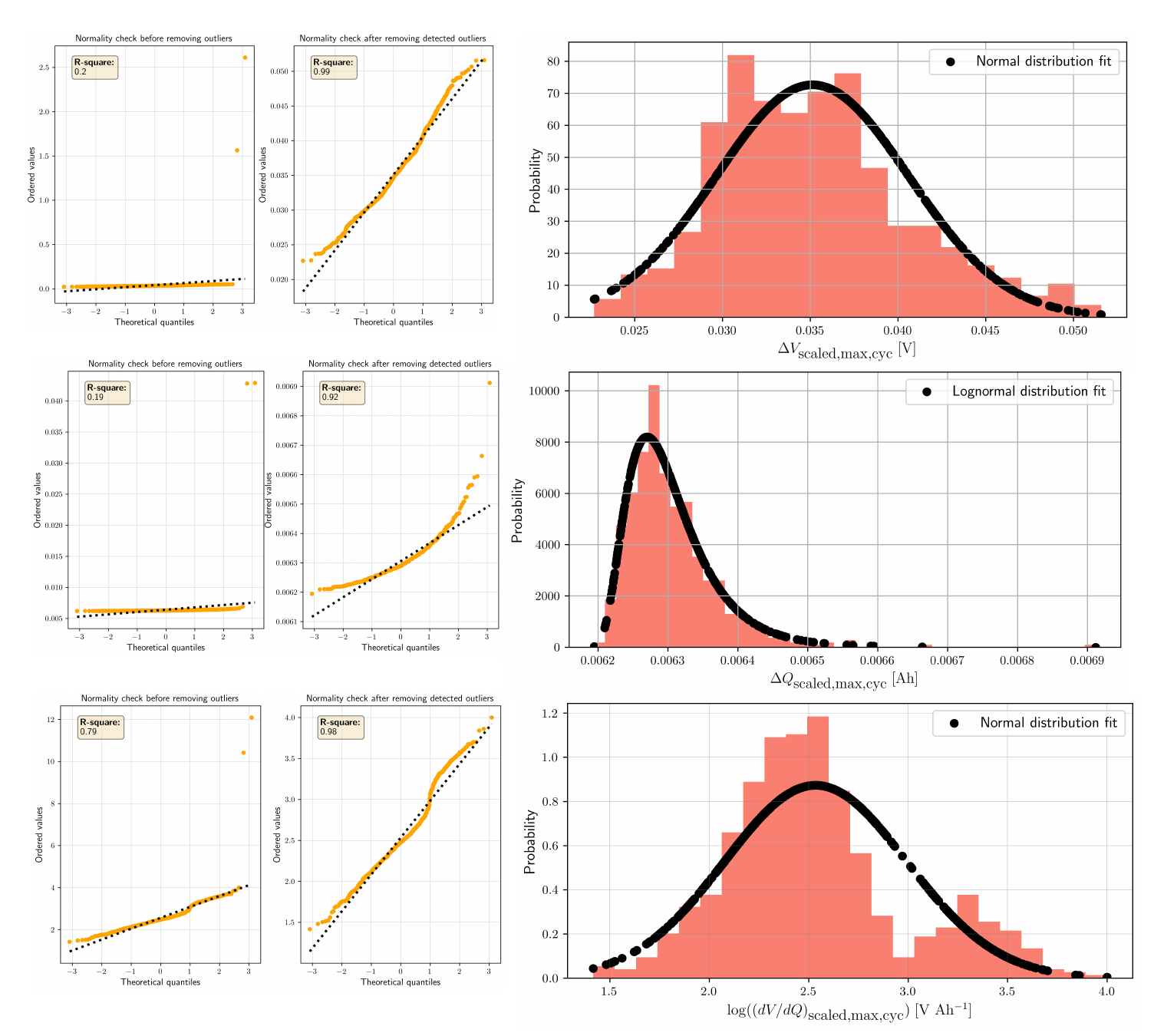}
\caption{\textbf{Feature normality check before and after removing outliers.} (a) For the feature $\Delta V_\textrm{scaled,max,cyc}$, $R^2$ of the probability plot is only 0.2 when anomalies are present in the dataset. As the two major anomalous cycles have outlier scores beyond $\mu \pm 3\sigma$, they can be detected using the statistical \acrshort{SD} method. After removing the detected outliers, the $R^2$-score improves to $0.99$. The histogram of the feature without anomalies also indicates that the feature $\Delta V_\textrm{scaled,max,cyc}$ can be approximated by a Gaussian distribution. (b) The feature $\Delta Q_\textrm{scaled,max,cyc}$ has two anomalies with an $R^2$-score of 0.19 in the probability plot. However, the feature distribution after removing the anomalies is non-Gaussian and exhibits left-skewness. Therefore, $\Delta Q_\textrm{scaled,max,cyc}$ in this example can be best described by a lognormal distribution instead of a normal distribution. (c) The feature $(\log({dV}/{dQ})_\textrm{scaled,max,cyc})$ has two anomalies with a $R^2$-score of 0.79 in the probability plot. This feature can also be described by a Gaussian distribution after removing the detected anomalies.}
\label{fig: normality_check}
\end{figure}

\subsubsection{Feature distribution evaluation}
\label{app: features_probcheck}

When applying \acrshort{SD} to detect anomalies in the battery datasets, we also checked whether the distribution of each transformed feature follows a normal distribution (see Figure \ref{fig: normality_check}). Here, we have generated a probability plot of each transformed feature against the theoretical quantiles of a normal distribution. The straight dotted line in the probability plot indicates a perfect fit to the normal distribution. If most data points fall approximately along a straight line, it implies that the feature is consistent with the normal distribution. Anomalies would appear as points far away from the main cluster and the straight line fit. Nevertheless, some deviation is still observable even after removing the anomalous cycles from the dataset. For example, the feature $\Delta Q_\textrm{scaled,max,cyc}$ from \colorbox{Gainsboro}{Cell 2017-05-12\textunderscore 5\textunderscore 4C-70per\textunderscore 3C\textunderscore CH17} in the Severson dataset\cite{Severson.2019} displays a convex shape from the straight line fit, which indicates the skewness in the feature. The histogram of the feature (after removing obvious anomalies) indicates that $\Delta Q_\textrm{scaled,max,cyc}$ can be best described by a lognormal distribution instead of a normal distribution.

\begin{figure}[!ht]
\centering
\includegraphics[width=\textwidth,trim= 4 4 4 4,clip]{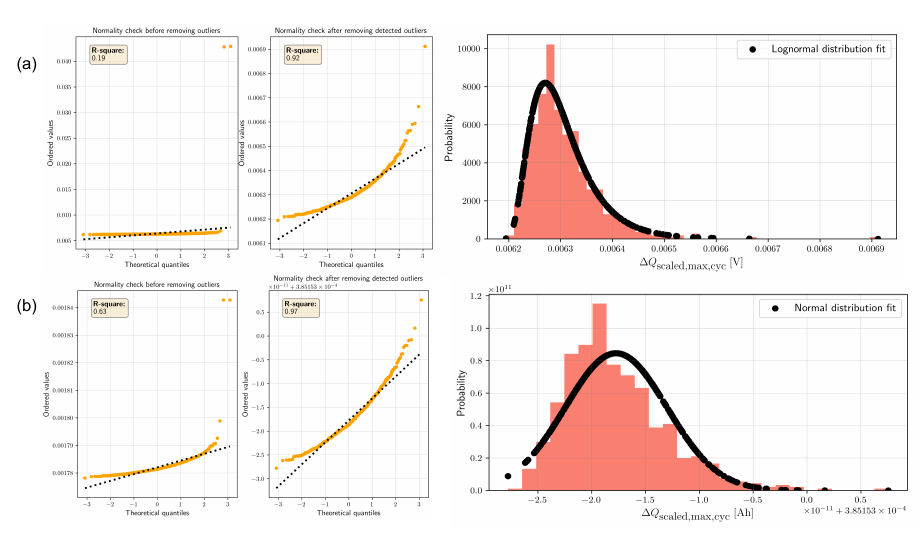}
\caption{\textbf{Yeo-Johnson transformation of the feature distribution.} If a feature is not normally distributed due to skewness in the datasets, the Yeo-Johnson power transformation can be used to reduce the skewness in the feature distribution. (a) The feature $\Delta Q_\textrm{scaled,max,cyc}$ exhibits a convex shape, as indicated by the deviation from the straight-line fit of the probability plot, and the corresponding histogram can be best described by a lognormal distribution. (b) After applying the Yeo-Johnson power transformation, the skewness is reduced, and a normal distribution can be fitted with an $R^2$-score of 0.97. The histogram also looks less skewed after the Yeo-Johnson transformation.}
\label{fig: yeo_johnson_transformation}
\end{figure}

As statistical anomaly detection methods based on \acrshort{SD} and Z-score require the data to be normally distributed, these two methods cannot be applied directly to features with a lognormal distribution. In this case, we further used the Yeo-Johnson power transformation to reduce the skewness in the feature:\cite{yeo2000new}

\begin{equation}
y^{\lambda} = \left \{\begin{array}{rcl} ((y+1)^{\lambda} - 1)/\lambda & \mbox{if} \ & {\lambda \neq 0,\; y \geq 0}, \\
\log(y+1) & \mbox{if} \ & {\lambda = 0,\; y \geq 0}, \\ 
- [(-y+1)^{2-\lambda} - 1)]/(2-\lambda) & \mbox{if} \ & {\lambda \neq 2,\; y < 0}, \\
-\log(-y+1) & \mbox{if} \ & {\lambda = 2,\; y < 0}, \\ 
\end{array} \right.
\end{equation}

\noindent where $y$ is equivalent to a selected non-Gaussian feature in this work. The value of $\lambda$ is automatically tuned by maximizing the log-likelihood function.\cite{2020SciPy-NMeth} Figure \ref{fig: yeo_johnson_transformation} compares the effects before and after applying the Yeo-Johnson transformation to the feature $\Delta Q_\textrm{scaled,max,cyc}$, where the skewness has decreased to better reflect a normal distribution fit.

\subsection{\acrfull{MAD}}

Instead of using \acrshort{SD} to estimate the distance of a value from the distribution mean, \acrfull{MAD} is another statistical method that can be be used to flag potential anomalies in a one-dimensional distribution:\cite{rousseeuw1993alternatives}

\begin{equation}
\textrm{MAD} = |F_\textrm{MAD}|\times \textrm{median}(|x_i - M|),
\label{eq: MAD}
\end{equation} 

\noindent where $x_i$ represents each data point in the feature and $M$ is the median of the feature ($M = \textrm{median}(x)$). Then, the absolute difference from that median and median of those absolute deviations can be computed to calculate the \acrshort{MAD}-score. Here, $F_\textrm{MAD}$ is the \acrshort{MAD}-factor. After calculating the \acrshort{MAD}-score, the upper and lower limit of the \acrshort{MAD}-score can be estimated as follows:\cite{yang2019outlier}

\begin{align}
\textrm{MAD}_\textrm{lower} &= M - (3\times \textrm{MAD}) \nonumber \\
\textrm{MAD}_\textrm{upper} &= M + (3\times \textrm{MAD}),
\label{eq: MAD-limit}
\end{align}

\begin{figure}[!ht]
\centering
\includegraphics[width=\textwidth,trim= 4 4 4 4,clip]{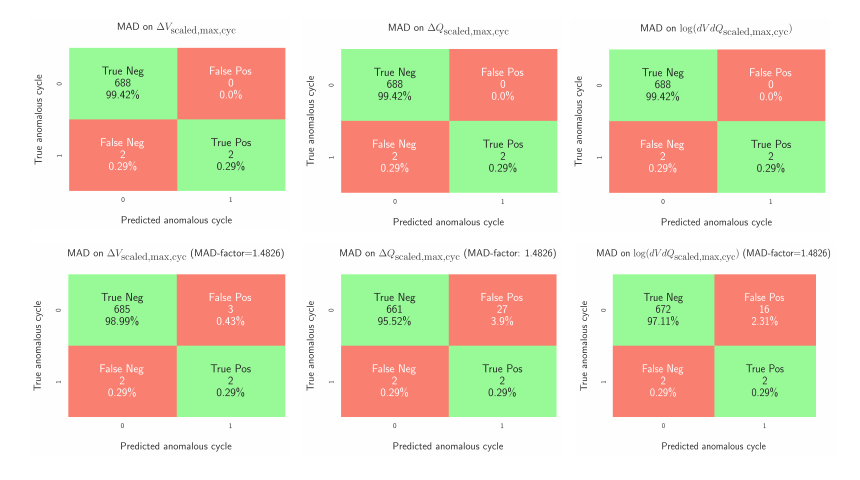}
\caption{\textbf{The effects of using different MAD-factors in anomaly detection.} In the first row across the top panel, $F_\textrm{MAD}$-factors were calculated from the reciprocal of the 75th-percentile of the standard distribution. In the second row across the bottom panel, $F_\textrm{MAD}$ was assumed to be $1.4826$. The \acrshort{MAD}-detector with the \acrshort{MAD}-factor of 1.4826 was found to be less accurate (\textit{i.e.} with more false positives) compared to the calculated \acrshort{MAD}-factor from the 75th-percentile of the standard distribution.}
\label{fig: stats_confusion_matrix_MAD}
\end{figure}

Any data points that exceed the $\textrm{MAD}_\textrm{lower}$ and $\textrm{MAD}_\textrm{upper}$ limits will be flagged as anomalous. We can assume that the \acrshort{MAD}-factor ($F_\textrm{MAD}$) is 1.4826 in Equation \ref{eq: MAD} if a specific feature has a normal distribution.\cite{rousseeuw1993alternatives,leys2013detecting,rosenmai2013using} However, if the data distribution is not Gaussian, then $F_\textrm{MAD}$ should be calculated from the reciprocal of the 75th-percentile of the corresponding standard distribution.\cite{rosenmai2013using} Figure \ref{fig: stats_confusion_matrix_MAD} benchmarks the anomalous cycle prediction using different \acrshort{MAD}-factors. Here we show that the commonly used \acrshort{MAD} factor ($F_\textrm{MAD} = 1.4826$) may lead to many false positives prediction if the underlying distribution is not Gaussian.

\subsection{\acrfull{IQR}}
\acrfull{IQR} is a statistical method that relies on splitting the entire data distribution into four parts, where each part is a quartile and \acrshort{IQR} can be calculated as the difference between the third quartile ($Q3$) and first quartile ($Q1$) of the feature distribution:

\begin{equation}
IQR = Q_3 - Q_1.
\label{eq: IQR}
\end{equation}

The first quartile ($Q_1$) is the median of the lower half of the datasets ($25^\textrm{th}$ percentile), whereas the third quartile ($Q_3$) is the median of the upper half of the datasets ($75^\textrm{th}$ percentile). \acrshort{IQR} can be visualized using boxplot, where the data distribution can be described based on a five number summary (\textit{i.e.} minimum, Q1, median, Q3 and maximum). By using the \acrshort{IQR} method, measurements are denoted as outliers if they fall outside the lower and upper bound:

\begin{align}
IQR_\textrm{lower} &= Q_1 - (1.5 \times IQR), \nonumber \\
IQR_\textrm{upper} &= Q_3 + (1.5 \times IQR),
\label{eq: IQR_limits}
\end{align}

The \acrshort{IQR}-detector was found to have the highest false positive prediction compared to the \acrshort{SD} and \acrshort{MAD}-detector. By comparing the limits of \acrshort{IQR}, \acrshort{MAD}-detector and \acrshort{SD}-detector, the \acrshort{IQR} limits were found to be most conservative. As a result, some of the normal cycles at the distribution tail were erroneously classified as outliers. \acrshort{IQR} range has been typically used in a boxplot to represent the distribution of a univariate feature, where data points outside the whiskers (\textit{i.e.} $IQR_\textrm{lower}$ and $IQR_\textrm{upper}$ limits) are considered as outliers. However, as shown in this benchmarking study, using boxplot to identify anomalies may not always be reliable as some normal cycles towards the distribution tail may be wrongly classified as outliers (see Figure \ref{fig: compare_stats_limit}).

\begin{figure}[!ht]
\centering
\includegraphics[width=8.5cm]{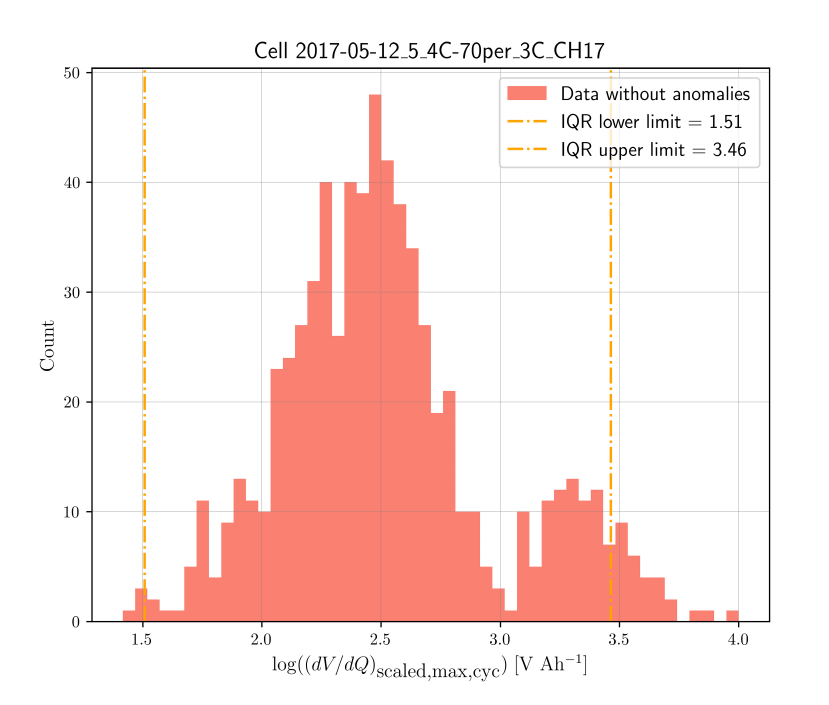}
\caption{\textbf{Histogram of the logarithmic maximum scaled differential voltage feature, $\log({dV}/{dQ})_\textrm{scaled,max,cyc}$} The histogram represents data for the cell-index \colorbox{Gainsboro}{Cell 2017-05-12\textunderscore 5\textunderscore 4C-70per\textunderscore 3C\textunderscore CH17} in the Severson dataset after removing major anomalies, in which the dashed-dotted lines indicate the \acrfull{IQR} thresholds used for anomaly detection. The lower and upper IQR limits, 1.51 and 3.46 respectively, define the normal operating range, with values outside this range considered as potential anomalies. As the \acrshort{IQR} limits are most conservative, many normal cycles towards the distribution tail were flagged as outliers, leading to many \acrfull{FP} in the confusion matrix.}
\label{fig: compare_stats_limit_IQR}
\end{figure} 

\begin{figure*}[ht!]
\centering
\includegraphics[width=\textwidth]{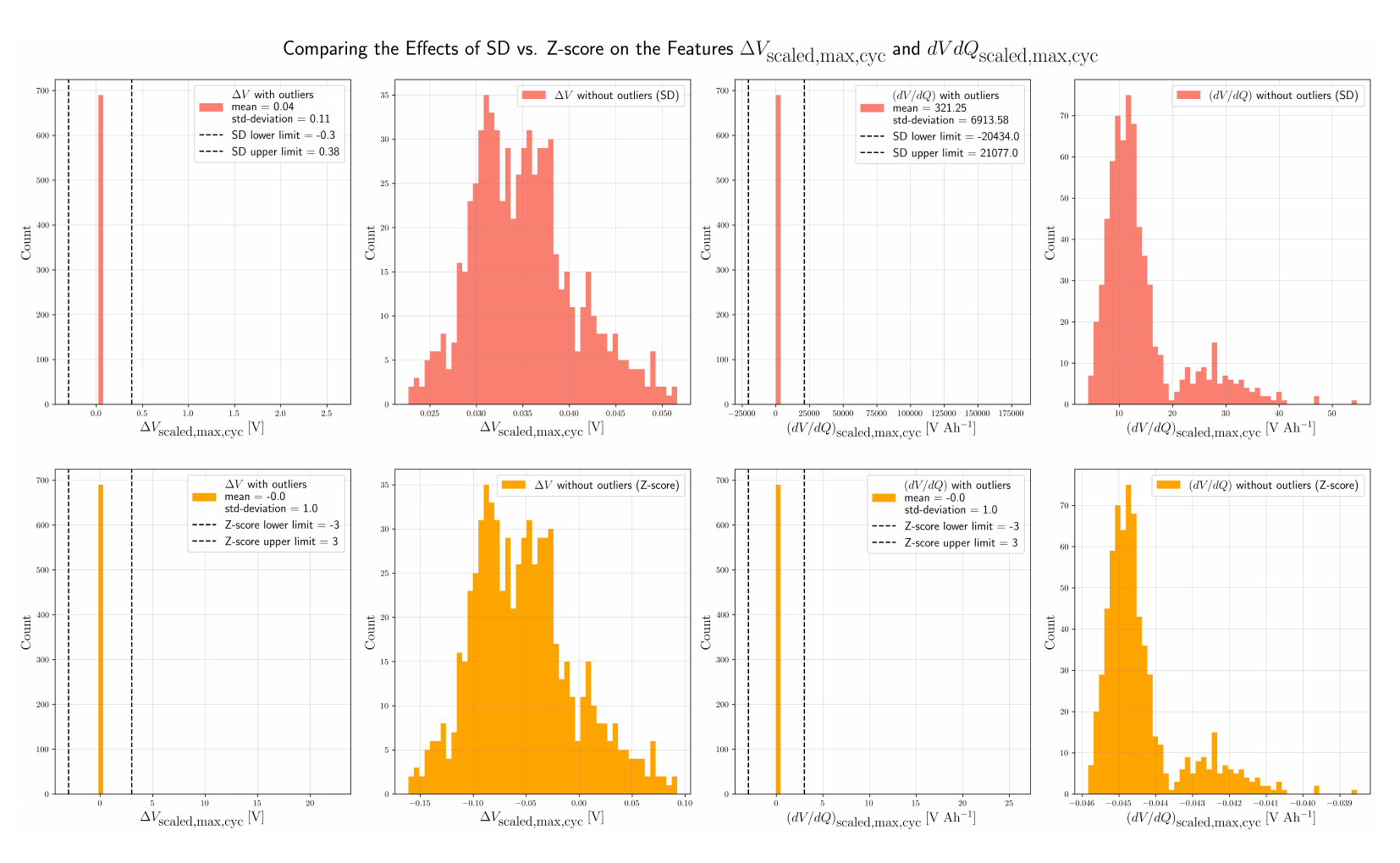}
\caption{\textbf{The difference between \acrfull{SD} and Z-score on detecting anomalous cycles in the battery systems.} The top panel shows the histograms with and without outliers (salmon color) when using \acrshort{SD} to detect anomalies in the feature $\Delta V_\textrm{scaled,max,cyc}$ and $({dV}/{dQ})_\textrm{scaled,max,cyc}$, whereas the bottom panel displays the histograms with and without outliers (orange color) when using Z-score to detect the anomalies across the same two features. By calculating the Z-score of each feature, both features (irrespective of the original value range in the salmon histograms) are standardized to have a mean of zero with a standard deviation of one. The Z-score limits of both features are also standardized to have a value range between -3 and 3.}
\label{fig: compare_zscore_sd}
\end{figure*}

\subsection{Z-score}

The fourth statistical method we have implemented for benchmarking anomaly detection is the Z-score: 

\begin{equation}
z_i = \frac{x_i - \mu}{\sigma},
\label{eq: standard z-score}
\end{equation}

\noindent where $x_i$ denotes the value of each data point in the univariate distribution, $\mu$ and $\sigma$ are the mean and standard deviation of the selected feature distribution. In principle, Z-score has a similar working principle like \acrshort{SD}, as both methods rely on the mean and standard deviation to threshold outliers. However, Z-score standardizes each feature to have a zero mean with a unity standard deviation, making the comparison between different features with different values range possible. 

Figure \ref{fig: compare_zscore_sd} compares the difference when using the detector \acrshort{SD} and Z-score to screen for anomalies in the features $\Delta V_\textrm{scaled,max,cyc}$ and $({dV}/{dQ})_\textrm{scaled,max,cyc}$. Due to the very different range of these two features (see the histogram of $\Delta V_\textrm{scaled,max,cyc}$ (with outliers) compared to the histogram of $({dV}/{dQ})_\textrm{scaled,max,cyc}$ (with outliers) in the top panel), their means ($\mu_{\Delta V} = 0.04$ versus $\mu_{(dV/dQ)} = 321.25$) and standard deviations also vary significantly ($\sigma_{\Delta V} = 0.11$ versus $\sigma_{(dV/dQ)} = 6913.58$). A a result, $\Delta V_\textrm{scaled,max,cyc}$ has a \acrshort{SD} limits between -0.3 and 0.38, whereas $({dV}/{dQ})_\textrm{scaled,max,cyc}$ has a \acrshort{SD} limits between $-20434$ and $21077$.

When we calculated the Z-score of each feature, we also transformed and standardized the corresponding feature range (see the ranges of both features in the bottom panel represented by the orange histograms). Regardless of the original value range in the salmon histogram, both features now have a mean of zero with a standard deviation of one. The Z-score limits are also standardized to have a value range between -3 and 3. This example highlights the advantage of using Z-score to detect anomalies, which enables an effective comparison of the anomaly score between different features. It is important to note that Z-transformation only standardizes the feature range without changing the distribution shape (see Figure \ref{fig: compare_zscore_sd}).  

In general, Z-score measures how many standard deviations a data point is from the mean of a dataset. If the calculated Z-score has a value of 0, this means the value is exactly at the mean.\cite{colan2013and} A Z-score of 1 denotes that the value is 1 standard deviation above the mean, whereas a Z-score of -2 means the value is 2 standard deviations below the mean.\cite{colan2013and} In this example, both cycles 0 and 40 were considered as outliers because their outlier scores after Z-transformation are more 3 standard deviations from the mean. Nevertheless, like the statistical method \acrshort{SD}, Z-score may not be effective if the underlying data distribution is not Gaussian.

\subsection{Modified Z-score}

The last statistical anomaly detection method benchmarked in our work is the modified Z-score that can be used to detect anomalies if the distribution is not normally distributed. Instead of using the mean and standard deviation (as shown by Equation \ref{eq: standard z-score}), the modified z-score uses the median and median absolute deviations to quantify how far a value is from the center of the dataset:\cite{yaro2024outlier}

\begin{equation}
\textrm{Modified Z-score} = \frac{x_i - \textrm{median}(X)}{\textrm{MAD}},
\label{eq: Modified z-score}
\end{equation} 

A coefficient of 0.6745 is sometimes added to the numerator of Equation \ref{eq: Modified z-score} to make the modified z-score equivalent to the standard Z-score for a Gaussian distribution. One should note that this coefficient is actually the reciprocal of the \acrshort{MAD}-factor ($F_\textrm{MAD} = 1.4826$).\cite{yaro2024outlier} If the distribution is not Gaussian, then this coefficient should be calculated from the 75th-percentile of the corresponding standard distribution.

\begin{table*}[!h]
\caption{Characteristics, advantages and limitations of statistical anomaly detection methods}
\centering
% The number next to resizebox fit the table into page
% When setting to \textwidth, the table will be fitted into the A4 page
\resizebox{\textwidth}{!}{
\begin{tabularx}{23cm}{p{2.5cm}ccccc} \\ \toprule

 & \makecell{Standard\\Deviation (SD)} & \makecell{Median\\Absolute Deviation (MAD)} & \makecell{Interquartile\\Range (IQR)} & Z-score & \makecell{Modified\\Z-score} \\ \midrule
Central tendency & {\makecell*[{{p{3.5cm}}}]{Mean}} & {\makecell*[{{p{3.5cm}}}]{Median}} & {\makecell*[{{p{3.5cm}}}]{Median}} & {\makecell*[{{p{3.5cm}}}]{Mean}} & {\makecell*[{{p{3.5cm}}}]{Median}} \\

Variability & {\makecell*[{{p{3.5cm}}}]{Standard deviation}} & {\makecell*[{{p{3.5cm}}}]{Median absolute deviation}} & {\makecell*[{{p{3.5cm}}}]{Difference between the first and third quartile}} & {\makecell*[{{p{3.5cm}}}]{Standard deviation}} & {\makecell*[{{p{3.5cm}}}]{Median absolute deviation}} \\

Statistical limits & {\makecell*[{{p{3.5cm}}}]{$\mu \pm 3\sigma$}} & {\makecell*[{{p{3.5cm}}}]{$|x_i - M|/\textrm{MAD} > 3$}} & {\makecell*[{{p{3.5cm}}}]{$Q_1 - (1.5 \times IQR)$ \\ $Q_3 + (1.5 \times IQR)$}} & {\makecell*[{{p{3.5cm}}}]{$|Z| > 3$}} & {\makecell*[{{p{3.5cm}}}]{$|Z_\textrm{mod}| > 3.5$}} \\

Advantages & {\makecell*[{{p{3.5cm}}}]{Easy to use.}} & {\makecell*[{{p{3.5cm}}}]{\acrshort{MAD} can be used for skewed distribution.}} & {\makecell*[{{p{3.5cm}}}]{Like \acrshort{MAD}, \acrshort{IQR} also works for skewed distribution.}} & {\makecell*[{{p{3.5cm}}}]{Standardize each feature to have a zero mean with a unity standard deviation.}} & {\makecell*[{{p{3.5cm}}}]{Similar to \acrshort{MAD} and \acrshort{IQR}, skewed distributions can be described by modified Z-score.}} \\

Limitations & {\makecell*[{{p{3.5cm}}}]{\acrshort{SD} assumes data to follow a Gaussian distribution; highly sensitive to extreme outliers that could skew the mean and standard deviation.}}

& {\makecell*[{{p{3.5cm}}}]{Using the common MAD factor ($F_\text{MAD} = 1.4826$) with non-Gaussian or asymmetrical distribution could lead to inaccurate anomaly detection.\cite{leys2013detecting, rosenmai2013using,rousseeuw1993alternatives}}} 

& {\makecell*[{{p{3.5cm}}}]{A conservative IQR threshold may increase the risk of false positives (Type I errors).}} 

& {\makecell*[{{p{3.5cm}}}]{Like \acrshort{SD}, Z-score also assumes a Gaussian distribution; highly sensitive to extreme outliers that could skew the mean and standard deviation.}} 

& {\makecell*[{{p{3.5cm}}}]{Similar to \acrshort{MAD}, using the common MAD factor ($F_\text{MAD} = 1.4826$) for modified Z-score with non-Gaussian distribution could lead to inaccurate anomaly detection.\cite{leys2013detecting, rosenmai2013using,rousseeuw1993alternatives}}}  \\
\bottomrule
\end{tabularx}}
\label{table: stats_benchmark_outliers}
\end{table*}

\begin{figure*}[ht!]
\centering
\includegraphics[width=\textwidth]{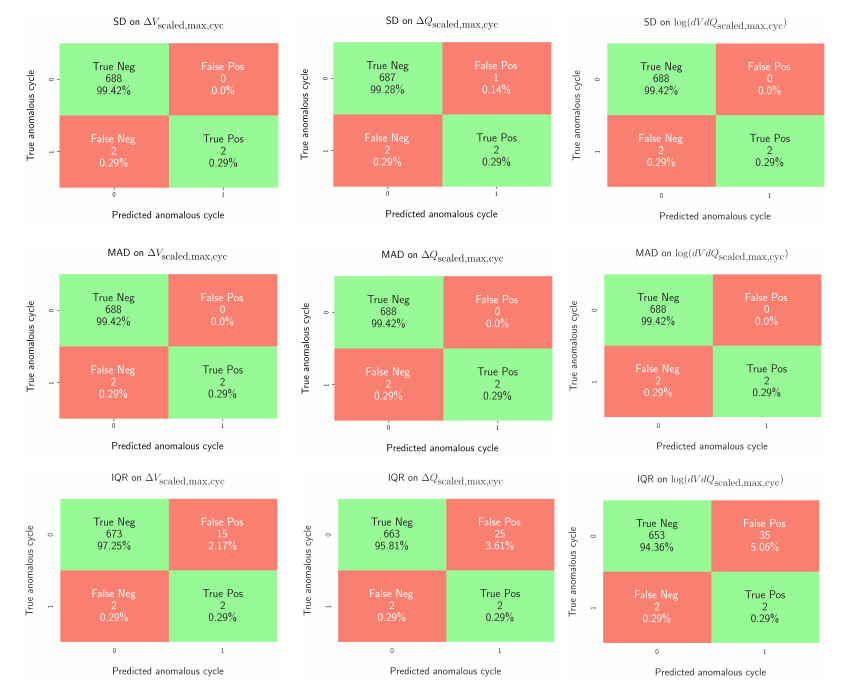}
\caption{\textbf{Confusion matrices to benchmark different statistical anomaly detection methods.} The first columns corresponds to the first extracted feature - maximum scaled voltage difference per cycle ($\Delta V_\textrm{scaled,max,cyc}$), whereas the second column represents the second feature - maximum scaled capacity difference per cycle ($\Delta Q_\textrm{scaled,max,cyc}$) and the third column denotes the combined feature - maximum scaled ratio of voltage to capacity difference per cycle ($({dV}/{dQ})_\textrm{scaled,max,cyc}$). The first row corresponds to using \acrfull{SD} to detect point anomalies of these extracted features. The second row denotes the confusion matrices when using \acrfull{MAD}, whereas the third row represents the confusion matrices of applying \acrfull{IQR} to distinguish between normal and anomalous cycles. \acrfull{TP} means the specific statistical method correctly identifies an anomalous cycle, whereas \acrfull{TN} means a cycle is correctly predicted as normal (represented as light green in the confusion matrix). The salmon color in the confusion matrix represents \acrfull{FP} and \acrfull{FN}, in which \acrfull{FP} indicates that the statistical method incorrectly identifies a normal cycle as anomalous (Type I error), and \acrfull{FN} means the method incorrectly predicts an anomalous cycle as normal (Type II error).}
\label{fig: stats_confusion_matrix}
\end{figure*}

\subsection{Confusion Matrices}

Figure \ref{fig: stats_confusion_matrix} compares the performance of different statistical methods (\textit{i.e.}, \acrshort{SD}, \acrshort{MAD}, and \acrshort{IQR}) for detecting anomalous battery cycles in the Severson datasets based on the extracted features. While each row represents a different method, the columns correspond to the three extracted features: maximum scaled voltage difference per cycle ($\Delta V_\textrm{scaled,max,cyc}$), maximum scaled capacity difference per cycle ($\Delta Q_\textrm{scaled,max,cyc}$), and their combined feature $\log({dV}/{dQ})_\textrm{scaled,max,cyc}$.

Each confusion matrix compares predicted versus true anomalous cycles labeled in the benchmarking datasets, where \acrfull{TP} indicate correctly identified anomalous cycles, \acrfull{TN} represent correctly classified normal cycles, \acrfull{FP} denote normal cycles incorrectly identified as anomalies (Type I error) and \acrfull{FN} indicate anomalous cycles incorrectly classified as normal (Type II error). The confusion matrices show that \acrshort{SD} and \acrshort{MAD} achieve high accuracy with minimal false positives and negatives. Due to the conservative statistical limits from the \acrshort{IQR} method, we show that anomaly detection with the \acrshort{IQR} method could cause up to 5\% of \acrshort{FP} compared to the \acrshort{SD} and \acrshort{MAD} methods.

\section{Statistical Anomalies Detection}
\subsection{Visual inspection of flagged anomalies}
\label{app: visual_check_anomalies}

\begin{figure}[!ht]
\centering
\includegraphics[width=\textwidth,trim=4 4 4 4,clip]{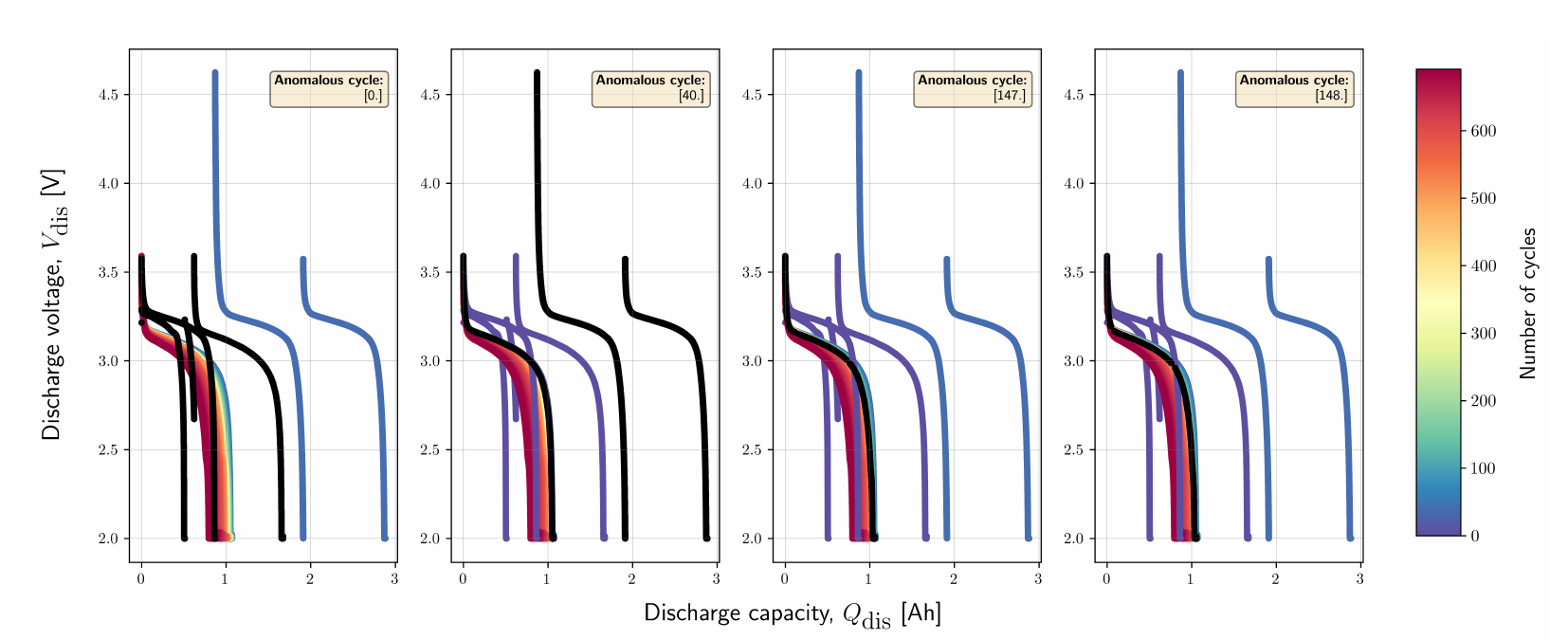}
\caption{\textbf{Visualization of identified anomalous discharge cycles for a representative lithium-ion cell.} Discharge voltage-capacity profiles are shown for cycles across the full lifetime of the cell, with color indicating the number of cycles. Each panel highlights a specific anomalous cycle (black curve) with the corresponding cycle index noted in the figure inset. Cycles 0 and 40 are annotated as major anomalies due to their distinct deviations from the typical discharge patterns, whereas cycles 147 and 148 are considered as minor anomalies.}
\label{fig: visual_check_anomalies_major}
\end{figure}

\begin{figure}[!ht]
\centering
\includegraphics[width=\textwidth,trim=4 4 4 4,clip]{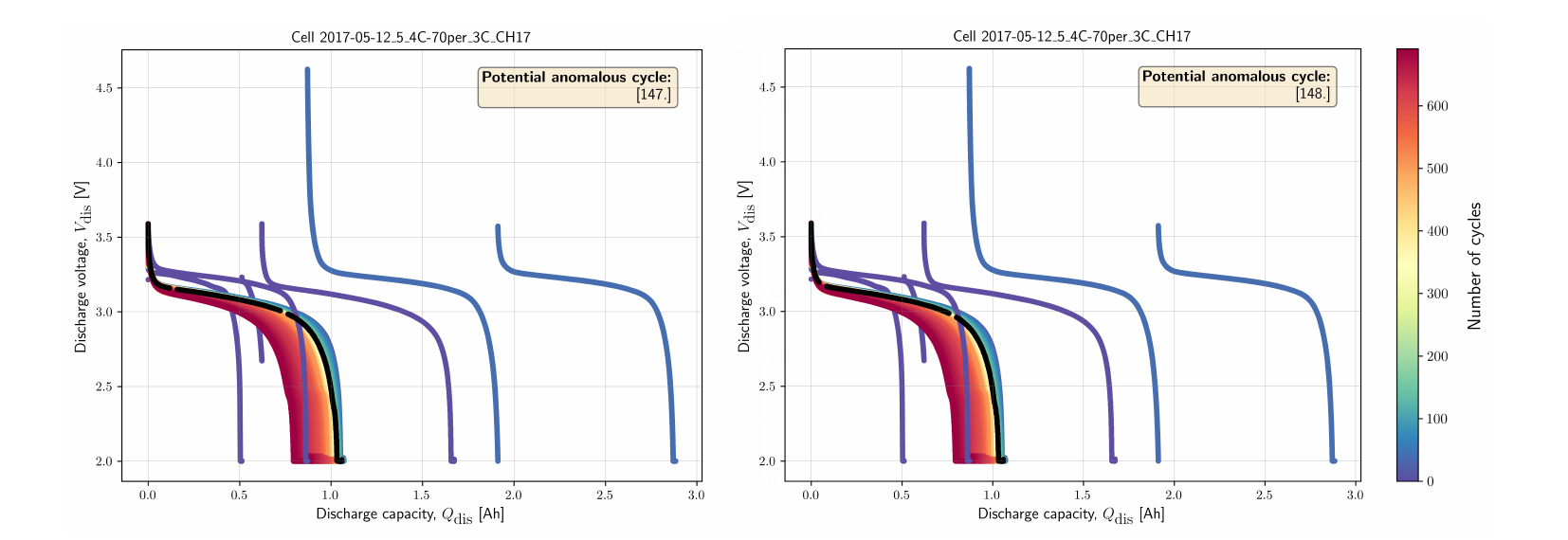}
\caption{\textbf{Enlarged visualization of anomalous discharge cycles for cycles 146 and 147.} Although these two cycles do not display clear deviations from the normal discharge pattern, they were identified as minor anomalies due to subtle discontinuities in their cycling patterns, implying potential measurement inconsistencies.}
\label{fig: visual_check_anomalies_minor}
\end{figure}

Figure \ref{fig: visual_check_anomalies_major} and Figure \ref{fig: visual_check_anomalies_minor}illustrate the identified major and minor anomalies from \colorbox{Gainsboro}{Cell 2017-05-12\textunderscore 5\textunderscore 4C-70per\textunderscore 3C\textunderscore CH17} in the Severson dataset.\cite{Severson.2019} Here, cycles [0, 40, 147, 148] are annotated as outliers in the benchmarking datasets. Because cycles 0 and 40 exhibit obvious anomalous cycling profiles compared to normal cycles, these two cycles are identified as major anomalies that could potentially affect cell performance and safety in real-world battery operations. On the other hand, cycles 147 and 148 were identified as minor anomalies due to some discontinuities in the discharge voltage-capacity curves. While these two cycles do not necessarily have detrimental effects on the cell performance, discontinuities in the cycling data could nevertheless cause biased learned patterns or alter distance metrics in the downstream \acrshort{ML} predictions, especially in models sensitive to feature completeness.

\newpage

\section{Distance-based Anomalies Detection}
\label{app: distance_based_anomaly_detection}

\begin{figure*}[ht!]
\centering
\includegraphics[width=\textwidth]{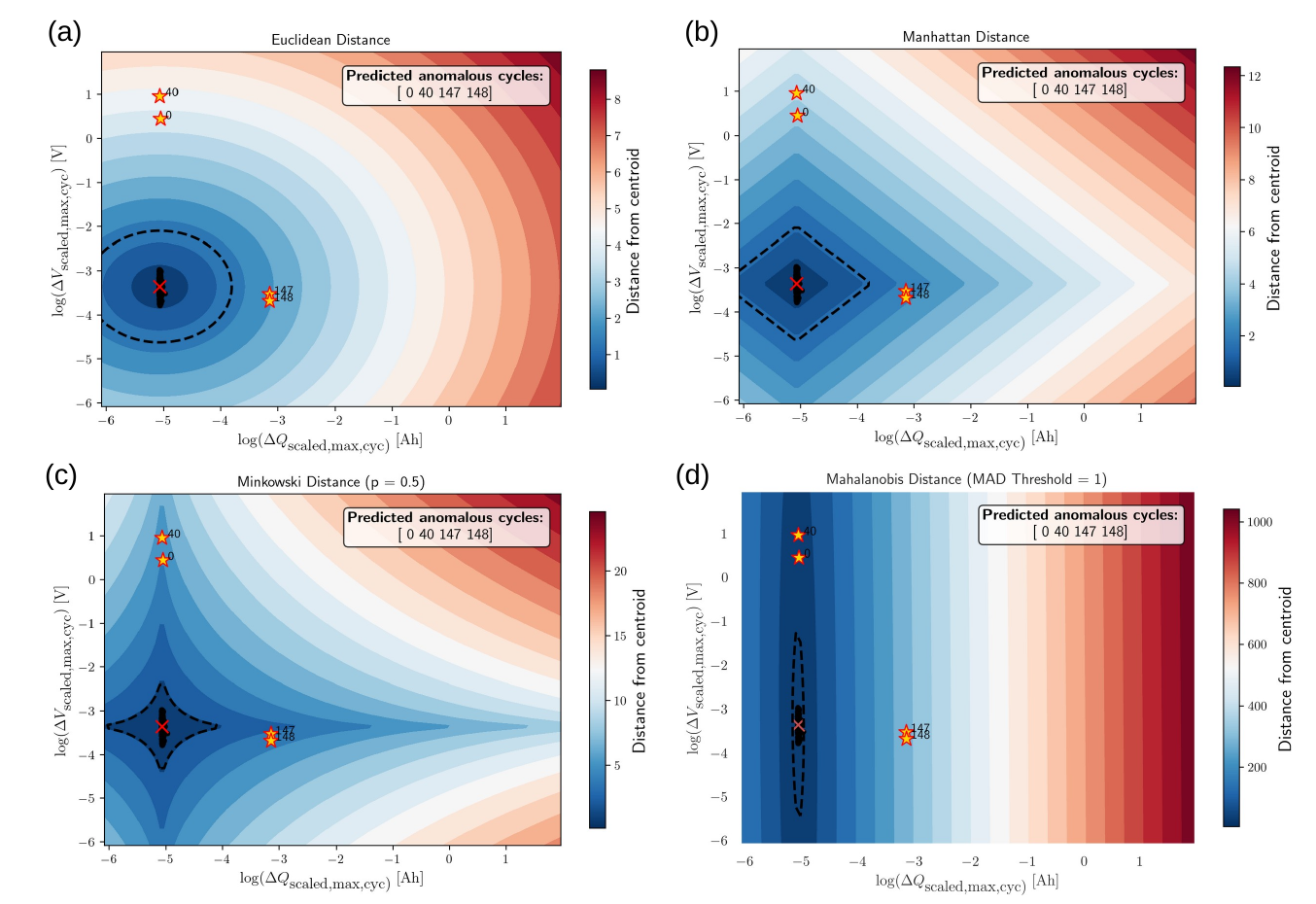}
\caption{\textbf{Visualization of distance-based anomaly detection methods applied to battery cycle data using two features: $\log(\Delta Q_\textrm{scaled,max,cyc})$ and $\log(\Delta V_\textrm{scaled,max,cyc})$.} Here, the distance from the centroid in the feature space is not normalized for (a) Euclidean distance, (b) Manhattan distance, (c) Minkowski distance ($p=0.5$), and (d) Mahalanobis distance (with $\textrm{MAD threshold} = 1$).}
\label{fig: distance_contour_map}
\end{figure*}

While Figure \ref{fig: distance_contour_map_norm} illustrates the normalized distance from the centroid for all four distance metrics evaluated in this work, Figure \ref{fig: distance_contour_map} shows the non-normalized version of the same metrics for comparison. 

\begin{figure*}[ht!]
\centering
\includegraphics[width=\textwidth]{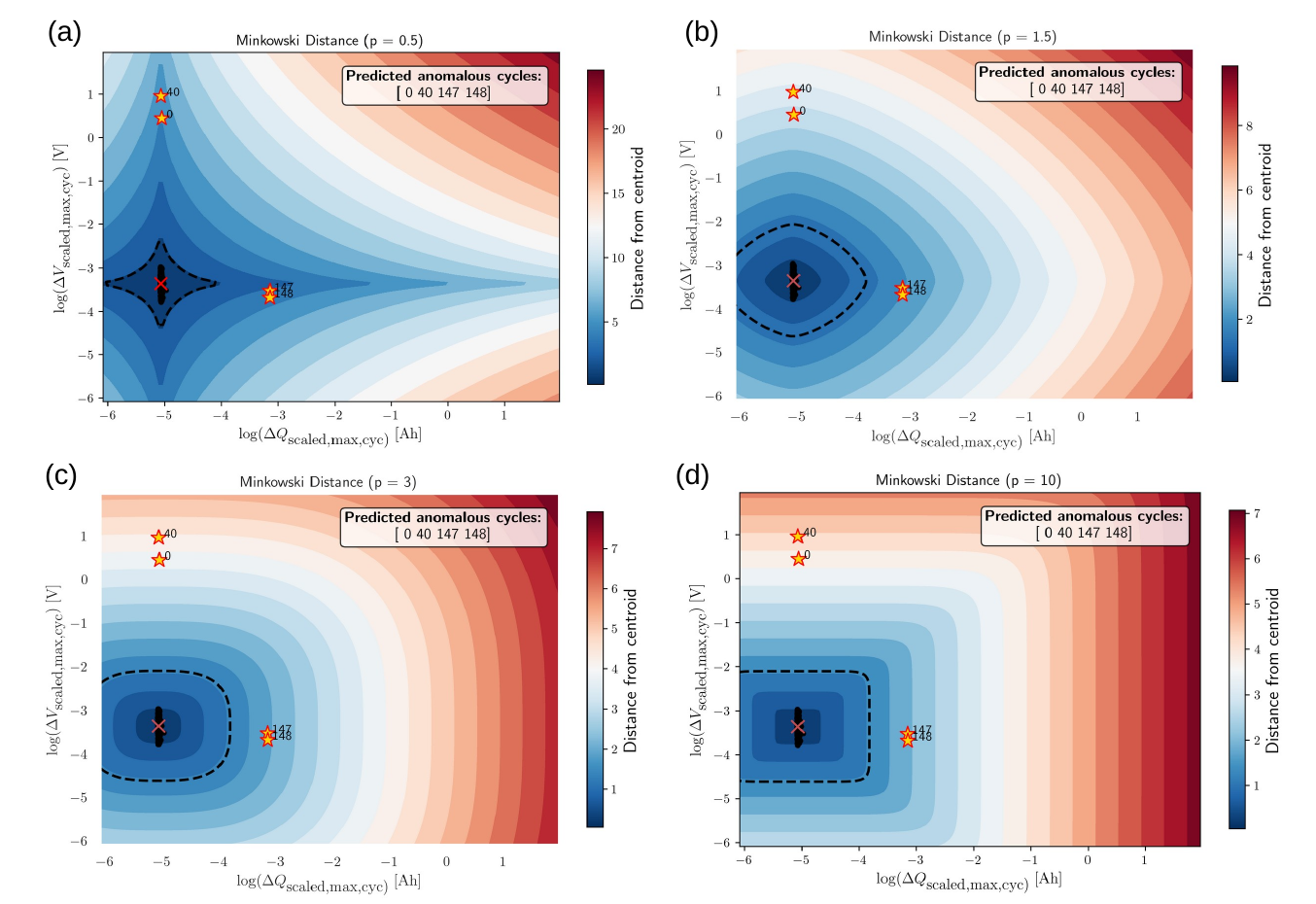}
\caption{\textbf{Effect of Minkowski distance order $p$ on anomaly detection in the feature space.} Contour maps illustrate the distance from the centroid in the two-dimensional feature space defined by $\log(\Delta Q_\textrm{scaled,max,cyc})$ and $\log(\Delta V_\textrm{scaled,max,cyc})$. Panels (a-d) correspond to the Minkowski distance orders of $p=0.5$, $1.5$, $3$, and $10$, respectively. The red "X" denotes the centroid of the data distribution, while the dashed contour represents the anomaly detection boundary. Orange stars indicate the locations of predicted anomalous cycles, along with their corresponding indices. The inset in each panel lists the predicted anomalous cycles, which remain consistent across different $p$-values.}
\label{fig: distance_contour_map_minkowski}
\end{figure*}

Figure \ref{fig: distance_contour_map_minkowski} demonstrates the effect of varying Minkowski distance order ($p$) on the anomaly detection performance within a two-dimensional feature space defined by $\log\Delta V_\textrm{scaled,max,cyc}$ and  $\log\Delta Q_\textrm{scaled,max,cyc}$. Panels (a-d) correspond to different Minkowski distance orders $p = 0.5$, $1.5$, $3$, and $10$, respectively. The contour maps represent the distance from the centroid of the data distribution (marked as red "X"), while the dashed black contour indicates the anomaly detection boundary. Orange stars mark the locations of predicted anomalous cycles (\textit{i.e.} cycles 0, 40, 147, and 148), which remain consistent across all $p$ values. While a small Minkowski distance order ($p = 0.5$) yields a diamond-like contour map, a large Minkowski distance order ($p = 10$) yields a square-like contour map. Nevertheless, the results show that varying $p$-values does not affect the detection outcome, indicating robustness of the anomaly identification process with respect to the Minkowski distance order for this dataset.

\begin{figure*}[ht!]
\centering
\includegraphics[width=\textwidth]{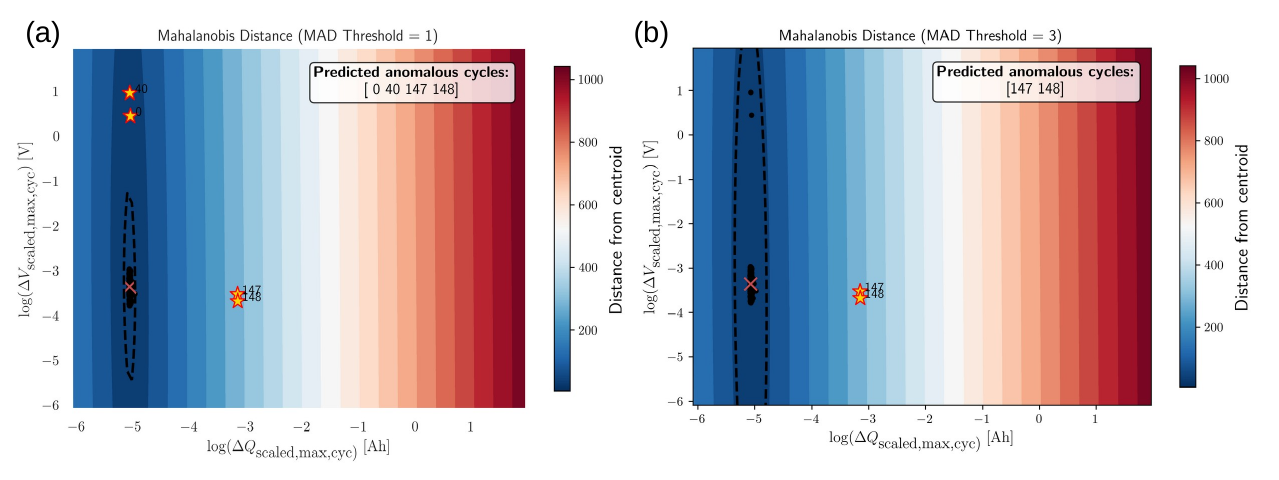}
\caption{\textbf{Effect of Mahalanobis distance threshold on anomaly detection in the feature space.} Panels (a-b) correspond to Mahalanobis distance analyses with \acrfull{MAD} thresholds of 1 and 3, respectively. The inset in each panel lists the predicted anomalous cycles, showing that higher \acrshort{MAD} thresholds result in fewer detected anomalies, as these anomalies are aligned with the joint distribution.}
\label{fig: distance_contour_map_mahalanobis}
\end{figure*}

Figure \ref{fig: distance_contour_map_mahalanobis} illustrates the effect of varying the Mahalanobis distance threshold on anomaly detection in the same feature space. At a lower \acrshort{MAD} threshold ($\textrm{MAD}=1$), four cycles (0, 40, 147, 148) are detected as anomalies, while at a higher \acrshort{MAD} threshold ($\textrm{MAD}=3$), only two cycles are identified as anomalies. This is because the Mahalanobis distance considers the covariance among variables to determine the multidimensional distance from a point to a distribution, which leads to the observed elliptical contour shape. While anomalies that deviate from the joint distribution can be identified easily using the Mahalanobis distance (\textit{e.g.}, cycles 0 and 40), this distance metric may miss anomalies that are aligned with the distribution (cycles 0 and 40). Therefore, to mitigate missing true anomalies, a smaller \acrshort{MAD} threshold instead of a larger \acrshort{MAD} threshold should be used.

\newpage

\section{Data-driven Anomaly Detection}

\subsection{Multivariate Anomalies}
\label{app: multivariate_anomalies}

\begin{figure}[!ht]
\centering
\includegraphics[width=8cm,trim=4 4 4 4,clip]{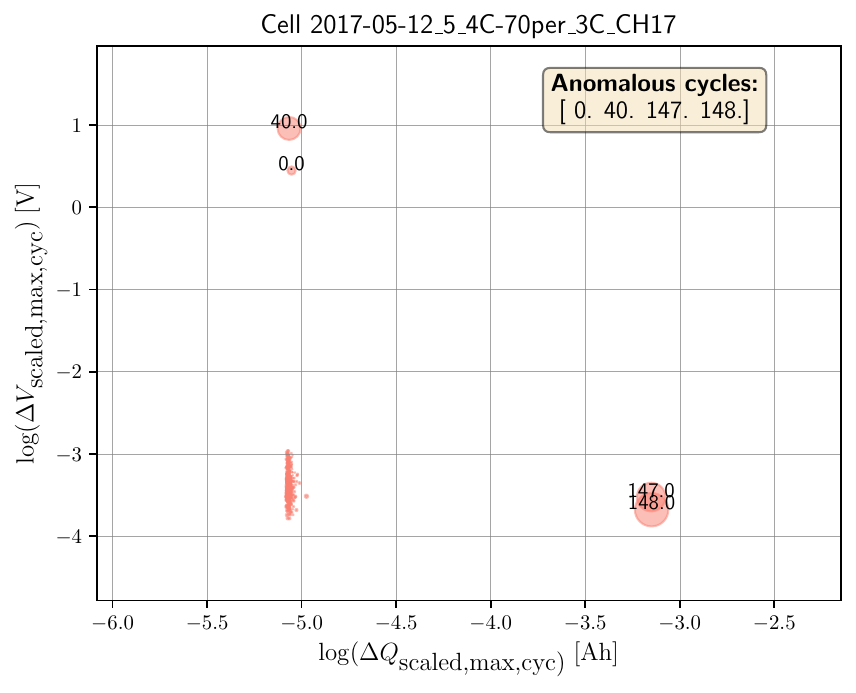}
\caption{\textbf{An example of a bubble plot with multivariate anomalies.} Anomalous battery cycling pattern due to spikes in both voltage and capacity measurements, represented by the extracted features $\log(\Delta V_\textrm{scaled,max,cyc})$ and $\log(\Delta Q_\textrm{scaled,max,cyc})$, can be observed together in a multivariate bubble plot. }
\label{fig: multivariate_anomalies}
\end{figure}

Figure \ref{fig: multivariate_anomalies} presents a multivariate bubble plot illustrating anomalous battery cycles identified from combined voltage and capacity features. The plot maps $\log(\Delta V_\textrm{scaled,max,cyc})$ against $\log(\Delta Q_\textrm{scaled,max,cyc})$, where each bubble represents a cycle. Four anomalous cycles (0, 40, 147, and 148) are highlighted, showing distinct deviations from the main cluster. These anomalies correspond to abnormal patterns in both voltage and capacity behavior, demonstrating how multivariate analysis effectively captures complex deviations in battery cycling data that may not be apparent in univariate analysis.

\subsection{Anomaly Contour Map}
\label{app: anomaly_score_map}

\begin{figure*}[ht!]
\centering
\includegraphics[width=\textwidth]{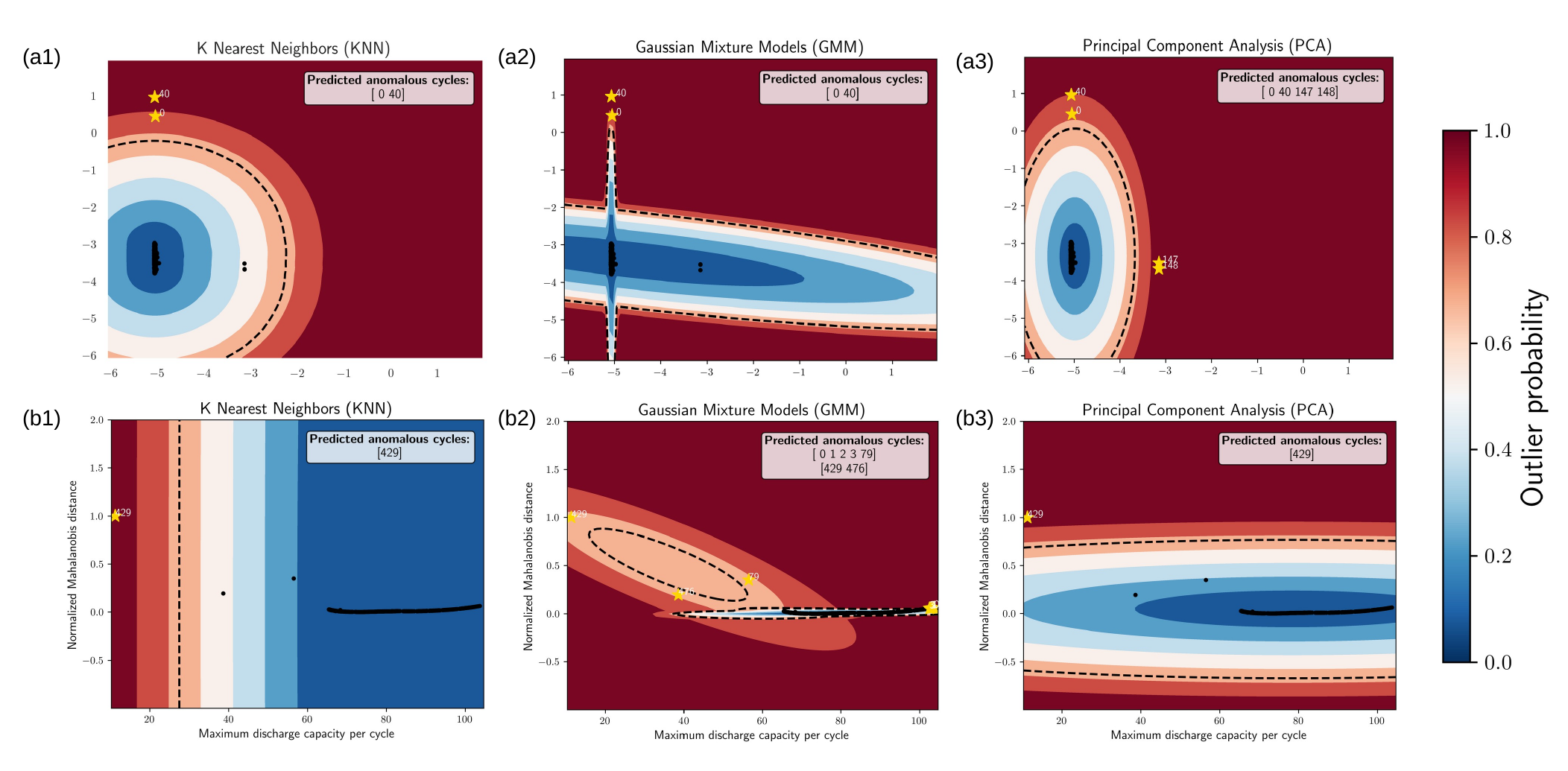}
\caption{\textbf{Anomaly score maps for three unsupervised models (\textit{i.e.,} \acrfull{KNN}, \acrfull{GMM} and \acrfull{PCA}) applied to two different datasets.} The top panel (a1-a3) shows results for the Severson dataset, with the feature space defined by $\log(\Delta Q_\textrm{scaled,max,cyc})$ and $\log(\Delta V_\textrm{scaled,max,cyc})$. The bottom panel (b1-b3) presents results for the Tohoku dataset, using maximum discharge capacity per cycle and normalized Mahalanobis distance as features. In each contour map, the color scale represents the outlier probability (from blue for normal cycles to red for highly anomalous cycles), and the dashed contour outlines the anomaly boundary. Orange stars indicate the detected anomalous cycles, with their indices listed in the inset of each panel. The predicted anomalous cycles and anomaly score maps of each model differ depending on the underlying algorithms and their hyperparameter configurations.}
\label{fig: anomaly_score_map_part2}
\end{figure*}

Figure \ref{fig: anomaly_score_map_part2} compares anomaly score maps from three unsupervised models, namely \acrshort{KNN}, \acrshort{GMM}, and \acrshort{PCA}, which are applied to the Severson (top row) and Tohoku (bottom row) datasets. The contour maps display outlier probabilities, where higher anomaly likelihood is represented by red regions, and lower anomaly likelihood is represented by blue regions. Dashed lines represent the anomaly detection boundaries, and orange stars denote predicted anomalous cycles. These results correspond to the baseline model predictions without hyperparameter tuning and illustrate how some true anomalies may be missed when using default hyperparameters for unsupervised \acrshort{ML} models.

\begin{figure*}[ht!]
\centering
\includegraphics[width=\textwidth]{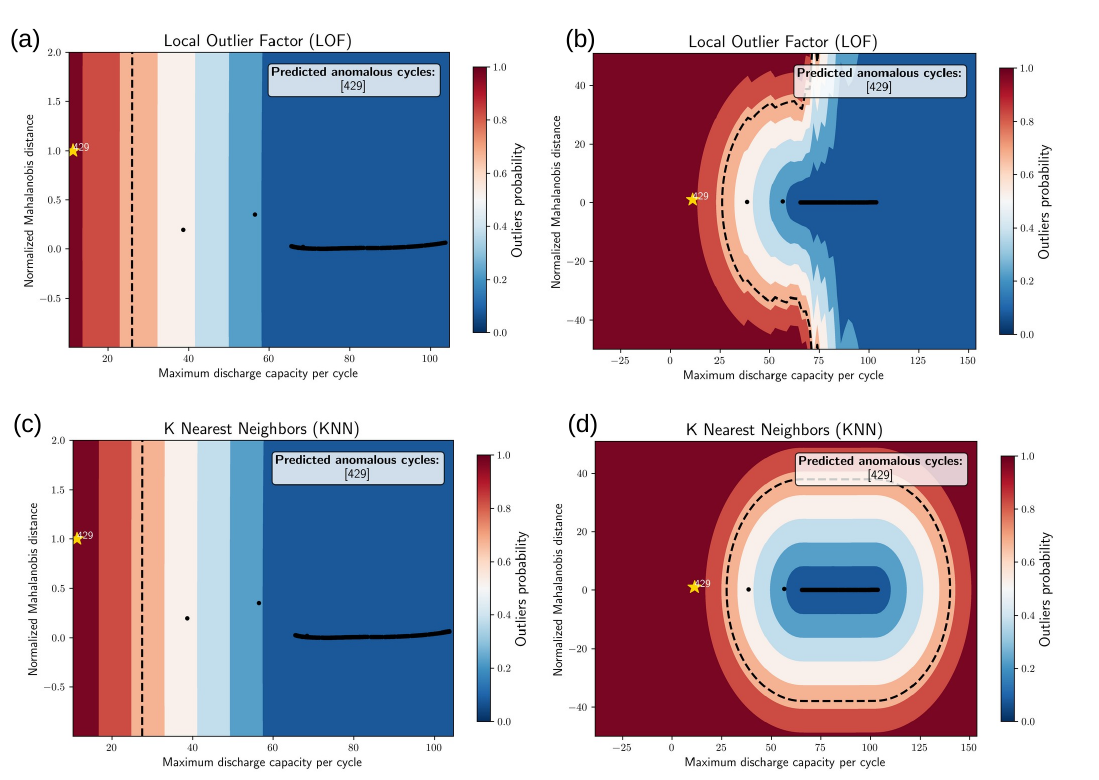}
\caption{\textbf{Anomaly score maps with different feature scaling range generated using \acrfull{LOF} and \acrfull{KNN} methods for the Tohoku datasets.} The plots display the feature space defined by maximum discharge capacity per cycle and normalized Mahalanobis distance, with color contours indicating the outlier probability (from blue for normal data to red for high anomaly likelihood). Panels (a) and (b) correspond to \acrshort{LOF}, while (c) and (d) correspond to \acrshort{KNN}, each with different feature scaling ranges. While enlarging the feature range does not change the detection outcome for both models, the anomaly contour shape of these two models can be visualized better using larger grids.}
\label{fig: knn_lof_grid_contour_map}
\end{figure*}

Figure \ref{fig: anomaly_score_map_part2}(a1) and Figure \ref{fig: anomaly_score_map_part2}(b1) illustrate the anomaly score map predicted by \acrshort{KNN} models. The default distance metric used by the \acrshort{KNN} model is the Minkowski distance with a distance order of 2 ($p=2$), corresponding to the use of the Euclidean distance metric for anomaly detection. As a result, the predicted anomaly score map by \acrshort{KNN} also illustrates the typical circular shape of a Euclidean distance (see Section \ref{sec: distance_based_anomaly_detection}). Due to a smaller feature range used for the Tohoku datasets, the anomaly score map of \acrshort{KNN} seems to differ from the anomaly score map in the Severson datasets. However, if the feature scaling range is enlarged (see Figure \ref{fig: knn_lof_grid_contour_map}(c) and \ref{fig: knn_lof_grid_contour_map}(d)), the typical circular shape of a \acrshort{KNN} model when using Minkowski distance with $p=2$ can be observed. While the anomaly contour shape of a \acrshort{KNN} anomaly detection model is predominantly influenced by the distance metric, the anomaly contour shape of \acrshort{GMM} depends on the covariance type and number of mixture components. In this work, as two input features are used to train the model, we assume two Gaussian distributions in the mixture ($n_\textrm{components} = 2$). The baseline \acrshort{GMM} model assumes that each component has its own general covariance matrix (\textit{i.e.,} $\textrm{covariance type} = \textrm{full}$). This configuration leads to the anomaly score map shown in Figure \ref{fig: anomaly_score_map_part2}(a2) and \ref{fig: anomaly_score_map_part2}(b2). Similar to the \acrshort{GMM} model, we assume \acrshort{PCA} to have two principal components. In reality, the number of principal components can be determined by the metric "explained variance ratio", which measures the proportion of the total variance in the input features that is explained by each principal component. 

Figure \ref{fig: knn_lof_grid_contour_map} compares anomaly score maps for the Tohoku datasets generated using \acrshort{LOF} and \acrshort{KNN} models under different feature scaling ranges. The feature space is defined by maximum discharge capacity per cycle and normalized Mahalanobis distance, with color contours representing outlier probability (blue for normal data and red for anomalies). Panels (a) and (b) show LOF results, while (c) and (d) show KNN results. Increasing the feature range (right-hand panels) does not change the anomaly detection outcome, but improves the visualization of the anomaly score map for these two models.

\subsection{Hyperparameters Tuning}
\label{app: hyperparameters_tuning}

\begin{figure*}[ht!]
\centering
\includegraphics[width=\textwidth]{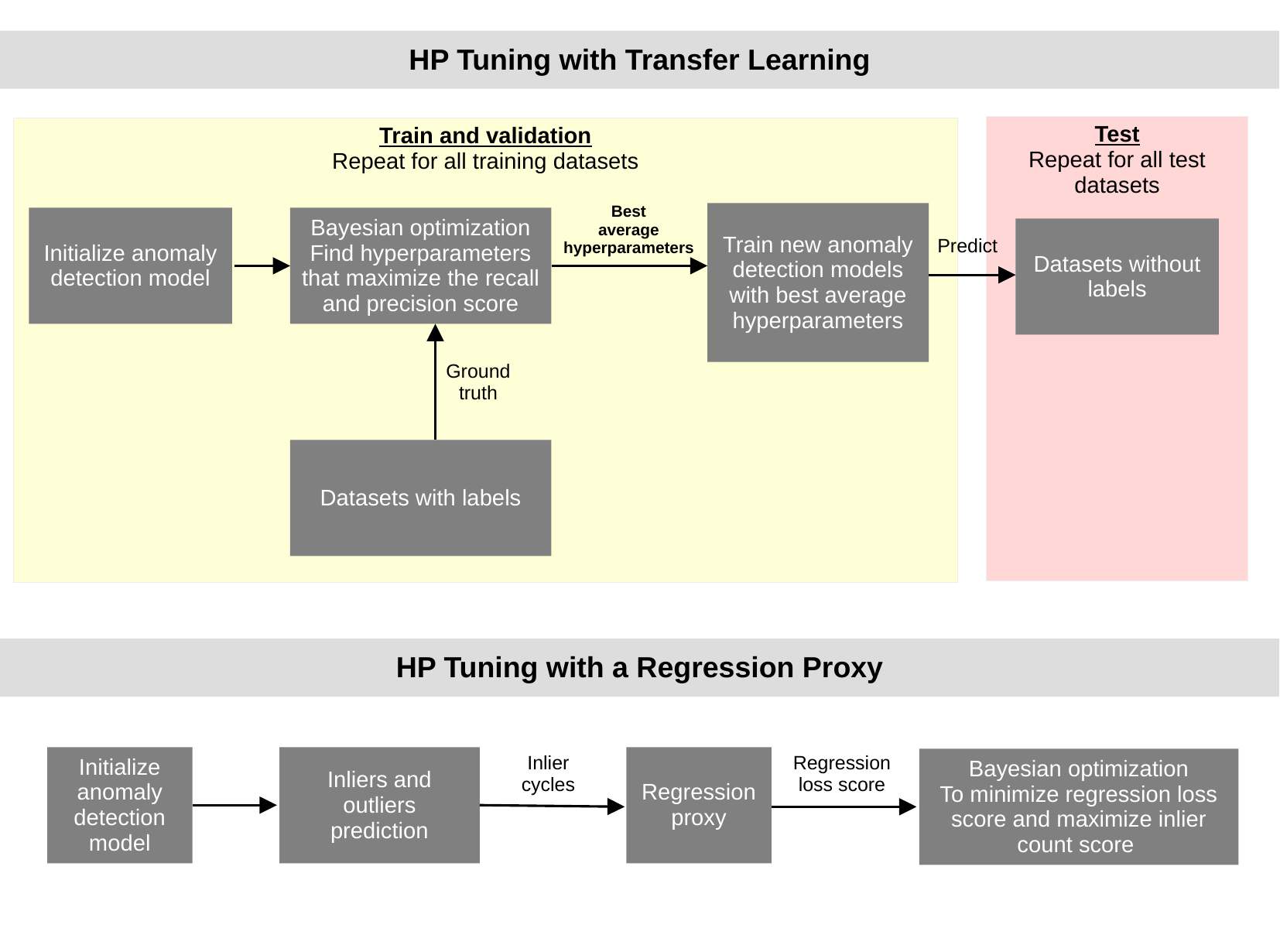}
\caption{\textbf{Comparison of hyperparameter tuning approaches for anomaly detection models.} The schematic illustrates two strategies: (top) Hyperparameter (HP) tuning with transfer learning, and (bottom) HP tuning with regression proxy. In the transfer-learning-based method, the anomaly detection model is first initialized and trained using labeled datasets. Bayesian optimization is employed to identify hyperparameters that maximize recall and precision based on ground truth labels. The best average hyperparameters from all training datasets are then transferred to initialize models for prediction on unlabeled test datasets.
In the regression-proxy approach, the model predicts inliers and outliers, and the inlier cycles are passed to a regression proxy. Bayesian optimization is then performed to minimize the regression loss score while maximizing the inlier count score, enabling hyperparameter tuning in the absence of ground truth.}
\label{fig: hp_tuning_schema}
\end{figure*}

Figure \ref{fig: hp_tuning_schema} illustrates two hyperparameter tuning methods for anomaly detection models, which are transfer-learning-based tuning and regression-proxy-based tuning. In the transfer-learning approach (top), the model is trained on labeled datasets using Bayesian optimization to find hyperparameters that maximize the recall and precision score. The best averaged hyperparameters are then applied to new, unlabeled test datasets for prediction. 

On the other hand, in the regression proxy approach (bottom), the model predicts inliers and outliers, and the inlier cycles are used as input for a regression model. Bayesian optimization is then performed to minimize the regression loss while maximizing the inlier count score, allowing effective hyperparameter tuning even without labeled data. While several methods exist for selecting a representative solution from the Pareto front, we utilize a simple yet effective frequency-based approach. This method identifies the most frequently occurring objective value pairs across optimization trials. Leveraging the Tree-structured Parzen Estimator algorithm from Bayesian optimization, which inherently samples more densely in regions associated with high-performing solutions, our approach naturally highlights stable and robust trade-offs. These high-frequency pairs (regression loss and inlier count scores) are indicative of algorithmically preferred regions in the objective space. Furthermore, the discrete nature of the objective values in our application facilitates straightforward grouping and selection of a representative compromise solution. Finally, all the trials with the identified most frequent objective value pair are aggregated based on the nature of the hyperparameter to get one single model configuration.

\newpage

\subsection{Model Evaluation Metrics}
\label{app: model_evaluation_metrics}

In this work, we evaluate the performance of different anomaly detection methods using five metrics, which include (1) Accuracy, (2) Precision, (3) Recall, (4) F1-score, and (5) \acrfull{MCC}. This section outlines the theoretical background, along with the corresponding equations for the five implemented metrics.

\textbf{Accuracy.} The first metric, Accuracy, means the number of correctly classified observations over the total number of observations:

\begin{equation}
\textrm{Accuracy} = \frac{TP+TN}{TP + TN + FP + FN},
\end{equation}
 
\noindent where $TP$ and $TN$ denote True Positives and True Negatives, whereas $FP$ and $FN$ represent False Positives and False Negatives. 

\textbf{Precision.} The second metric, Precision, represents the fraction of correctly predicted anomalies. It is defined as the ratio of true positive anomalies (correctly identified anomalies) to the total number of observations classified as anomalies (both true positives and false positives), as shown by

\begin{equation}
\textrm{Precision} = \frac{TP}{TP + FP}.
\end{equation}

It is essential to note that the metric precision only penalizes false positive predictions, meaning a model that produces no false positives has a precision of 1. 

\textbf{Recall.} On the other hand, recall (also known as sensitivity) measures the model's ability to identify all actual anomalies. It is the ratio of true positive anomalies to the total number of actual anomalies, which can be described mathematically by

\begin{equation}
\textrm{Recall} = \frac{TP}{TP + FN}.
\end{equation}

While the metric precision penalizes false positive predictions, the metric recall penalizes false negative predictions. A model that predicts no false negatives has a recall of 1. One should note that the evaluation metrics precision and recall are often in tension, meaning that improving the precision score usually reduces the recall score, and vice versa. For anomaly detection in battery cycling protocols, predicting false negatives has a much higher cost than predicting false positives. While investing additional time to sort out false alarms later may result in unnecessary interventions, missing true anomalies in real-time battery operations could lead to severe implications, such as toxic chemical reactions or even worse, battery fire incidents. In other words, prioritizing the recall score to reduce the probability of missing true anomalies is more critical for battery operations.

\textbf{F1-score.} To find the balance between optimizing the precision and recall score, we should also consider evaluating the model effectiveness based on the F1-score, which is the harmonic mean of precision and recall:

\begin{equation}
\begin{aligned}
\textrm{F1-score} &= 2 \left[ \frac{\textrm{Precision} \times \textrm{Recall}}{\textrm{Precision} + \textrm{Recall}} \right], \\
&= \frac{2TP}{2TP + FN + FP}.
\end{aligned}
\label{eq: F1-score}
\end{equation}

\textbf{\acrfull{MCC}.} The F1-score focuses on the model's ability to predict the positive classes and penalizes false positives and false negatives equally. However, as shown by Equation \ref{eq: F1-score}, F1-score ignores true negatives, which means the performance metric does not provide a complete picture of overall anomaly detection performance across all four classes. In this case, a more robust metric to evaluate the anomaly detection will be the \acrshort{MCC}-score, which includes true positives, true negatives, false positives, and false negatives, therefore, providing a balanced measure of the detector's performance:

\begin{equation}
\textrm{MCC} = \frac{TP \times TN - FP \times FN}{\sqrt{(TP + FP)(TP + FN)(TN + FP)(TN+FN)}}.
\end{equation}

Unlike the F1-score, the \acrshort{MCC}-score ranges between -1 and 1. A \acrshort{MCC}-score of +1 indicates perfect prediction, whereas a score of -1 denotes total disagreement between the prediction and actual observation across all four classes. A score of 0 indicates random prediction. Even if the class labels are swapped (\textit{i.e.}, redefining which class is positive), the MCC score will remain consistent, whereas the F1 Score will change significantly.

\end{appendices}
\end{document}